\documentclass[acmtog,nonacm]{acmart}

\renewcommand\footnotetextcopyrightpermission[1]{} %

\usepackage{booktabs} 
\usepackage{xr-hyper}

\usepackage[ruled]{algorithm2e} 

\SetAlFnt{\small}
\SetAlCapFnt{\small}
\SetAlCapNameFnt{\small}
\SetAlCapHSkip{0pt}

\setcopyright{rightsretained}
\acmJournal{TOG}
\acmYear{2021}\acmVolume{40}\acmNumber{6}\acmArticle{1}\acmMonth{12} \acmDOI{10.1145/3478513.3480487}

\usepackage{soul}
\usepackage{amsmath}

\usepackage{amsfonts}
\usepackage{mathtools}
\usepackage{bm}
\usepackage{xfrac}
\usepackage{xspace}
\usepackage[makeroom]{cancel}
\usepackage{makecell}
\usepackage{multirow}

\usepackage{subcaption}

\usepackage{color}
\usepackage{xcolor}
\usepackage{colortbl}
\usepackage{lipsum}
\usepackage{ wasysym }

\usepackage[abs]{overpic}

\definecolor{myyellow}{rgb}{1,1, 0.6}
\definecolor{myorange}{rgb}{1, 0.8, 0.6}
\definecolor{myred}{rgb}{1, 0.6, 0.6}
\definecolor{newcolor}{HTML}{3f37c9}

\newcommand\rurl[1]{%
  \href{https://#1}{\nolinkurl{#1}}%
}

\newcommand{\figref}[1]{Fig.~\ref{#1}}
\newcommand{\tabref}[1]{Tab.~\ref{#1}}
\newcommand{\secref}[1]{Sec.~\ref{#1}}

\newcommand{\equref}[1]{Eq.~(\ref{#1})}

\newcommand{\mat}[1]{\bm{\mathrm{#1}}}

\newcommand{\latent}{\mat{\omega}_i}
\newcommand{\appearance}{\mat{\psi}_i}
\newcommand{\spatial}{\mat{x}}
\newcommand{\hyper}{\mat{w}}
\newcommand{\viewdir}{\mat{d}}

\newcommand{\sethree}[0]{\mathfrak{se}(3)}

\newcommand{\etal}{{et al}.\@ }

\def\figcell#1#2#3{\begin{subfigure}{#1\columnwidth}\centering\includegraphics[width=\textwidth]{#2}
\def\temp{#3}\ifx\temp\empty\else\caption*{\centering{#3}}\fi\end{subfigure}}

\def\figcellc#1#2#3{\begin{subfigure}{#1\columnwidth}\centering\includegraphics[width=\textwidth]{#2} \def\temp{#3}\ifx\temp\empty\else\caption{#3}\fi\end{subfigure}}

\def\figcellt#1#2#3#4{\begin{subfigure}{#1\columnwidth}\centering\includegraphics[width=\textwidth,#4]{#2} \def\temp{#3}\ifx\temp\empty\else\caption*{#3}\fi\end{subfigure}}

\newcommand{\figcellbox}[1]{\fcolorbox{black}{white}{#1}}

\newcommand{\figcellboxwhite}[1]{\fcolorbox{white}{white}{#1}}

\def\figcelltb#1#2#3#4{\begin{subfigure}{#1\columnwidth}\centering\figcellbox{\includegraphics[width=\textwidth,#4]{#2}} \def\temp{#3}\ifx\temp\empty\else\caption*{#3}\fi\end{subfigure}}

\def\figcelltbwhite#1#2#3#4{\begin{subfigure}{#1\columnwidth}\centering\figcellboxwhite{\includegraphics[width=\textwidth,#4]{#2}} \def\temp{#3}\ifx\temp\empty\else\caption*{#3}\fi\end{subfigure}}

\def\figcellb#1#2#3{\begin{subfigure}{#1\columnwidth}\centering\figcellbox{\includegraphics[width=\textwidth]{#2}}
\def\temp{#3}\ifx\temp\empty\else\caption*{\centering{#3}}\fi\end{subfigure}}

\makeatletter
\newcommand{\thickhline}{%
    \noalign {\ifnum 0=`}\fi \hrule height 1pt
    \futurelet \reserved@a \@xhline
}
\newcolumntype{"}{@{\hskip\tabcolsep\vrule width 1pt\hskip\tabcolsep}}
\makeatother

\newcommand{\textimage}[4]{
	\begin{overpic}[width=2.35cm,unit=1mm,clip,trim=#1]{figures/vrig_results_v2/#2}
	\put (9.9,0.7) {\sethlcolor{white}\footnotesize\hl{$#3 / #4$}}
    \end{overpic}
}

\begin{document}
\title{HyperNeRF: A~Higher-Dimensional~Representation for~Topologically~Varying~Neural~Radiance~Fields}

\begin{anonsuppress}
\author{Keunhong Park}
\authornotemark[1]
\thanks{* Work done while the author was an intern at Google.}
\affiliation{
 \institution{University of Washington}
 \country{USA}}
\author{Utkarsh Sinha}
\affiliation{
 \institution{Google Research}
 \country{USA}}
\author{Peter Hedman}
\affiliation{
 \institution{Google Research}
 \country{USA}}
\author{Jonathan T. Barron}
\authornotemark[2]
\affiliation{
 \institution{Google Research}
 \country{USA}}
\author{Sofien Bouaziz}
\authornotemark[2]
\authornotemark[3]
\thanks{\ddag Work done while the author was at Google.}
\affiliation{
 \institution{Facebook Reality Labs}
 \country{USA}}
\author{Dan B Goldman}
\authornotemark[2]
\affiliation{
 \institution{Google Research}
 \country{USA}}
\author{Ricardo Martin-Brualla}
\authornotemark[2]
\affiliation{
 \institution{Google Research}
 \country{USA}}
\author{Steven M. Seitz}
\authornotemark[2]
\thanks{\dag Sorted alphabetically.}
\affiliation{
 \institution{University of Washington, Google Research}
 \country{USA}}

\renewcommand\shortauthors{Park, K. et al}
\end{anonsuppress}

\begin{abstract}
Neural Radiance Fields (NeRF) are able to reconstruct scenes with unprecedented fidelity, and various recent works have extended NeRF to handle dynamic scenes. A common approach to reconstruct such non-rigid scenes is through the use of a learned deformation field mapping from coordinates in each input image into a canonical template coordinate space. However, these deformation-based approaches struggle to model changes in topology, as topological changes require a discontinuity in the deformation field, but these deformation fields are necessarily continuous.
We address this limitation by lifting NeRFs into a higher dimensional space, and by representing the 5D radiance field corresponding to each individual input image as a slice through this ``hyper-space''.
Our method is inspired by level set methods, which model the evolution of surfaces as slices through a higher dimensional surface. 
We evaluate our method on two tasks: (i) interpolating smoothly between ``moments'', i.e., configurations of the scene, seen in the input images while maintaining visual plausibility, and (ii) novel-view synthesis at fixed moments. We show that our method, which we dub \emph{HyperNeRF}, outperforms existing methods on both tasks. 
Compared to Nerfies, HyperNeRF reduces average error rates by 4.1\% for interpolation and 8.6\% for novel-view synthesis, as measured by LPIPS. 
Additional videos, results, and visualizations are available at \rurl{hypernerf.github.io}.
\end{abstract}

%
%

%
%



\fboxsep=0pt 
\fboxrule=0.4pt 


\begin{teaserfigure}
    
	\captionsetup[sub]{labelformat=parens}
	\includegraphics[width=\textwidth]{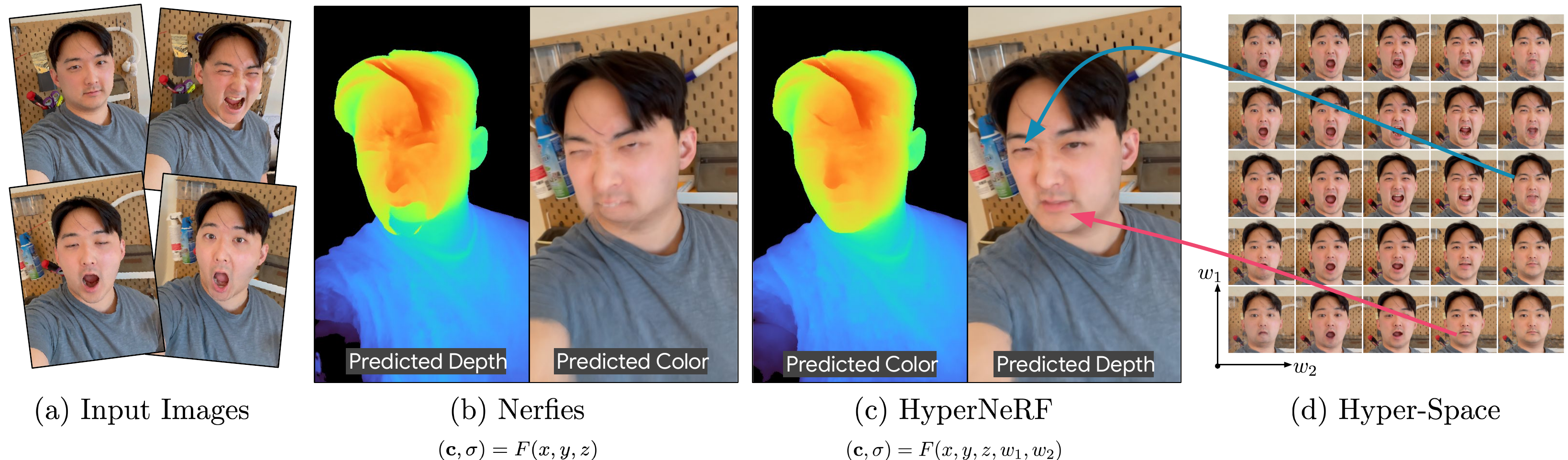}
  \caption{
  Neural Radiance Fields (NeRF)~\cite{mildenhall2020nerf} when endowed with the ability to handle deformations~\cite{park2020nerfies} are able to capture non-static human subjects, but often struggle in the presence of significant deformation or topological variation, as evidenced in (b).
  By modeling a family of shapes in a high dimensional space shown in (d), our Hyper-NeRF model is able to handle topological variation and thereby produce more realistic renderings and more accurate geometric reconstructions, as can be seen in (c).
  }
  \label{fig:teaser}
\end{teaserfigure}

\maketitle

\section{Introduction}
\label{sec:introduction}

Many real-world motions involve changes in topology.  Examples include cutting a lemon, or tearing a piece of paper.  While not strictly a genus change, motions like closing your mouth cause changes in surface connectivity that can also be considered ``topological’’.  Such changes in topology often cause problems for algorithms that seek to reconstruct moving three dimensional scenes, as they cause motion discontinuities or singularities.

A clever approach for addressing topology changes is to represent the 3D scene as a {\em level set} in a 4D volume.  Pioneered in the late 1980s \cite{osher1988fronts}, level set methods model moving scenes as static objects in a higher dimensional ambient space, and topological changes as {\em smooth} (rather than discontinuous) transformations.

In this paper, we adapt the level set framework for deformable neural radiance fields \cite{park2020nerfies}, to generate photorealistic, free-viewpoint renderings of objects undergoing changes in topology.  In doing so, we use modern tools like MLPs to significantly generalize the classical level set framework in key ways.  First, whereas classical level sets add a single ambient dimension, we can add any number of ambient dimensions to provide more degrees of freedom.  Second, rather than restrict level sets to hyper-planes, as is traditional, we allow general, curved slicing manifolds, represented through MLPs.  

Our approach models each observation frame as a nonplanar slice through a hyperdimensional NeRF --- a HyperNeRF. Previous methods using higher dimensional inputs require either substantial regularization or additional supervision. In contrast, our method retains a deformation field, which has previously demonstrated strong ability to fuse information across observations, and instead of regularizers, we use an optimization strategy that encourages smooth behavior in the higher dimensions. This enables our method to reconstruct  high-quality geometry even when some poses are observed from only a small range of angles. 

Our method enables users to capture photorealistic free-viewpoint reconstructions of a wide range of challenging deforming scenes from {\em monocular video}, e.g., waving a mobile phone in front of a moving scene. %
We demonstrate the quality of our method on two tasks: (i) interpolating smoothly between ``moments'' while maintaining visual plausibility, and (ii) novel-view synthesis with fixed moments. Our method, \emph{HyperNeRF}, produces sharper, higher quality results with fewer artifacts on both tasks.

\section{Related Work}

\subsection{Non-Rigid Reconstruction}

A common approach in non-rigid reconstruction techniques is to decompose a scene into a canonical model of scene geometry (which is fixed across frames) and a deformation model that warps the canonical scene geometry to reproduce each input image.
The difficulty of this task depends heavily on the inherent ambiguity of the problem formulation. Using only a monocular video stream is convenient and inexpensive, but also introduces significant ambiguity which must be ameliorated through the use of factorization \cite{bregler2000recovering} or regularization \cite{torresani2008nonrigid}.
On the opposite end of the spectrum, complicated and expensive capture setups using multiple cameras and depth sensors can be used to overconstrain the problem, thereby allowing 3D scans to be registered and fused to produce high quality results \cite{collet2015high, dou2016fusion4d}.
Machine learning techniques have been used effectively for non-rigid reconstruction, when applied to direct depth sensors~\cite{bozic2020neural, schmidt2015dart}.
Our method requires only monocular RGB images from a conventional smartphone camera as input --- no depth sensors or multi-view capture systems are required.

Several works have used learning techniques to solve for deformation based models of shape geometry~\cite{Niemeyer_2019_ICCV,jiang2020shapeflow}, though because these works model only geometry and not radiance, they cannot be applied to RGB image inputs and do not directly enable view synthesis.
Yoon~\etal~\cite{yoon2020novel} use a combination of multi-view cues as well as learned semantic priors in the form of monocular depth estimation to
recover dynamic scenes from moving camera trajectories. In contrast, our approach requires no training data other than the input sequence being used as input, as is typical in NeRF-like models.
Neural Volumes~\cite{lombardi2019neural} represents deformable scenes using a volumetric 3D voxel grid and a warp field, which are directly predicted by a convolutional neural network.
As we will demonstrate, Neural Volumes's high fidelity output relies on the use of dozens of synchronized cameras, and does not generalize well to monocular image sequences.
The Deformable NeRF technique of \citet{park2020nerfies} uses NeRF-like learned distortion fields alongside the radiance fields of \citet{mildenhall2020nerf} to recover ``nerfies'' of human subjects, and is capable of generating photorealistic synthesized views of a wide range of non-stationary subjects. We build directly upon this technique, and extend it to better support subjects that not only move and deform, but that also vary topologically.

\subsection{Neural Rendering}

The nascent field of ``neural rendering'' aims, broadly, to use neural networks to render images of things. This is an emerging area of study that is changing rapidly, but progress in the field up through 2020 is well-documented in the survey report of \citet{tewari2020neuralrendering}. The dominant paradigm in this field has, until recently, been framing the task of synthesizing an image as a sort of ``image to image translation'' task, in which a neural network is trained to map some representation of a scene into an image of that scene~\cite{isola2017image}. This idea has been extended to incorporate tools and principles from the graphics literature, by incorporating reflectance or illumination models into the ``translation'' process~\cite{Meka2019, sun2019single}, or by using proxy geometries~\cite{thies2019deferred, aliev2019neural} or conventional renderings~\cite{kim2018deepvideoportraits, fried2019text, martin2018lookingood, meshry2019neural} as a harness with which neural rendering can then be guided. Though these techniques are capable of producing impressive results, they are often hampered by a lack of consistency across output renderings --- when rendering is performed independently by a ``black box'' neural network, there is no guarantee that all renderings of scene will correspond to a single geometrically-consistent 3D world.

Research within neural rendering has recently begun to shift away from this ``image to image translation'' paradigm and towards a ``neural scene representation'' paradigm. Instead of ``rendering'' images using a black box neural network that directly predicts pixel intensities, scene-representation approaches use the weights of a neural network to directly model some aspect of the physical scene itself, such occupancy~\cite{mescheder2019occupancy}, distance ~\cite{park2019deepsdf}, surface light field~\cite{yariv2020multiview}, or a latent representation of appearance~\cite{sitzmann2019deepvoxels, sitzmann2019srns}. The most effective approach within this paradigm has been constructing a neural radiance field (NeRF) of a scene, and using a conventional multilayer perceptron (MLP) to parameterize volumetric density and color as a function of spatial scene coordinates~\cite{mildenhall2020nerf}.
NeRF's usage of classical volumetric rendering techniques has many benefits in addition to enabling photorealistic view synthesis: renderings from NeRF must correspond to a single coherent model of geometry, and the gradients of volumetric rendering are well-suited to gradient-based optimization. Note that, in NeRF, a neural network is not used to render an image --- instead, an analytical physics-based volumetric rendering engine is used to render a scene whose geometry and radiance happen to be parameterized by a neural network.

Though NeRF produces compelling results on scenes in which all content is static, it fails catastrophically in the presence of moving objects. As such, a great deal of recent work has attempted to extend NeRF to support dynamic scenes. We will separate these NeRF variants into two distinct categories: \emph{deformation-based} approaches apply a spatially-varying deformation to some canonical radiance field~\cite{park2020nerfies, tretschk2021nonrigid, pumarola2020dnerf}, and \emph{modulation-based} approaches directly condition the radiance field of the scene on some property of the input image and modify it accordingly ~\cite{xian2020videonerf, li2020nsff, li2021neural, gafni2021nerface}.

\emph{Deformation-based} NeRF variants follow the tradition established by the significant body of research on non-rigid reconstruction~\cite{newcombe2015dynamicfusion}, and map observations of the subject onto a template of that subject. Nerfies~\cite{park2020nerfies}, D-NeRF~\cite{pumarola2020dnerf}, and NR-NeRF~\cite{tretschk2021nonrigid} all define a continuous deformation field which maps observation coordinates to canonical coordinates which are used to query a template NeRF.
Because multiple observations of the subject are used to reconstruct a single canonical template, and only the deformation field is able to vary across images, this approach  yields a well-constrained optimization problem similar to basic NeRF.
Similar to how NeRF parameterizes radiance and density using a coordinate-based MLP, these deformation-based NeRF variants use an MLP to parameterize the deformation field of the scene, and are thereby able to recover detailed and complicated deformation fields. However, this use of a \emph{continuous} deformation field means that these techniques are unable to model any topological variations (e.g., mouth openings) or transient effects (e.g, fire). Topological openings or closings require a discontinuity in the deformation field at the seam of the closing, which MLPs cannot model easily. This is a consequence of the fact that coordinate-based MLPs with positional encoding perform interpolation with a band-limited kernel, when viewed through the lens of neural tangent kernels~\cite{tancik2020fourfeat}.

\emph{Modulation-based} or \emph{latent-conditioned} NeRF variants adopt the well established technique of conditioning a neural network with a latent code to modulate its output. 2D generative models such as GANs~\cite{goodfellow2014generative} or VAEs~\cite{kingma2013auto} input latent code to a 2D CNN which then outputs a corresponding image. This technique is also used to encode a space of 3D shapes; for example, DeepSDF~\cite{park2019deepsdf} and IM-NET~\cite{chen2019learning} modulate a signed distance field (SDF) based on input latent codes, while Occupancy Networks~\cite{mescheder2019occupancy} modulate an occupancy field. Similarly, SRNs~\cite{sitzmann2019srns} modulate a latent representation which is decoded by an LSTM.

Following these footsteps, instead of modeling the scene as a single NeRF that is warped to explain individual images, modulation-based NeRF techniques provide additional information (e.g., the image's timestamp or a latent code) as input to the MLP, directly changing the radiance field of the scene.
As such, these techniques are capable of modeling any deformation, topological change, or even complex phenomena such as fire. However, because these modulated NeRFs may have completely different radiance and density across input images, these techniques result in a severely under-constrained problem and allows for trivial, non-plausible solutions. This issue can be addressed by providing additional supervision such as depth and optical flow (as in Video-NeRF~\cite{xian2020videonerf} and NSFF~\cite{li2020nsff}) or by using a multi-view input captured from 7 synchronized cameras (as in DyNeRF~\cite{li2021neural}).
For facial avatars, \citet{gafni2021nerface} avoids this issue by using a face-centric coordinate frame from a 3D morphable face model~\cite{thies2016face2face}.
NeRF in the Wild also uses a similar modulated approach (albeit for a different task) in which latent codes are optimized and provided as input to an MLP, which also introduces ambiguities that are addressed by allowing those codes to only modify radiance but not density~\cite{martinbrualla2020nerfw}.

Our method can be thought of as a combination of deformation-based and modulation-based approaches: we use deformations to model motion in the scene, resulting in a well-behaved optimization, but we also extend NeRF's 3D input coordinate space to take additional higher-dimension coordinates as input, and allow for deformations along the higher dimensions as well as the spatial dimensions. As we will show, this approach is able to use the higher dimensions to capture changes in object topology, which a strictly deformation-based approach would be unable to model.

\section{Modeling Time-Varying Shapes}

\begin{figure}[t]
	\begin{subfigure}[b]{0.42\columnwidth}
    	\includegraphics[width=0.9\textwidth]{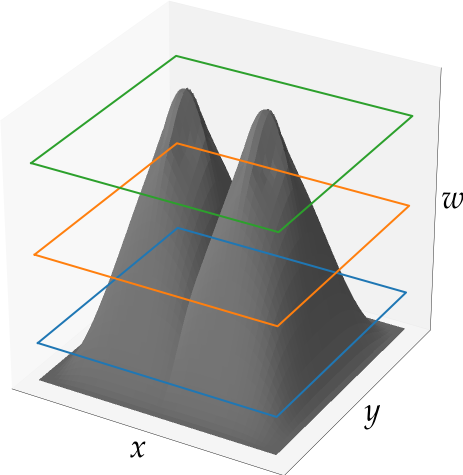}
    	\caption{3D Auxiliary Function}
        \label{fig:level_set_example_surf}
	\end{subfigure}
	\begin{subfigure}[b]{0.42\columnwidth}
    	\includegraphics[width=0.9\textwidth]{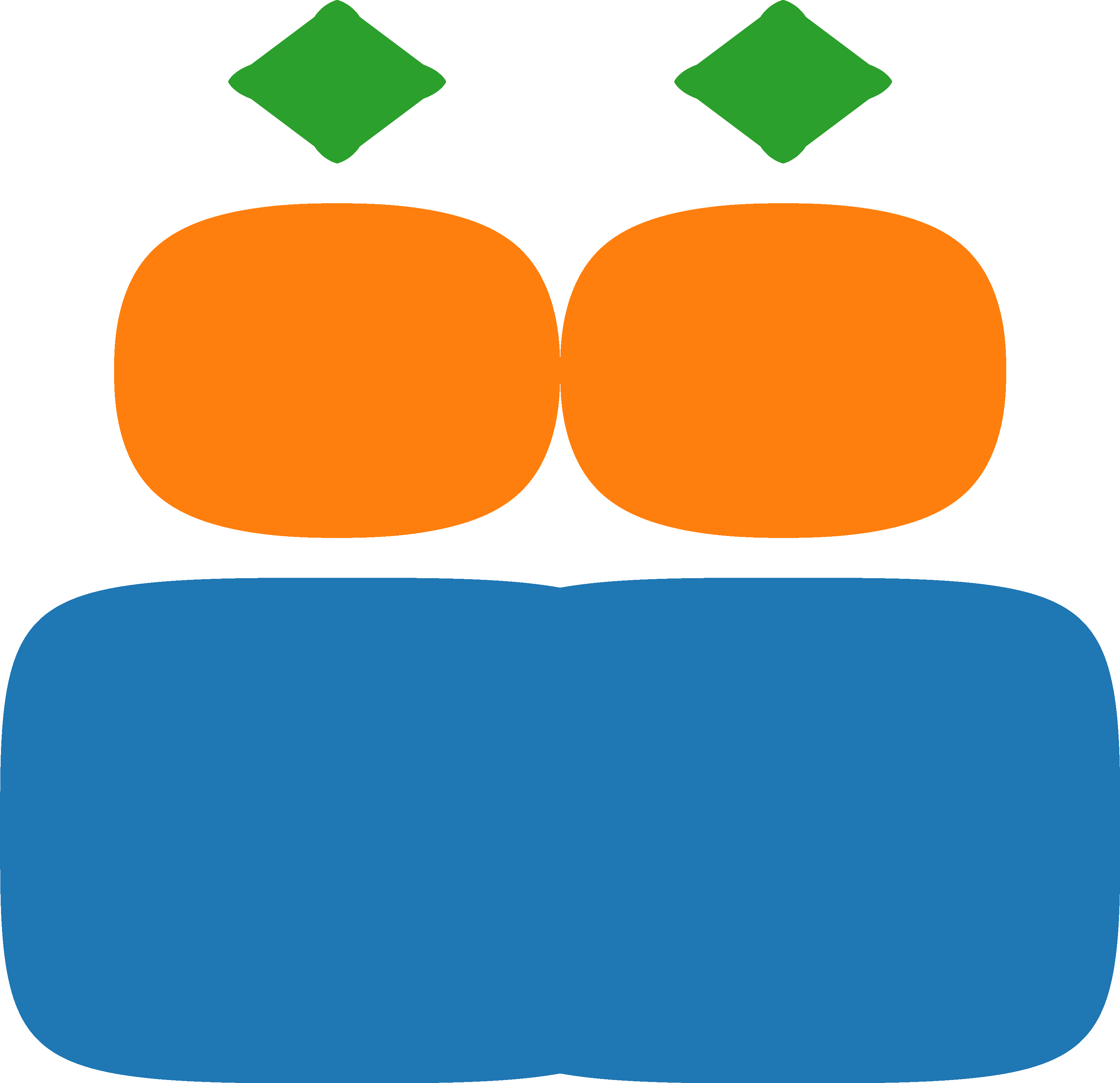}
    	\caption{2D Sliced Shapes}
        \label{fig:level_set_example_slice}
	\end{subfigure}
	\caption{
	Level-set methods provide a means to model a family of topologically-varying shapes (b) as slices of a higher dimensional auxiliary function (a).
    \label{fig:level_set_example}
    }
    \vspace{-12pt}
\end{figure}

Our method represents changes in scene topology by providing a NeRF with a higher-dimensional input. To provide an intuition and a justification for our formulation in higher dimensions, this section introduces level set methods which our method draws inspiration from. Note that our method is \emph{not} a level set method, though we will apply intuitions gained from these examples to our NeRF setting.
Note that the insights gained from the level set perspective also apply to all existing latent-conditioned neural representations, such as DeepSDF~\cite{park2019deepsdf}, which are equivalent to the axis-aligned slicing plane formulation described in \secref{sec:slicing_surfaces}. We believe this provides a good visual intuition for existing methods.

There are two common approaches for mathematically representing the surface of an object: a surface can be defined \emph{explicitly}, perhaps with a polygonal mesh, or \emph{implicitly}\footnote{Confusingly, coordinate-based approaches such as NeRF are sometimes referred to in the literature as ``implicit'' models, in reference to the idea that they encode scenes ``implicitly'' using the weights of a neural network. This new use of ``implicit'' is distinct from its common use in the literature, so to avoid confusion we will only use ``implicit'' according to its meaning in the level set literature.}, perhaps as the level set of a continuous function. Explicit representations of shape, though effective and ubiquitous, are often poorly suited to topological variation of a surface: slicing a polygonal mesh into two halves, for example, likely requires creating new vertices and redefining the edge topology of the mesh. This sort of variation is particularly difficult to express in the context of gradient-based optimization methods such as NeRF, as this transformation is discontinuous and therefore not easily differentiated.
In contrast, implicit surfaces provide a natural way to model the evolution of a surface in the presence of topological changes, such as when the surface develops a hole or splits into multiple pieces.

\subsection{Level Set Methods}
\label{sec:level_set_methods}

Level set methods model a surface implicitly as the zero-level set of an auxiliary function~\cite{osher1988fronts}. For example, a 2D surface can be defined as $\Gamma = \{(x, y) | S(x, y) = 0\}$, where $S : (x, y) \rightarrow s$ is a signed-distance function with $s>0$ for points inside the surface, and $s<0$ for points outside the surface. To model a surface that varies topologically with respect to some additional dimension (such as ``time''), one can add an additional dimension $w$, and thereby define a 3D surface $\Gamma = \{(x, y, w) | S(x, y, w) = 0\}$. We will refer to the space by these additional dimensions as the ``ambient'' space. The 2D surface at some $w_i$ ambient coordinate can then be expressed as the 2D cross-section of the 3D surface $\Gamma$ obtained by slicing it with the plane passing through $w=w_i$. See \figref{fig:level_set_example} for an illustration.

\begin{figure}[t]
	\centering
	\begin{subfigure}[b]{0.95\columnwidth}
    	\figcellt{0.24}{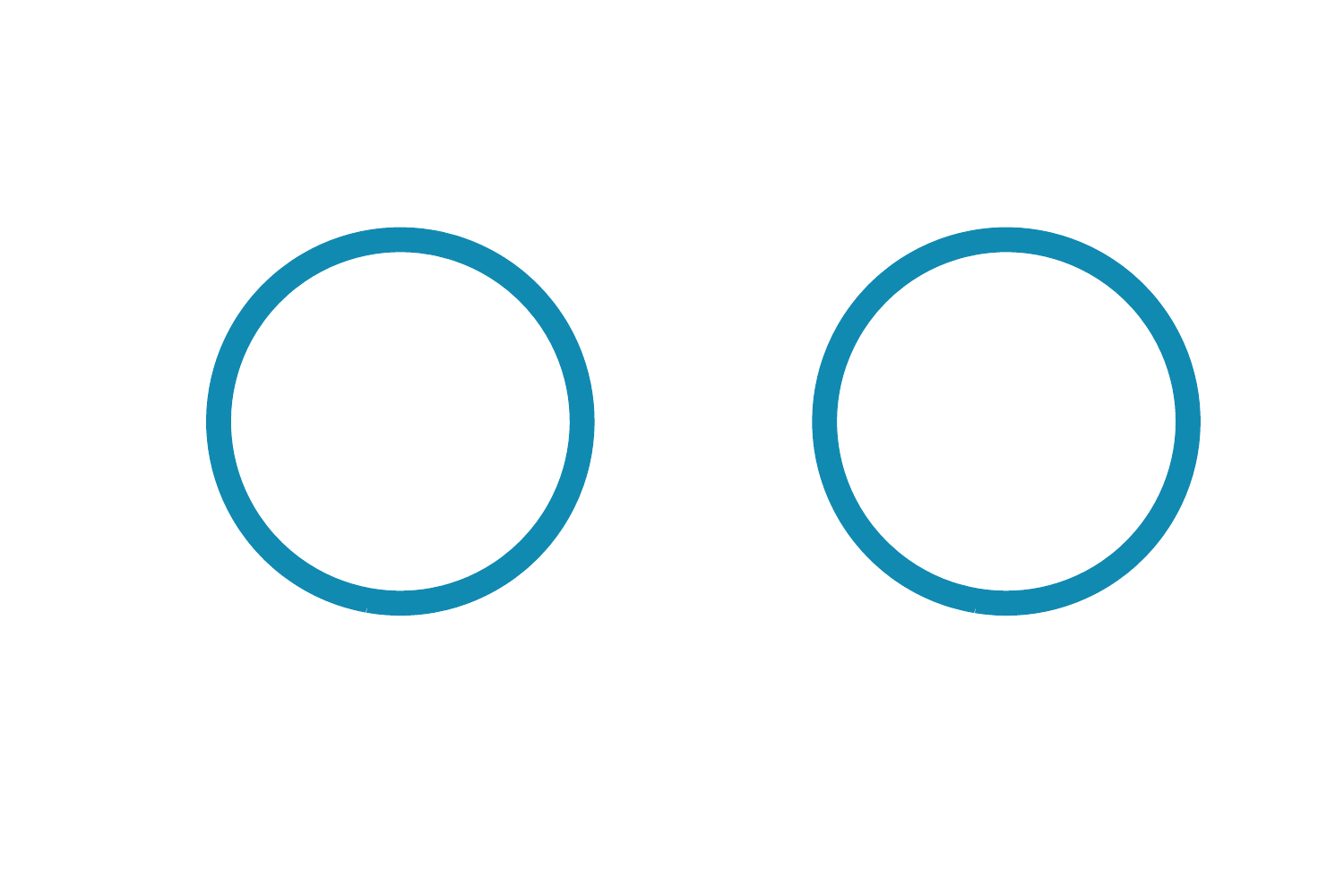}{Shape 1}{clip,trim=40 90 30 40}
    	\figcellt{0.24}{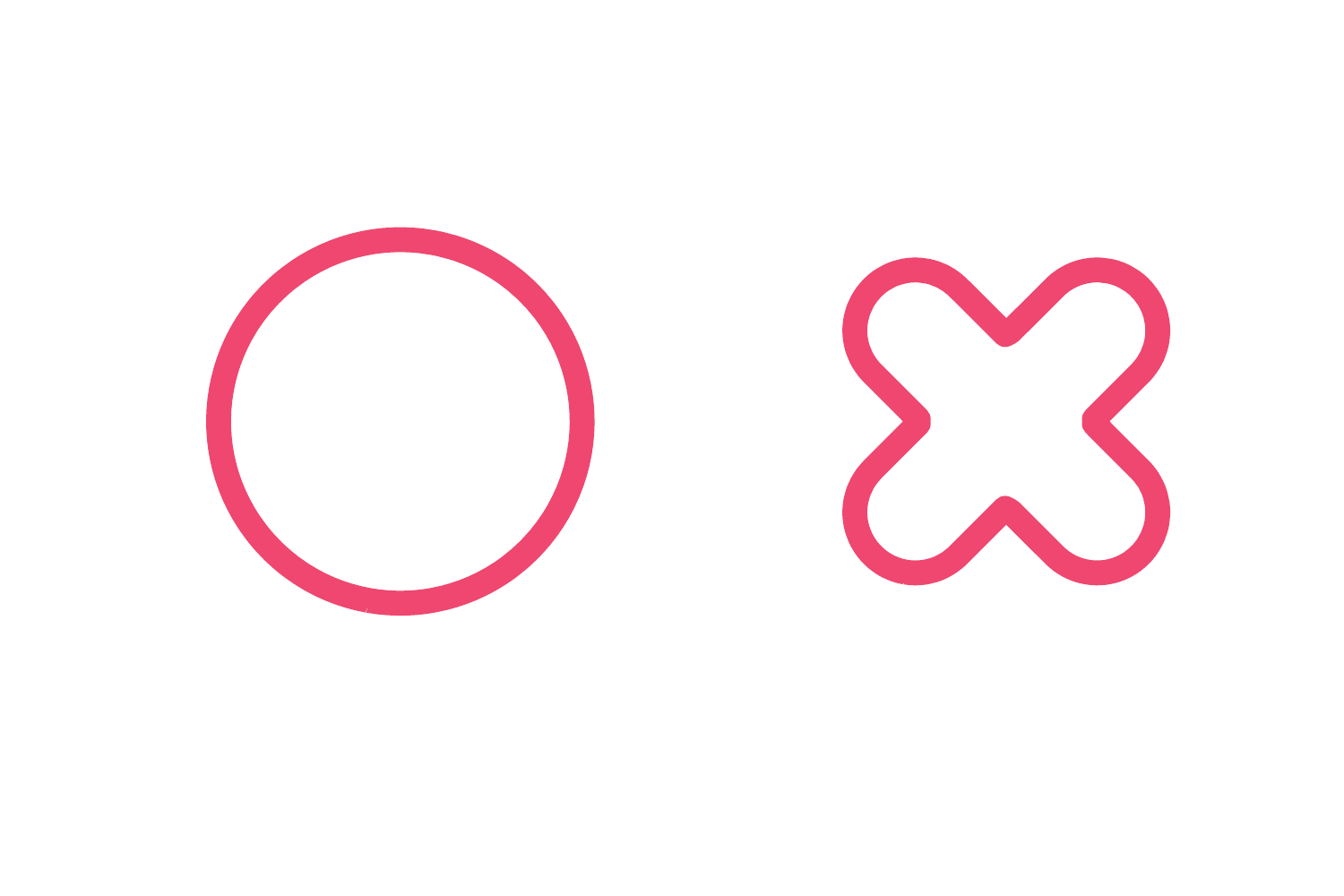}{Shape 2}{clip,trim=40 90 30 40}
    	\figcellt{0.24}{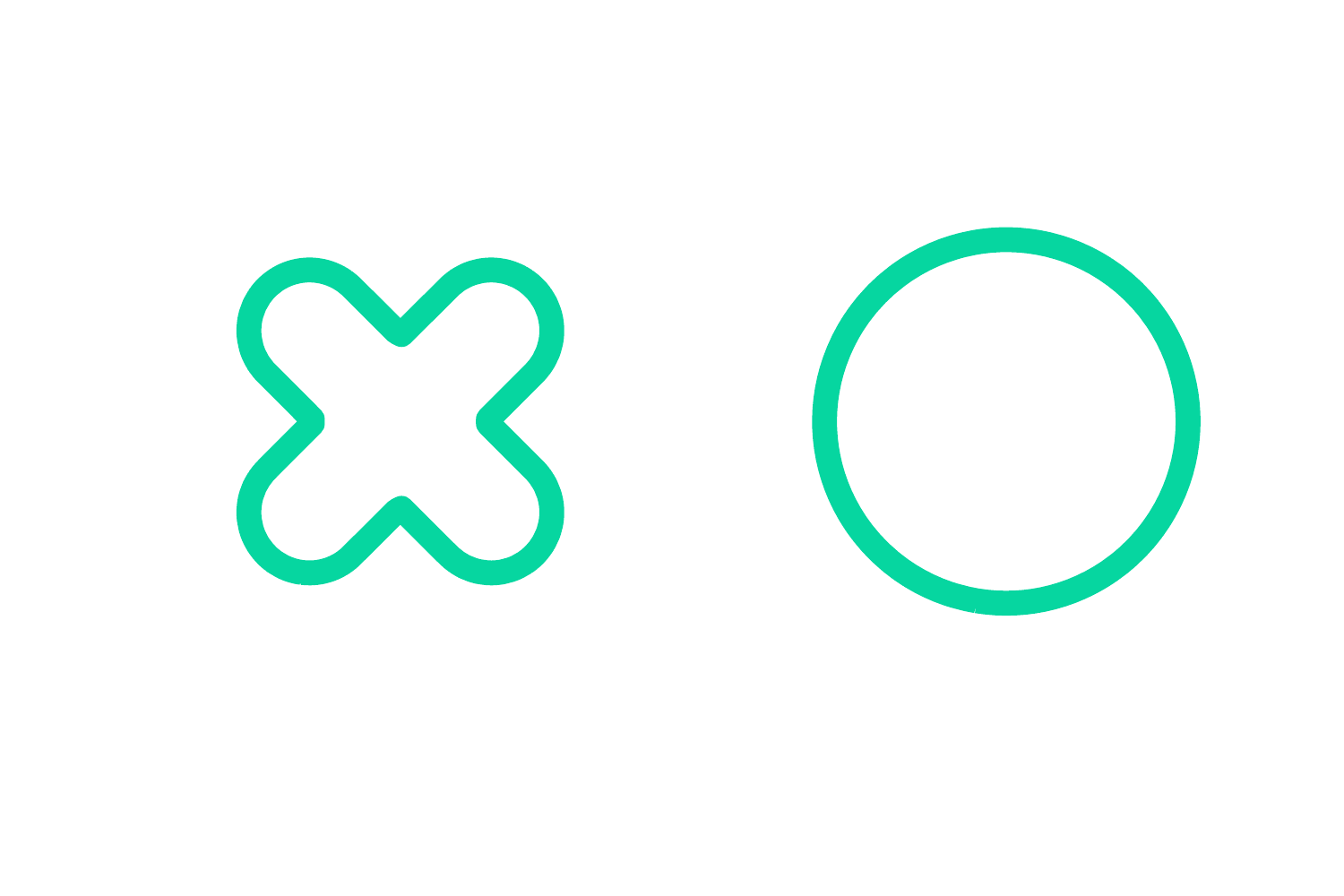}{Shape 3}{clip,trim=40 90 30 40}
    	\figcellt{0.24}{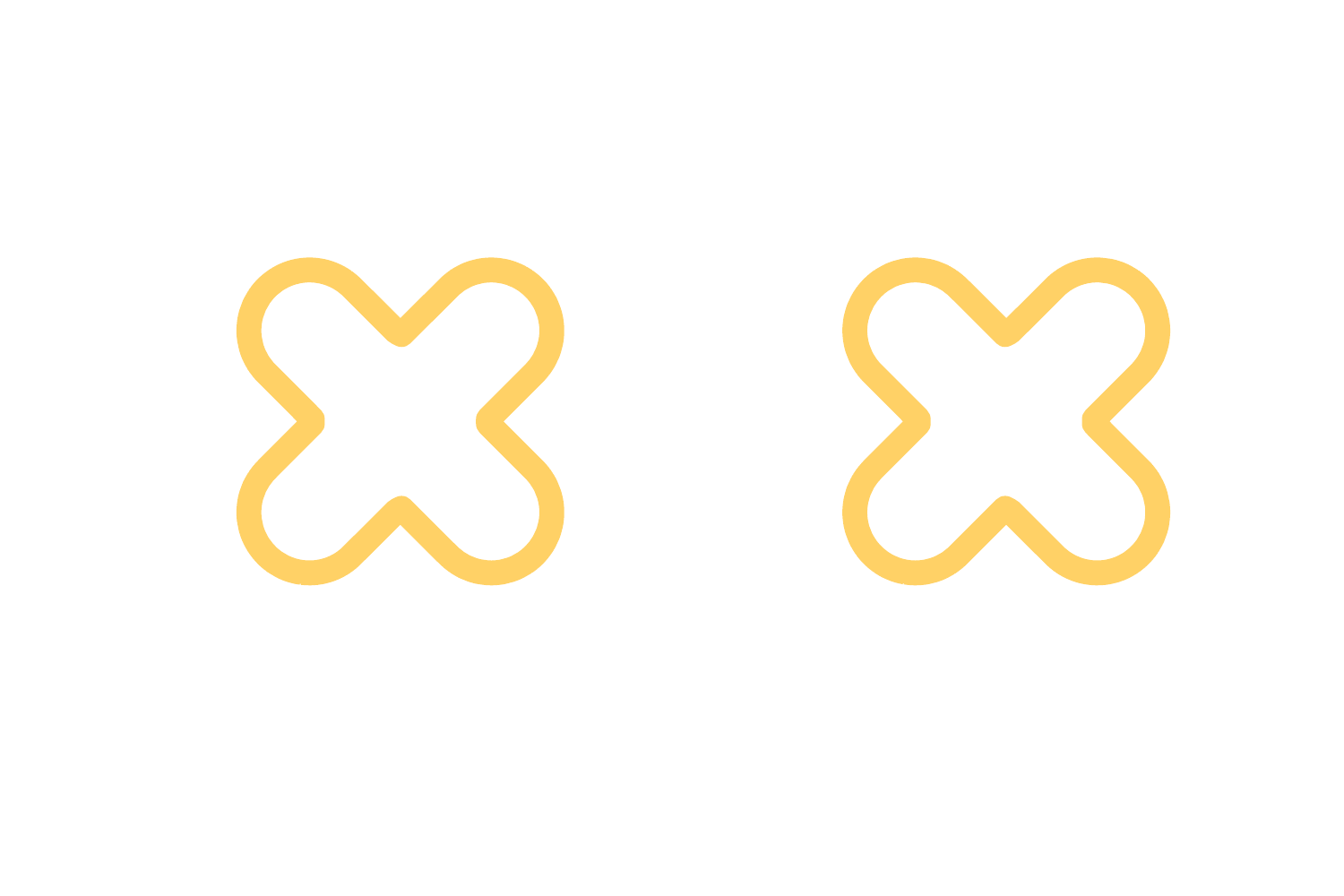}{Shape 4}{clip,trim=40 90 30 40}
    	\caption{Input Shapes}
        \label{fig:level_set_slice_input}
	\end{subfigure}
	\begin{subfigure}[b]{0.47\columnwidth}
    	\includegraphics[width=\textwidth]{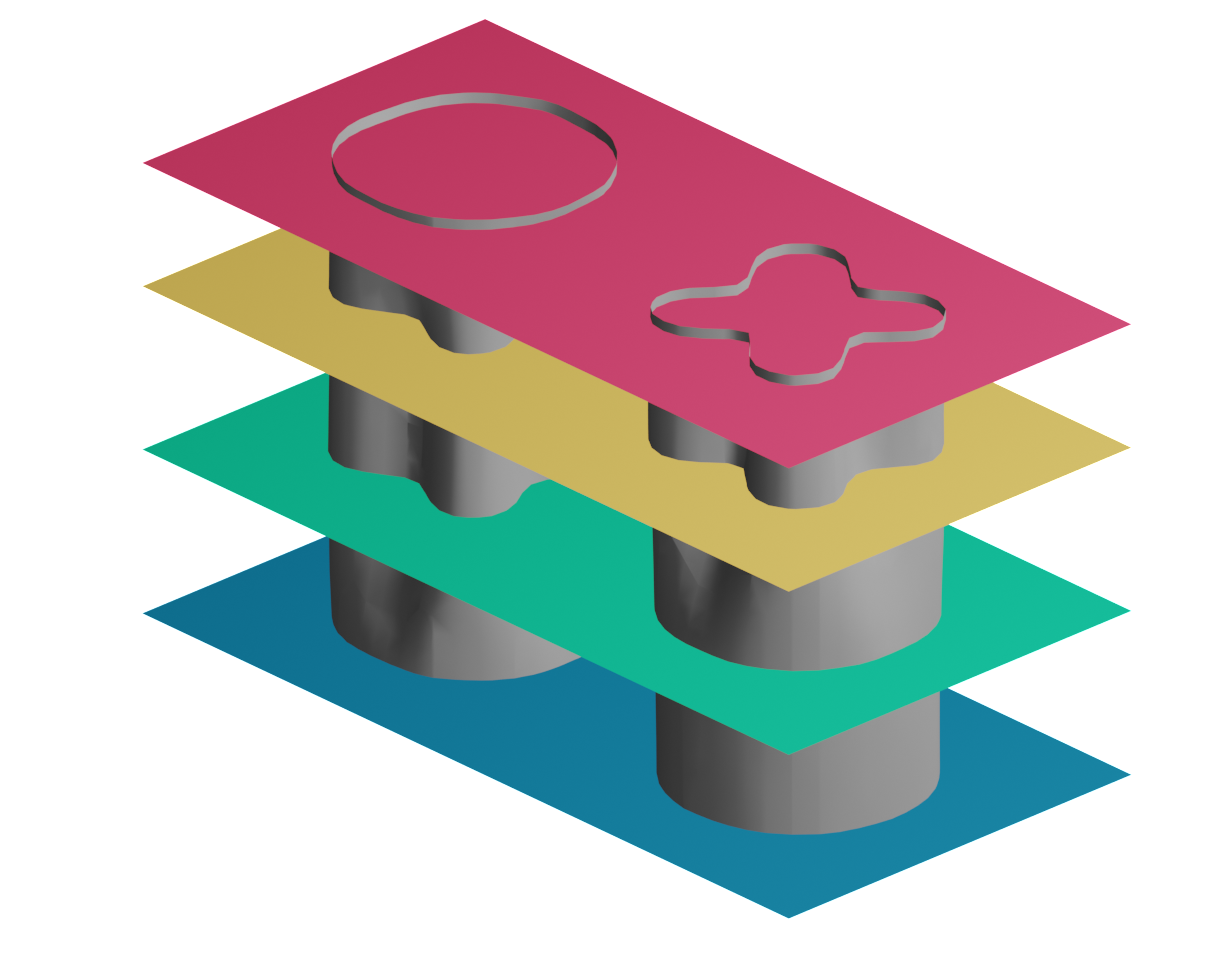}
    	\caption{Axis-aligned Slicing Plane (AP)}
        \label{fig:level_set_slice_ap}
	\end{subfigure}
	\begin{subfigure}[b]{0.47\columnwidth}
    	\includegraphics[width=\textwidth]{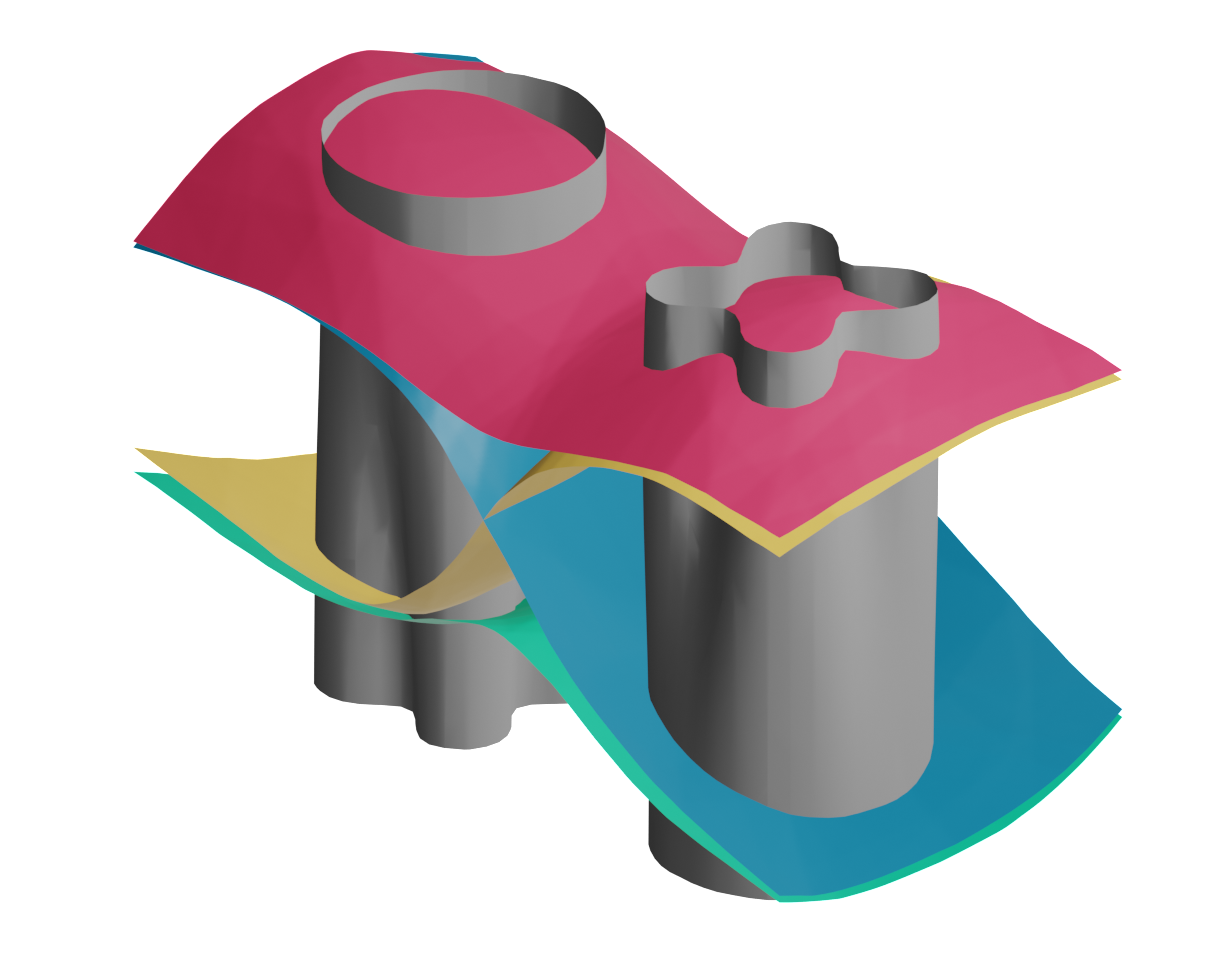}
    	\caption{Deformable Slicing Surface (DS)}
        \label{fig:level_set_slice_bs}
	\end{subfigure}
    
    \caption{Level-set methods model shapes as slices of a higher-dimensional surface, which we call an ``ambient surface''. We present two slicing methods: \emph{axis-aligned planes} (AP) slice the surface perpendicular to the higher-dimensional axis; \emph{deformable surfaces} (DS) can deform and thus different parts of the shape can reference varying parts of hyper-space. Axis-aligned planes require a copy of each shape separately, while deformable surfaces can share information and resulting in simpler ambient surfaces. We encode deformable slicing surfaces in the weights of an MLP.}
    \label{fig:level_set_slice}
\end{figure}

This idea can be extended to learn a collection of shapes. Given a set of implicitly-defined 2D shapes, one can learn an SDF $\Gamma$ which contains all such 2D shapes as individual slices of a canonical 3D surface. Following DeepSDF~\cite{park2019deepsdf}, let $S : (x,y,w) \rightarrow s$ be an MLP, where the per-shape ambient coordinates $\{w_i\}$ are learned as an embedding layer~\cite{bojanowski2018optimizing}. The weights of the MLP and the hyper coordinates can then be optimized using gradient descent. \figref{fig:level_set_slice} shows four different shapes being encoded in a single 3D SDF using this MLP. As with time-varying shapes, this approach gives us a natural way to interpolate between the shapes by interpolating between the learned slicing planes. As shown by DeepSDF, this formulation can be extended to an arbitrary number of spatial and ambient dimensions. For example, by formulating this same problem with 3 spatial dimensions and 256 ambient dimensions, we can learn 3D shapes as 3-dimensional cross sections of a 259-dimensional hyper-surface.

\subsection{Deformable Slicing Surfaces}
\label{sec:slicing_surfaces}

Level set methods work by ``slicing'' through a function, usually with an axis-aligned plane. The plane of this slice spans the spatial axes (the $x$ and $y$ axes in our 2D/3D example), and occupies a single ambient coordinate (the $w$ axis in our example) at all points. A consequence of using an axis-aligned slice is that every desired output shape must exist as a cross section cut by the slicing plane. In certain circumstances, this can lead to an inefficient use of space. For example, consider the set of shapes shown in \figref{fig:level_set_slice_input}, which are different permutations of the same two shapes. If axis-aligned slices are used, the ambient surface must contain a \emph{copy} of each of the four permutations as shown in \figref{fig:level_set_slice_ap}. This is inefficient considering there are only two possible sub-shapes---a circle and a cross. 

To address this, we introduce another MLP which encodes a \emph{deformable} slicing surface, such that the output shape is the cross-section sliced by a non-planar surface as in \figref{fig:level_set_slice_bs}. This allows different spatial locations to reference different parts of the ambient coordinate space, resulting in a more compact representation in hyperspace. We define the deformable slicing surface as an MLP $H : (\spatial,\mat{\omega}_i) \rightarrow \hyper$, where $\spatial$ is a spatial position, $\hyper$ is a position along the ambient axes, and $\mat{\omega}_i$ is a per-input latent embedding (whose dimensionality and meaning need not match that of the ambient coordinate). The SDF is then queried at the coordinate obtained by concatenating $\spatial$ and $H(\spatial,\mat{\omega}_i)$. As shown in \figref{fig:level_set_slice_bs}, this parameterization is able to slice through the ambient dimensions to model arbitrary mixtures of shapes. See the supplementary materials for details on the experiment for \figref{fig:level_set_slice}.

\begin{figure}[t]
	\centering
	\begin{subfigure}[b]{0.45\columnwidth}
    	\includegraphics[width=\textwidth]{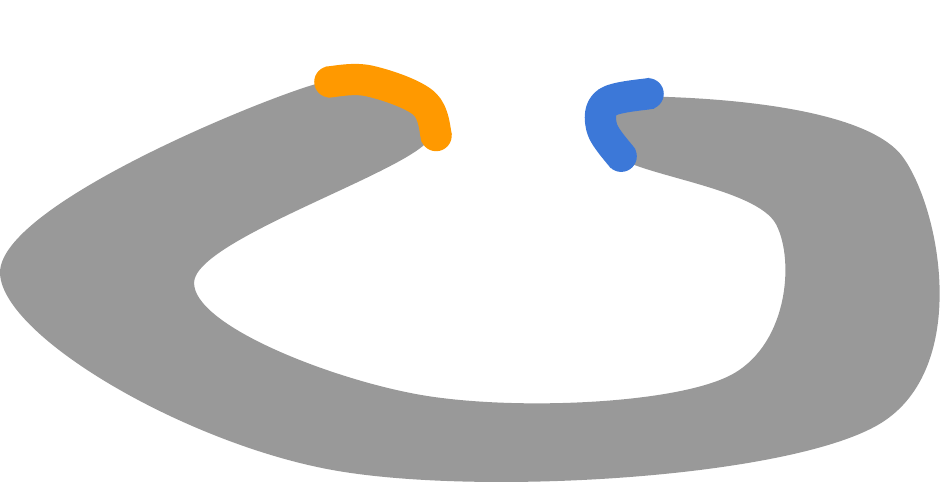}
    	\caption{Template}
        \label{fig:deformation_discontinuity_template}
	\end{subfigure}
	\begin{subfigure}[b]{0.45\columnwidth}
    	\includegraphics[width=\textwidth]{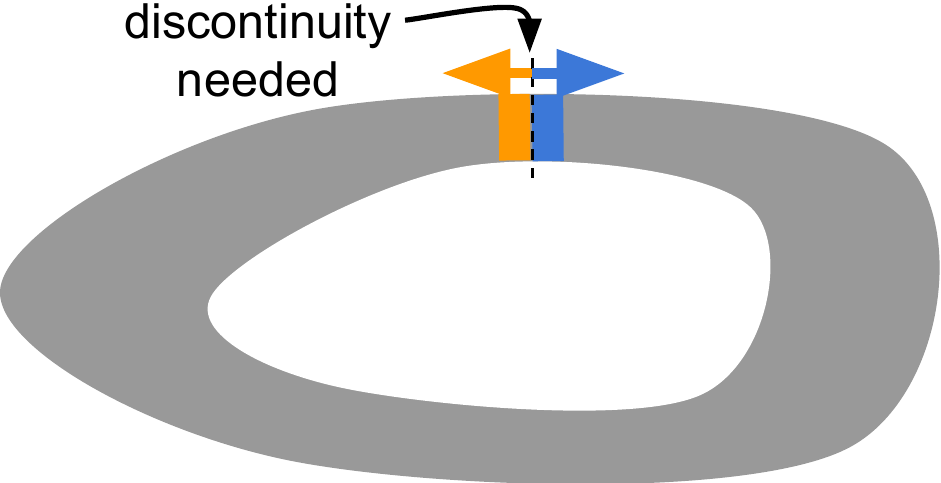}
    	\caption{Observation}
        \label{fig:deformation_discontinuity_observation}
	\end{subfigure}
	\vspace{-4pt}
    \caption{
    An example topological change where the ring opens at the marked seam. A deformation field referencing the template for each point in the observation frame would require a discontinuity at the seam, where an infinitesimal step from the orange position towards the blue position results in a big change in the deformation. If the template were the closed ring, the contents of the ring would be inaccessible using a deformation field.
    }
    \vspace{-10pt}
    \label{fig:deformation_discontinuity}
\end{figure}

\begin{figure*}[t!]
\centering
\includegraphics*[width=0.99\linewidth]{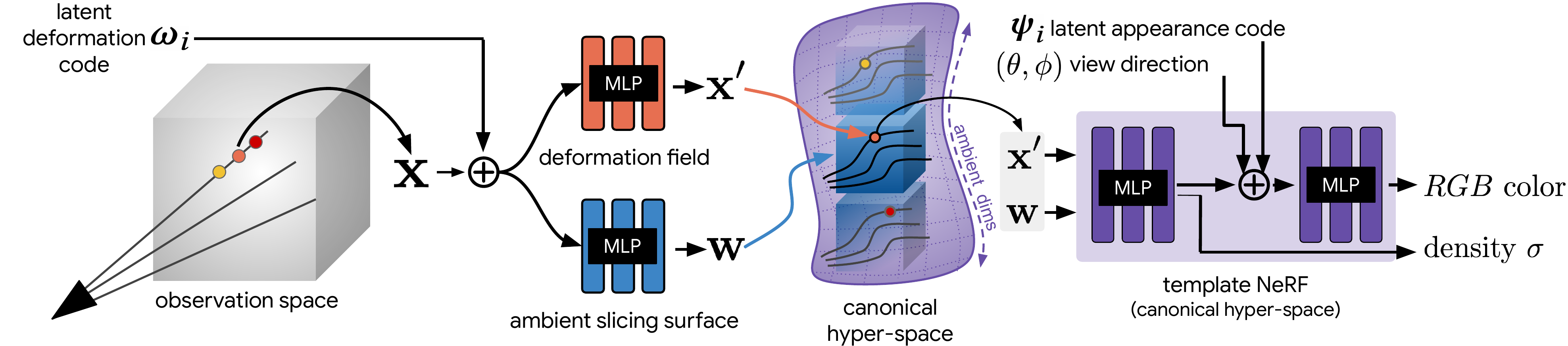}
\caption{
An overview of our model architecture. 
We associate a latent deformation code $\mat{\omega}_i$ and a latent appearance code $\mat{\psi}_i$ with each image $i$. 
Rays are cast from the camera in the space, and samples $\mat{x}$ along those rays are then concatenated with the image's latent deformation code $\mat{\omega}_i$ and provided as input to MLPs parameterizing a deformation field, which yields a warped coordinate $\mat{x}'$ and a slicing surface in hyperspace, which yields a coordinate in our ambient space $\mat{w}$.
Both outputs are concatenated to a mapped point $(\mat{x}',\mat{w})=(x,y,z,w_1,w_2,\ldots)$. The concatenation of  $(\mat{x}',\mat{w})$, the viewing direction $\mat{d}=(\theta,\phi)$, and the appearance code $\boldsymbol{\psi}_i$ are then used as inputs to the MLP parameterizing the template NeRF. The densities and colors produced by that MLP are then integrated along the ray as per the physics of volumetric rendering, as in \citet{mildenhall2020nerf}.
}
\label{fig:architecture}
\end{figure*}

\section{Method}
\label{sec:method}

Here we describe our method for modeling non-rigidly deforming scenes given a casually captured monocular image sequence. The focus of our method is to be able to accurately model the appearance and geometry of \emph{topologically varying} scenes. Before we introduce HyperNeRF, we first review its foundations: neural radiance fields (NeRF)~\cite{mildenhall2020nerf} as well as unconstrained and deformable extensions of it~\cite{martinbrualla2020nerfw, park2020nerfies}, all of which we will build upon.

\subsection{Review of NeRF, NeRF-W, and Deformable NeRF}
\label{sec:nerf}
NeRF represents a scene as a continuous, volumetric field of density and radiance, defined as $F: (\spatial,\mat{d}, \appearance) \rightarrow (\mat{c}, \sigma)$. The function $F$ is parameterized by a multilayer perception (MLP) and maps a 3D position $\spatial = (x,y,z)$ and viewing direction $\mat{d}=(\phi,\theta)$ to a color $\mat{c}=(r,g,b)$ and density $\sigma$. Instead of directly providing input coordinates to the MLP directly, NeRF first maps each input $\spatial$ (and $\viewdir$) using a sinusoidal positional encoding:
\begin{equation}
\resizebox{0.92\linewidth}{!}{%
$
\gamma(\spatial) = \left[\sin(\spatial),\cos(\spatial),\sin(2\spatial),\cos(2\spatial),\ldots,\sin(2^{m-1}\spatial),\cos(2^{m-1}\spatial)\right]^\mathrm{T},
$}
\end{equation}
where $m$ is a hyper-parameter that controls the number of sinusoids used by the encoding. As shown in \citet{tancik2020fourfeat}, this encoding allows the MLP to model high-frequency signals in low frequency domains, where the parameter $m$ serves to control smoothness of the learned representation by modifying the effective bandwidth of an interpolating kernel. 

\paragraph{Appearance Variation} To handle unconstrained ``in the wild'' images, the MLP in NeRF can be additionally conditioned on an appearance embedding $\appearance$ for each observed frame $i\in\{1,\ldots,n\}$, as shown in \citet{martinbrualla2020nerfw}. This allows NeRF to handle appearance variations between input frames, such as those caused by illumination variation or changes in exposure and white balance.

\paragraph{Deformations} Because a NeRF is only able to represent static scenes, we use the deformation field formulation proposed in Nerfies~\cite{park2020nerfies} to model non-rigid motion. Nerfies defines a mapping $T : (\spatial,\mat{\omega}_i)\rightarrow \spatial'$ that maps all observation-space coordinates $\spatial$ to canonical-space coordinates $\spatial'$, conditioned on a per-observation latent deformation code $\mat{\omega}_i$. The deformation field $T$ is parameterized by an MLP $W : (\spatial,\mat{\omega}_i) \rightarrow (\mat{r},\mat{v})$, where $(\mat{r},\mat{v})\in\sethree$ encode the rotation and translation. To encourage deformations to be as-rigid-as-possible, Nerfies proposes an elastic regularization loss which penalizes non-unit singular values of the Jacobian of the deformation field. 
We found that the elastic loss is ill-suited for some of our scenes due to the topological variations which directly violate its assumptions.
Please see \citet{park2020nerfies} for details.

\paragraph{Windowed Positional Encoding}

\citet{tancik2020fourfeat} showed that the number of frequencies $m$ in the positional encoding $\gamma$ controls the bandwidth of an MLP's Neural Tangent Kernel (NTK)~\cite{jacot2018neural}. A small value for $m$ results in a smooth estimator that may under-fit the data, while a large value of $m$ may result in over-fitting. \citet{park2020nerfies} and \citet{hertz2021progressive} use this property to implement a coarse-to-fine strategy when optimizing an MLP, by slowly narrowing the bandwidth of the NTK by weighting the frequency bands of the positional encoding with a window function. For example, \citet{park2020nerfies} uses the window function
\begin{align}
    w_j(\alpha) = \frac{1-\cos(\pi\operatorname{clamp}(\alpha-j,0,1))}{2}\,,
\end{align}
where $j\in\{0, \ldots, m-1\}$ is the index of the frequency band of the positional encoding. Linearly increasing $\alpha\in[0,m]$ is equivalent to sliding a truncated Hann window down the frequency bands of positional encoded features. The windowed positional encoding is then computed as:
\begin{equation}
\gamma_\alpha(\spatial) = \left[w_0(\alpha)\sin(\spatial),\ldots,w_{m-1}(\alpha)\cos(2^{m-1}\spatial)\right]^\mathrm{T}.
\label{eq:windowed_posenc}
\end{equation}
Nerfies uses this to optimize the deformation field in a coarse-to-fine manner which prevents getting stuck in sub-optimal local minima while being retaining the ability to represent high-frequency deformations. We do the same for our spatial deformation field.

\subsection{Hyper-Space Neural Radiance Fields}
\label{sec:hypernerf}

\begin{figure*}[t]
	\centering
	\begin{subfigure}{0.43\textwidth}
    	\centering
    	\includegraphics[height=40mm]{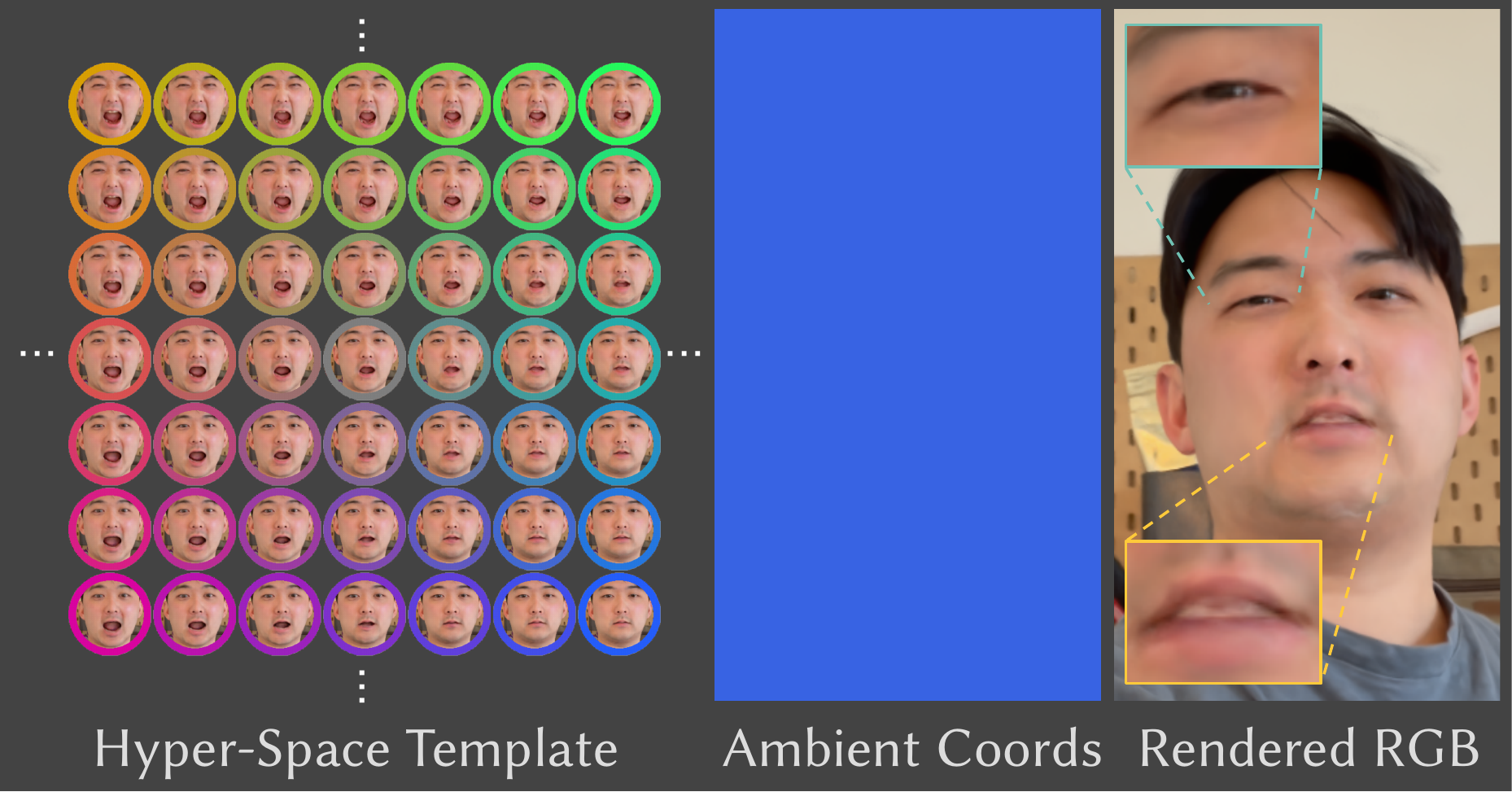}
    	\caption{Axis-Aligned Slicing Plane (AP)}
        \label{fig:ap_ds_comp_ap}
	\end{subfigure}
	\begin{subfigure}{0.12\textwidth}
    	\centering
    	\includegraphics[height=40mm,clip,trim=0 0 391pt 0]{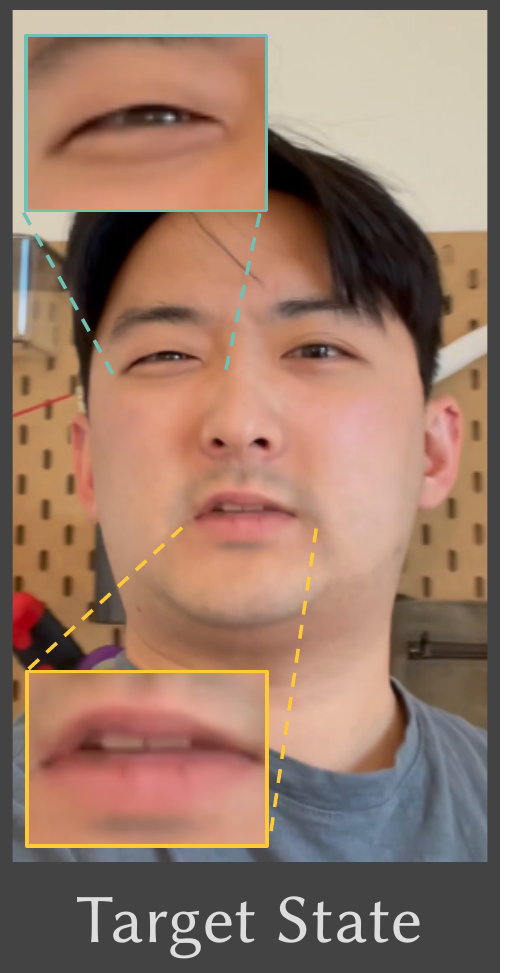}
    	\caption{}
        \label{fig:ap_ds_comp_target}
	\end{subfigure}
	\begin{subfigure}{0.43\textwidth}
    	\centering
    	\includegraphics[height=40mm]{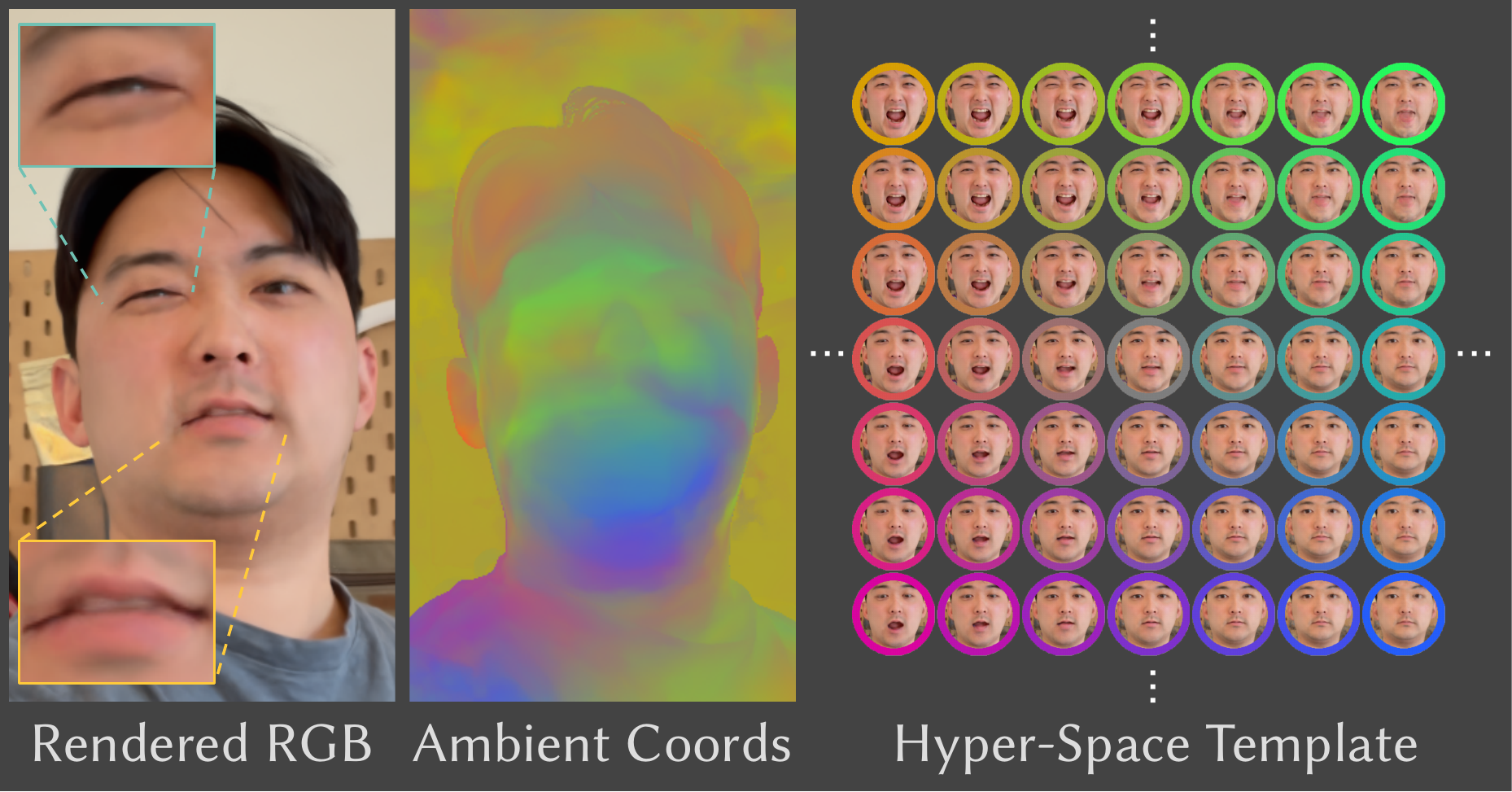}
    	\caption{Deformable Slicing Surface (DS)}
        \label{fig:ap_ds_comp_ds}
	\end{subfigure}

    \caption{We render the target state (b) from a novel view using axis-aligned slicing planes (AP, a) and deformable surfaces (DS, c). We show the hyper-space template rendered at ambient coordinates, sampled on a regular grid. We plot the ambient coordinate of each pixel --- the color of the coordinates correspond to the outlined color of each template sample. AP must wholly model each scene state and results in artifacts in the eyes, mouth, and chin. DS can more efficiently use the template since each part of the scene can refer to different parts of the template, resulting in sharper details.}
    \label{fig:ap_ds_comp}
    \vspace{-5pt}
\end{figure*}

Motion in a scene can be divided into two categories: (a) motions that preserve the topology of the scene; and (b) motions which change the apparent topology of the scene. Here we use the term ``topology'' in the sense we defined in \secref{sec:introduction}.
Deformation fields can effectively model topology-preserving motion, but cannot easily model changes in topology. This is because a change in topology necessarily involves a discontinuity in the deformation field. Our deformation fields are continuous, by virtue of being encoded within the weights of an MLP that behaves as a smooth interpolator, and therefore cannot represent such discontinuities. See \figref{fig:deformation_discontinuity} for a visual explanation. 

Here we present our method, \emph{HyperNeRF}, which extends neural radiance fields into higher dimension to allow topological variations.
In \secref{sec:level_set_methods} we introduced the level set method, which can naturally model topologically changing shapes as cross-sections of a higher-dimensional ambient surface. We use the same idea in the context of neural radiance fields. Our architecture is visualized in \figref{fig:architecture}.

\paragraph{Hyper-space Template} Deformable NeRFs~\cite{park2020nerfies} represent a scene in a canonical-space \emph{template} NeRF which is indexed by a spatial deformation field to render observation frames. Our idea is to embed the template NeRF in higher dimensions, where a slice taken by an intersecting high-dimensional slicing surface yields a full 3D NeRF, akin to slicing a 3D object to get a 2D shape as in \figref{fig:level_set_example}. Note that we are still rendering 3-dimensional scenes, and thus ray casting and volume rendering occur in the same manner as a normal NeRF.

We extend the domain of the template NeRF to a higher dimensional space: $(\spatial,\hyper)\in\mathbb{R}^{3+W}$, where $W$ is the number of higher dimensions (visualized in \figref{fig:ap_ds_comp_ap}). The formulation of the template NeRF is otherwise similar, with it being represented as an MLP
\begin{align}
    F: \left(\spatial,\hyper,\mat{d},\appearance\right) \rightarrow (\mat{c},\sigma)\,.
\end{align}

\paragraph{Slicing Surfaces} In \secref{sec:slicing_surfaces} and \figref{fig:level_set_slice} we showed how a 3D auxiliary surface could be cut with a slicing surface to obtain a 2D shape. In the same way, we can take cross-sections of a hyper-space NeRF. While our 2D examples only had a single ambient dimension for visualization purposes, we can have more. Slicing an auxiliary shape with a single ambient dimension involves a single slicing plane; slicing an auxiliary shape with more than one ambient dimension simply requires a slicing plane for each dimension. 
This is the geometric interpretation of evaluating the auxiliary function at a specific value of $\hyper$, which leaves the span of the spatial axes as the cross-section along the ambient dimensions.

Analogous to the 2D \emph{axis-aligned slicing plane (AP)} formulation described in \secref{sec:slicing_surfaces}, we can associate a specific slice of the template for each observation $i$ by directly optimizing the ambient coordinates $\{\hyper_i\}$ as with the deformation and appearance codes.
This approach forces all ray samples for an observation to share the same ambient coordinates, requiring all observed states to be explicitly modeled at a single ambient coordinate. This can lead to an inefficient use of the representation capacity of the template NeRF, since topological variations happening in spatially distance parts of the scene must be copied across multiple sub-spaces to represent different combinations of states (as with \figref{fig:level_set_slice_ap}). This can decrease reconstruction quality (\figref{fig:ap_ds_comp_ap}) and causes fading artifacts when interpolating between different states (\figref{fig:interp_ap_vs_bs}). 

We therefore use \emph{deformable slicing surfaces (DS)} --- introduced in \secref{sec:slicing_surfaces} --- which allow different spatial positions to be mapped to different coordinates along the ambient dimensions. This allows a more efficient use of the ambient dimensions (as in \figref{fig:level_set_slice_bs}), resulting in better reconstruction quality (\figref{fig:ap_ds_comp_ds}) and smoother interpolations between states compared to axis-aligned planes (\figref{fig:interp_ap_vs_bs}).

Similarly to the spatial deformation field, we define a deformable slicing surface field using an MLP. Each observation-space sample point $\spatial$ is mapped through the mapping $H : (\spatial,\mat{\omega}_i) \rightarrow \hyper$, where $\mat{\omega}_i$ is the latent deformation code (shared with the spatial deformation field), and $\hyper$ is a point in the ambient coordinate space which defines the cross-sectional subspace for the sample.  
Given the template NeRF $F$, the spatial deformation field $T$, and the slicing surface field $H$, the observation-space radiance field can be evaluated as:
\begin{align}
    \spatial' &= T(\spatial,\latent)\,, \\
    \hyper &= H(\spatial,\latent)\,, \\
    (\mat{c},\sigma) &= F(\spatial',\hyper,\viewdir,\appearance)\,.
\end{align}
where in practice we apply separate positional encodings to $\spatial$, $\viewdir$, and $\hyper$. We use a windowed positional encoding $\gamma_{\alpha}$ for the deformation field $T$ and $\gamma_{\beta}$ for the slicing surface field $H$, where $\alpha$ and $\beta$ are parameters for the windowed positional encoding defined in \equref{eq:windowed_posenc}.

We do not use the identity concatenation of the positional encoding for the ambient coordinates. 
\citet{barron2021mipnerf} showed that the identity encoding does not meaningfully affect performance or speed, and omitting it confers a specific advantage: we can collapse ambient dimensions during the initial phases of the optimization by setting $\beta = 0$, as described in the previous section.

\paragraph{Delayed use of Ambient Dimensions} Motion can be encoded by deforming points using the spatial deformation field $T$, or by moving along the ambient dimensions using the slicing surface $H$. If a motion does not involve a change in topology, we prefer using a spatial deformation over moving through the ambient dimensions. This is because spatial deformations result in a more constrained optimization, since several observations get mapped to the same point on the template. In addition, spatial deformations result in better interpolations and they only involve movement and cannot directly change the density of the template. We therefore delay the use of the ambient dimensions by using the windowed positional encoding (\secref{sec:nerf}) on the ambient coordinates $\hyper$. We disable the identity concatenation for the positional encoding so that the input can be completely turned off when $\beta=0$. We fix $\beta=0$ for a number of iterations and then linearly ease in the rest of the frequencies. This windowing is only applied to the ambient dimensions. The spatial coordinate $\spatial$ is not windowed, and the spatial deformation field uses its own windowing schedule from \citet{park2020nerfies}.

\fboxsep=0pt 
\fboxrule=1.5pt 

\definecolor{startcolor}{rgb}{0.067, 0.541, 0.698}
\definecolor{endcolor}{rgb}{0.937, 0.278, 0.435}

\newcommand{\kfbox}[1]{\fcolorbox{black}{white}{#1}}
\newcommand{\kfboxstart}[1]{\fcolorbox{startcolor}{white}{#1}}
\newcommand{\kfboxend}[1]{\fcolorbox{endcolor}{white}{#1}}
\newcommand{\latcell}[1]{\includegraphics[width=1.2cm,clip,trim=40 140 50 70]{#1}}
\newcommand{\latcellkfstart}[1]{\kfboxstart{\includegraphics[width=1.2cm,clip,trim=40 140 50 70]{#1}}}
\newcommand{\latcellkfend}[1]{\kfboxend{\includegraphics[width=1.2cm,clip,trim=40 140 50 70]{#1}}}

\begin{figure}[t]
    \setlength{\tabcolsep}{0.5pt}
    \renewcommand{\arraystretch}{0.25}
    \centering
    \begin{tabular}{ccccccc}
    
        \multirow{4}*[0.7cm]{\kfboxstart{\includegraphics[width=0.93cm,clip,trim=40 0 40 0]{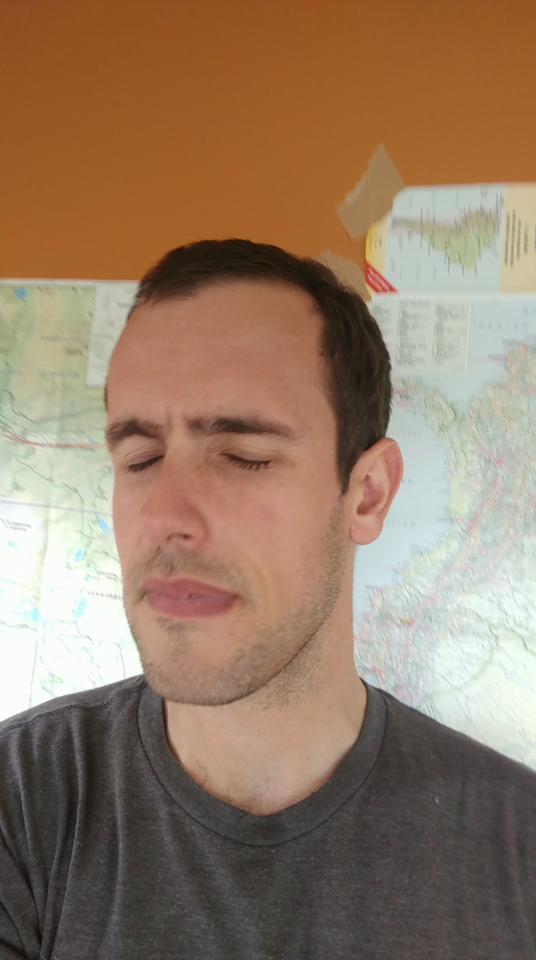}}\hspace{0.05cm}} &
        
        \latcellkfstart{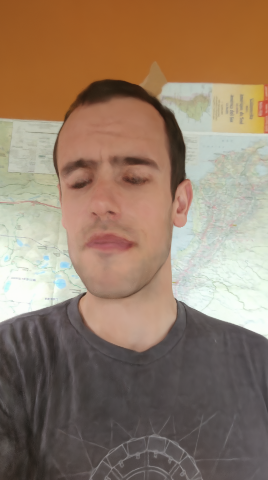} &
        \latcell{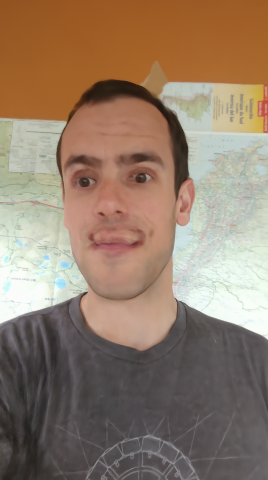} &
        \latcell{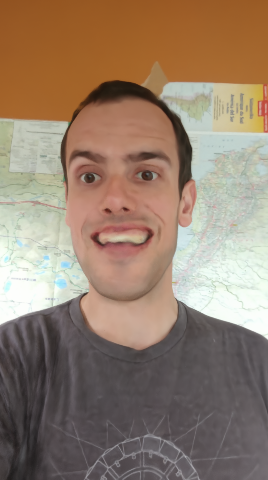} &
        \latcell{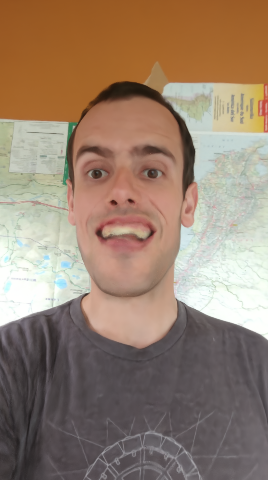} &
        \latcellkfend{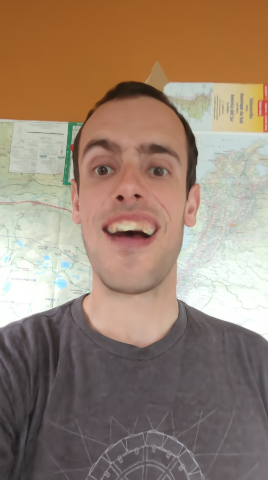} &
        
        \multirow{4}*[0.7cm]{\hspace{0.05cm}\kfboxend{\includegraphics[width=0.93cm,clip,trim=40 0 40 0]{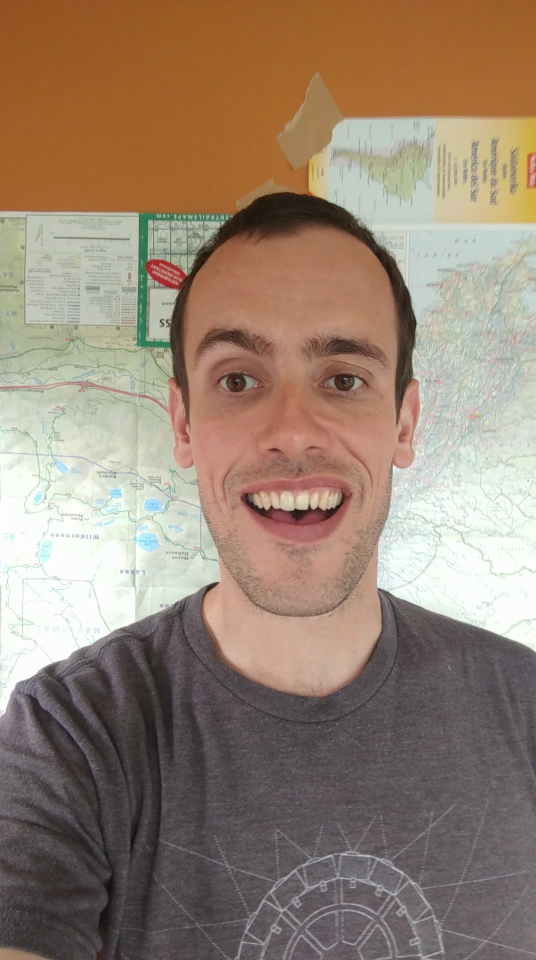}}}
        \vspace{2pt}
        \\
        & \multicolumn{5}{c}{{\small Axis-Aligned Slicing Plane (AP)}} & 
        \\
       
        {\small start}
        &
        \latcellkfstart{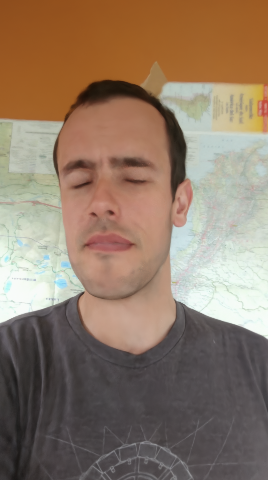} &
        \latcell{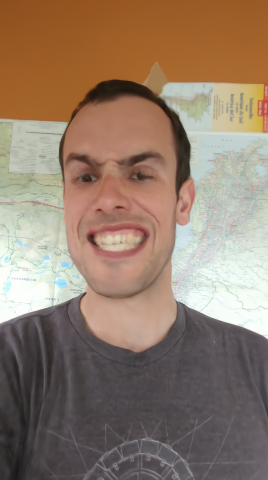} &
        \latcell{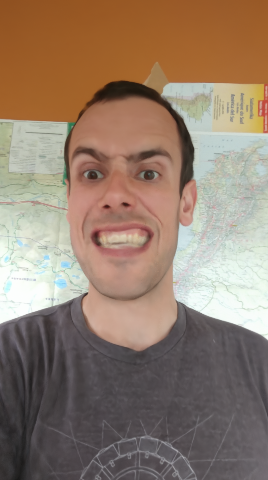} &
        \latcell{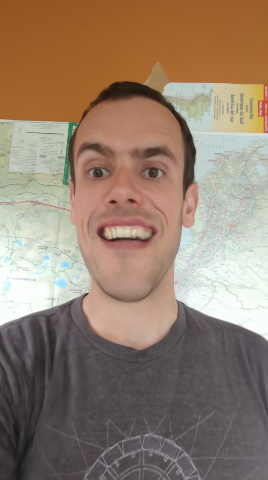} &
        \latcellkfend{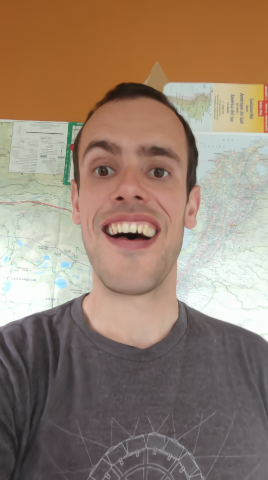} &
        
        {\small end}
        \vspace{2pt}
        \\
        & \multicolumn{5}{c}{{\small Deformable Slicing Surface (DS)}} & 
    \end{tabular}
    \vspace{-8pt}
	\caption{
	    Novel views synthesized by linearly interpolating the embedding $\latent$ of two frames from ``\textsc{Expressions 1}''. With (AP), all spatial points are rendered at the same ambient coordinate resultin in blurry-cross-fading artifacts during interpolation. With (DS), different spatial positions to reference varying parts of the ambient space, resulting in better interpolations.
	}
  \label{fig:interp_ap_vs_bs}
\end{figure}

\fboxsep=0pt 
\fboxrule=2.0pt 

\definecolor{bestcolor}{rgb}{0.85, 0.113, 0.188}
\definecolor{topcolor}{rgb}{0.937, 0.278, 0.435}
\definecolor{botcolor}{rgb}{0.067, 0.541, 0.698}

\newcommand{\topbox}[1]{\fcolorbox{topcolor}{white}{#1}}
\newcommand{\botbox}[1]{\fcolorbox{botcolor}{white}{#1}}

\newcommand{\resultcell}[2]{\fcolorbox{white}{white}{\includegraphics[width=1.6cm,unit=1mm,clip,trim=#1]{images/results/#2}}}
\newcommand{\topcell}[2]{\topbox{\includegraphics[width=1.6cm,unit=1mm,clip,trim=#1]{images/results/#2}}}
\newcommand{\botcell}[2]{\botbox{\includegraphics[width=1.6cm,unit=1mm,clip,trim=#1]{images/results/#2}}}
\newcommand{\gtcell}[2]{\multirow{2}{*}[0.4in]{\resultcell{#1}{#2}}}

\newcommand{\animage}[2]{\includegraphics[width=2.0cm,clip,trim=#1]{figures/vrig_results_v2/#2}}
\newcommand{\bestnum}[1]{\textcolor{bestcolor}{#1}}

\begin{figure*}[t!]
\resizebox{0.99\linewidth}{!}{
    \setlength{\tabcolsep}{0.4pt}
    \renewcommand{\arraystretch}{0.25}
    \begin{tabular}{@{}c@{\,}c@{\,}c@{}c@{}c@{}c@{\,\,}c@{\,}c@{}c@{}c@{}c@{}}
            
        \multirow{2}{*}[0.25in]{\makebox[20pt]{\raisebox{25pt}{\rotatebox[origin=c]{90}{\textsc{Tamping}}}} \hspace{-7pt}} &
        \gtcell{0 220 0 100}{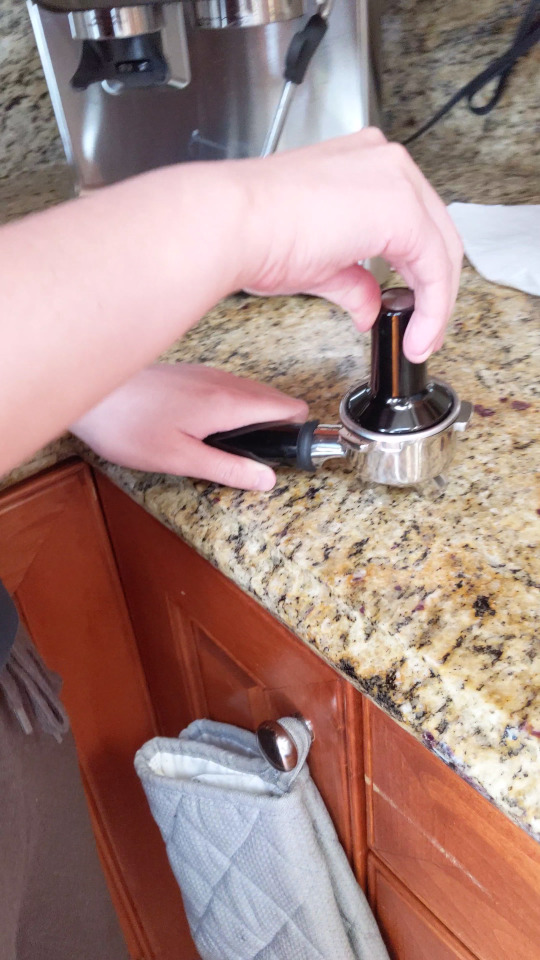} &
        \topcell{0 220 0 100}{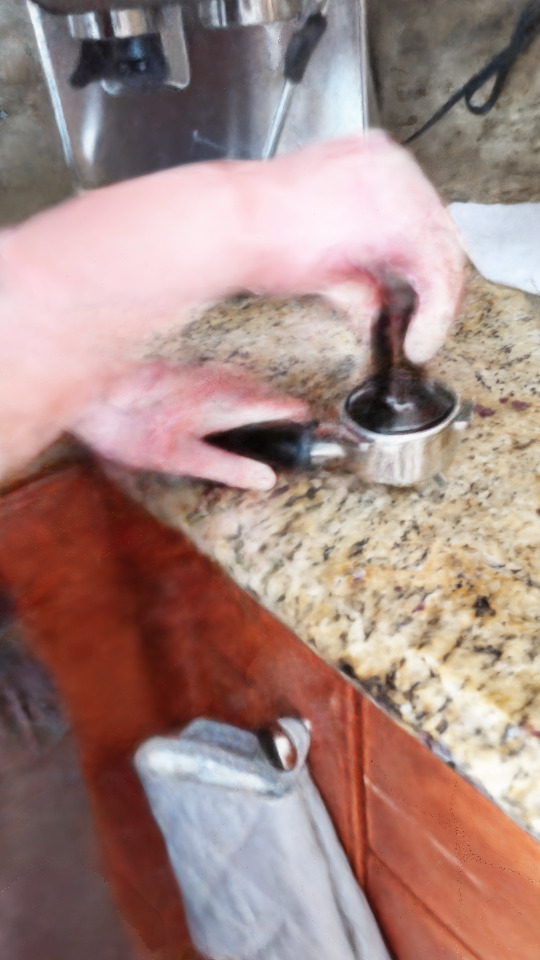} &
        \topcell{0 220 0 100}{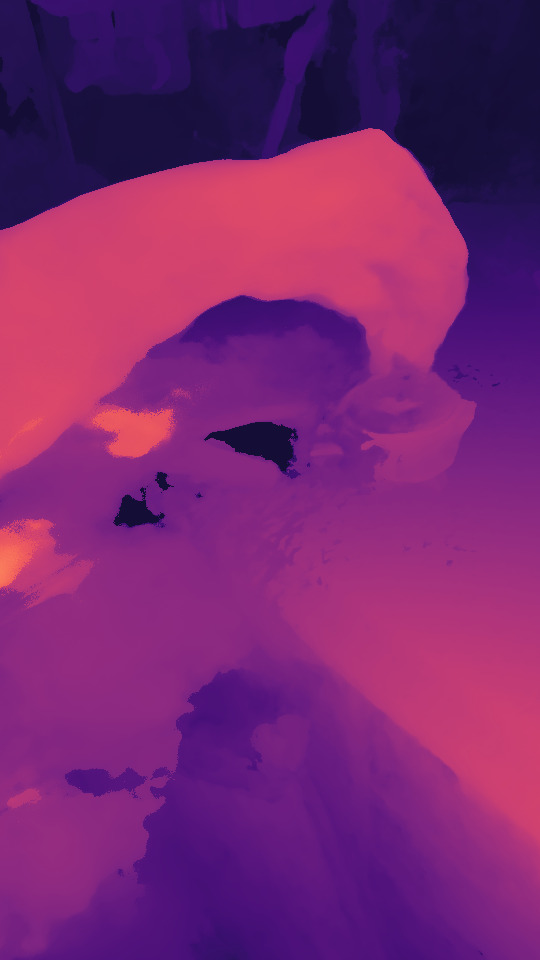} &
        \topcell{0 220 0 100}{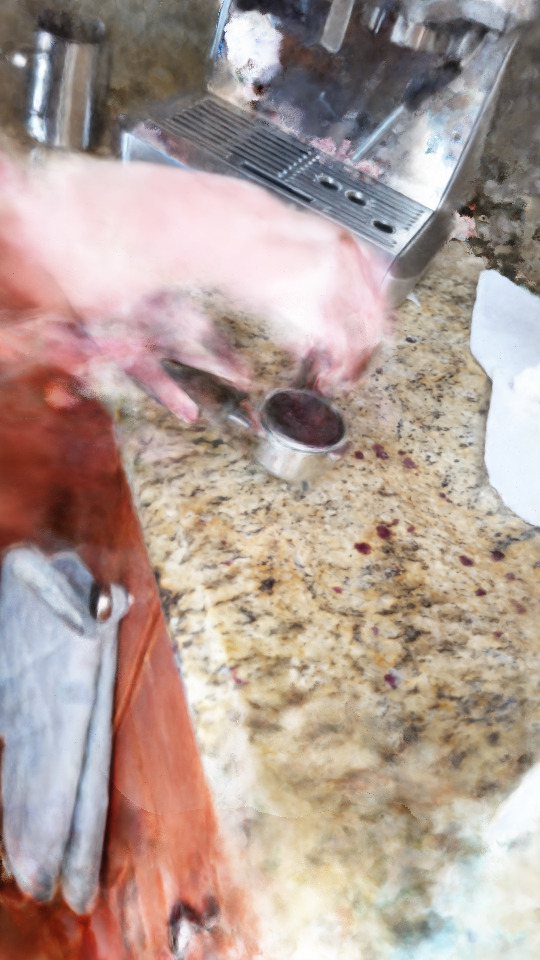} &
        \topcell{0 220 0 100}{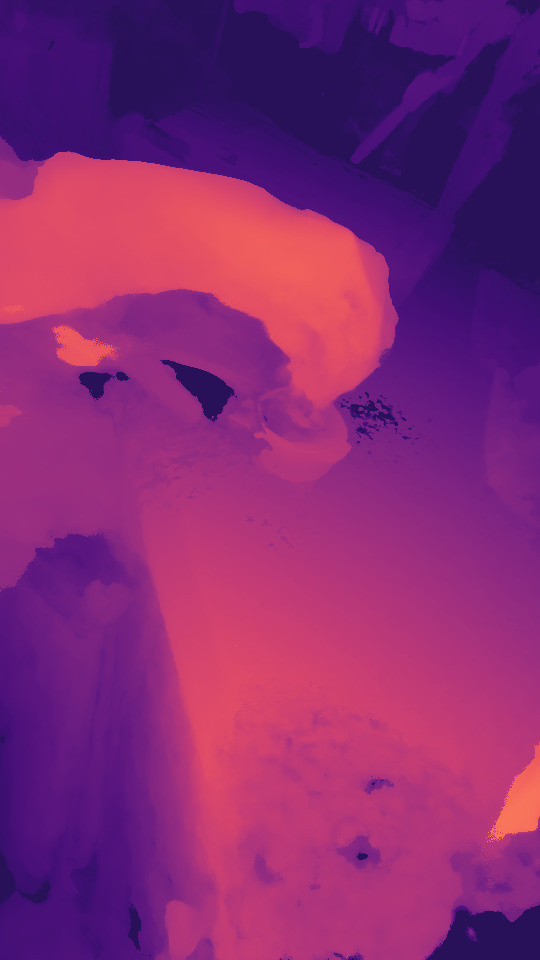} &
        \gtcell{0 220 0 100}{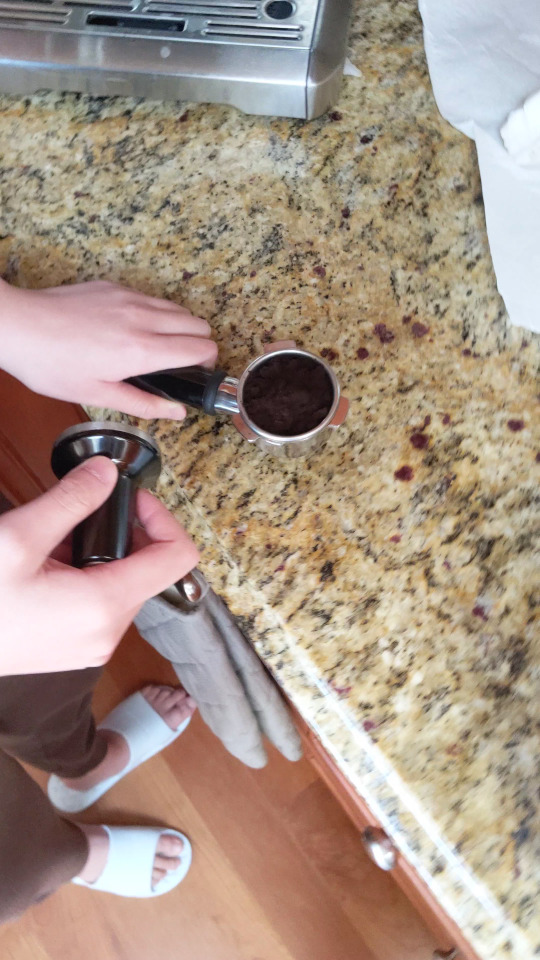} &
        \topcell{0 220 0 100}{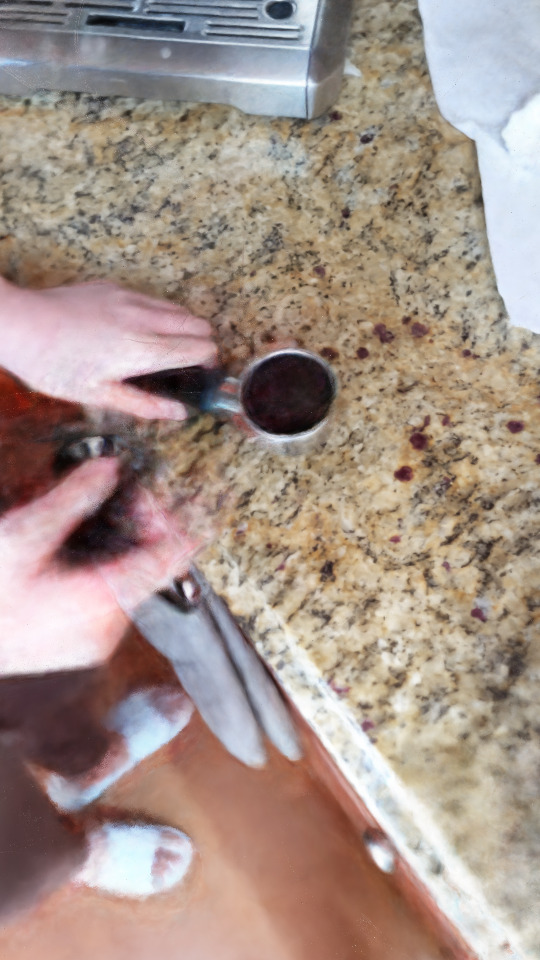} &
        \topcell{0 220 0 100}{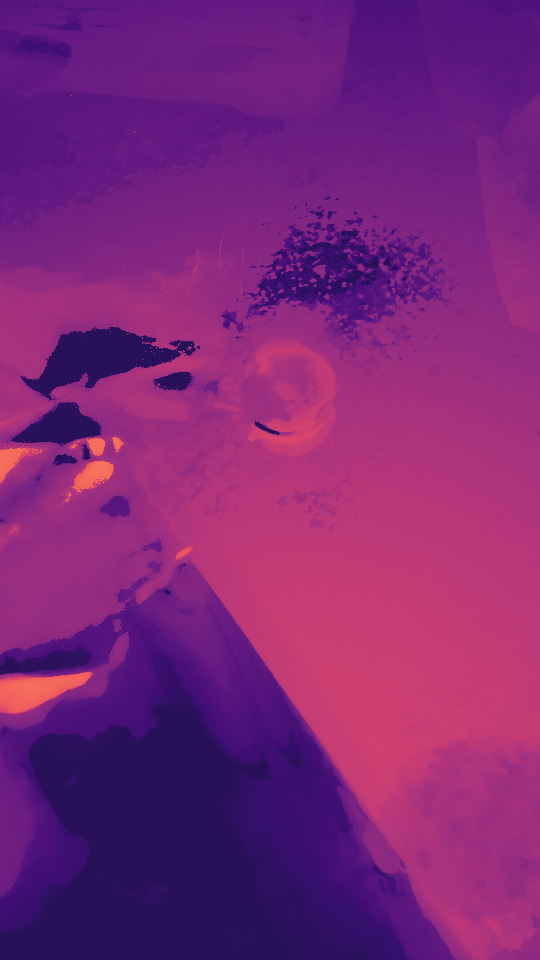} &
        \topcell{0 220 0 100}{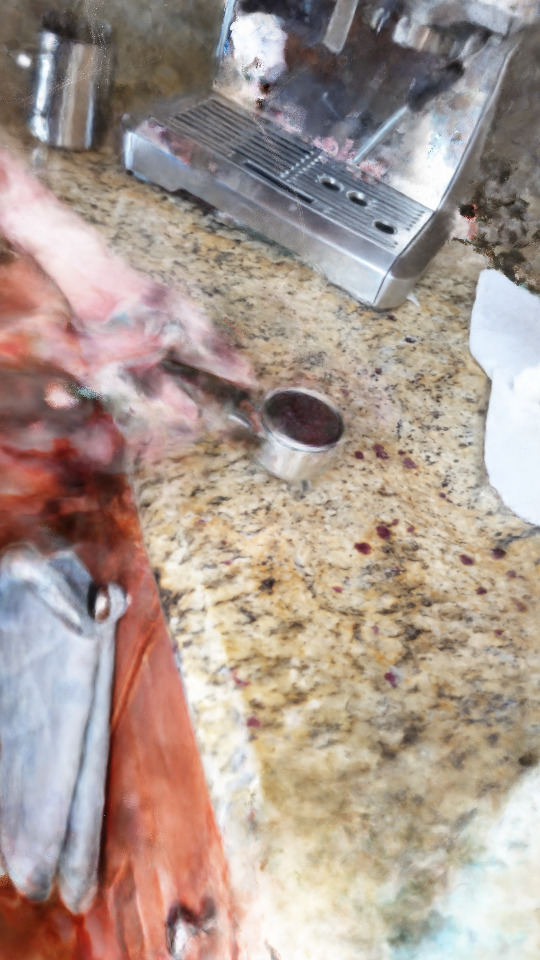}  &
        \topcell{0 220 0 100}{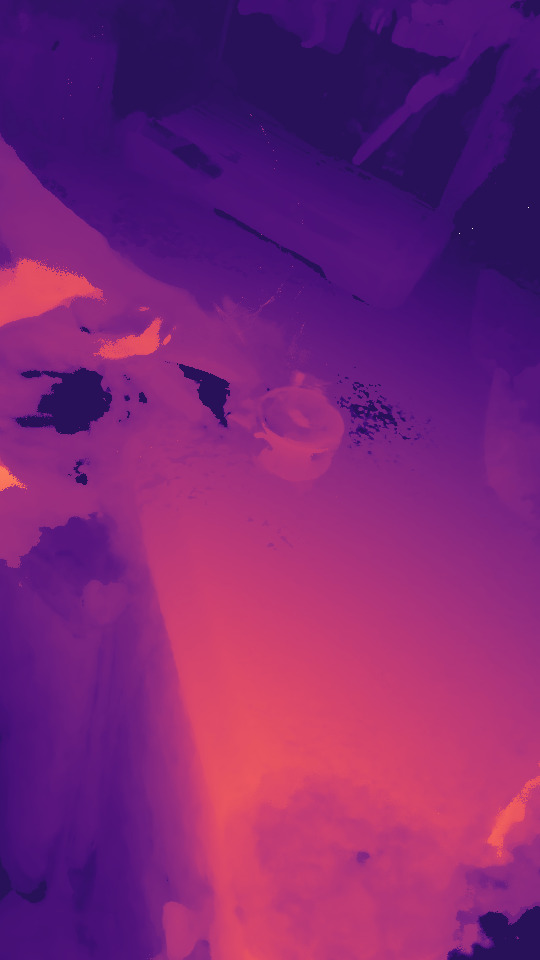}  \\
        &
        &
        \botcell{0 220 0 100}{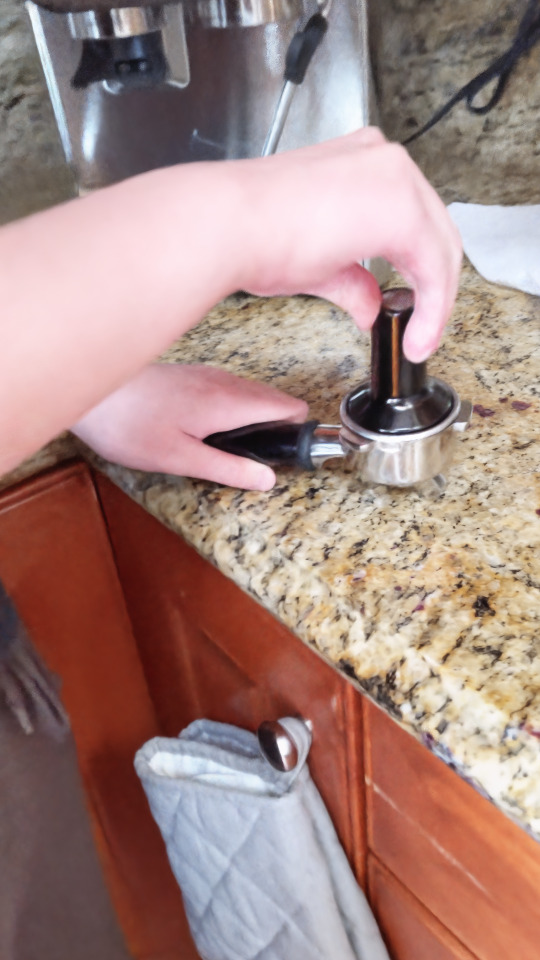} &
        \botcell{0 220 0 100}{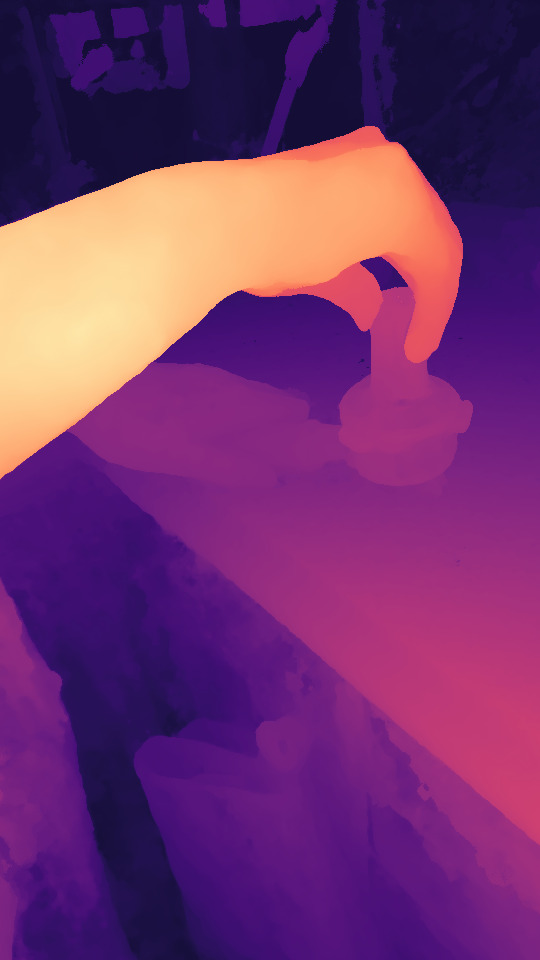} &
        \botcell{0 220 0 100}{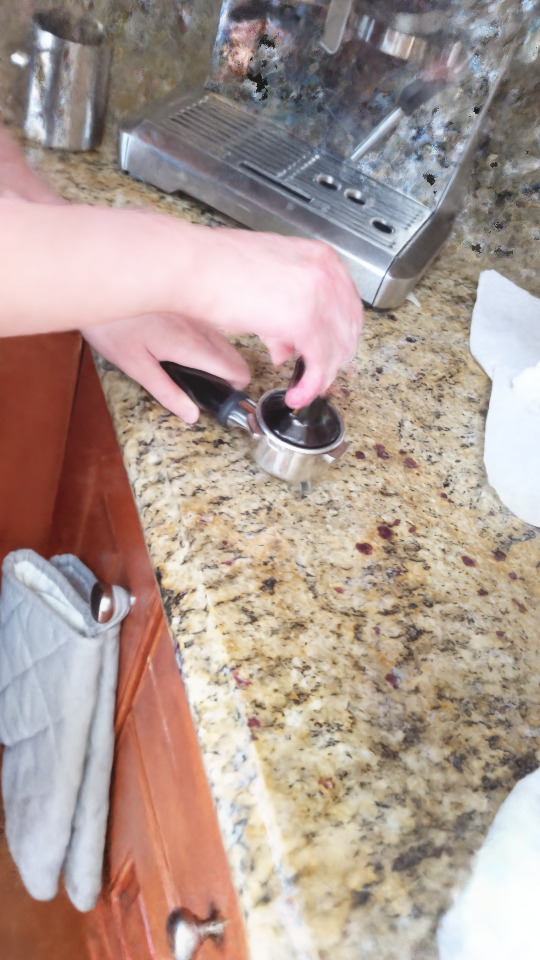} &
        \botcell{0 220 0 100}{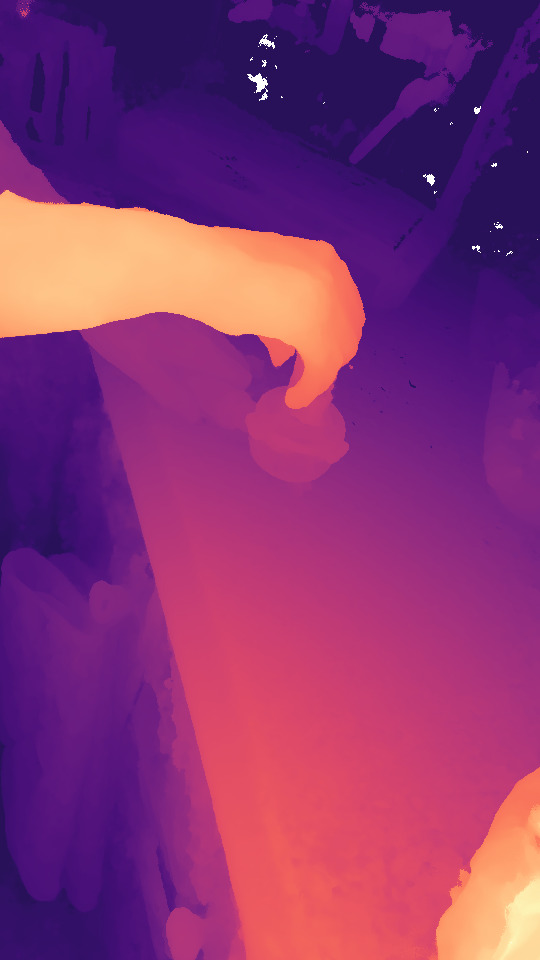} &
        &
        \botcell{0 220 0 100}{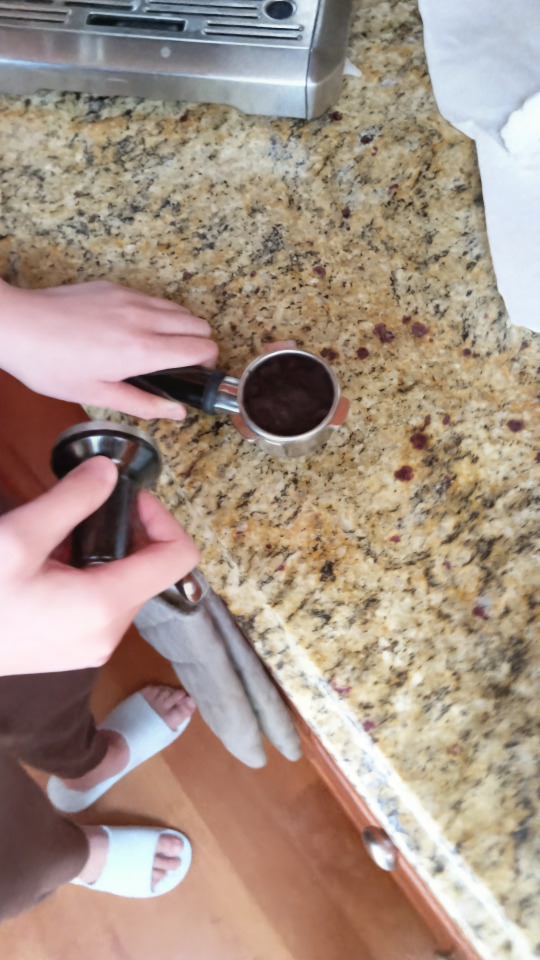} &
        \botcell{0 220 0 100}{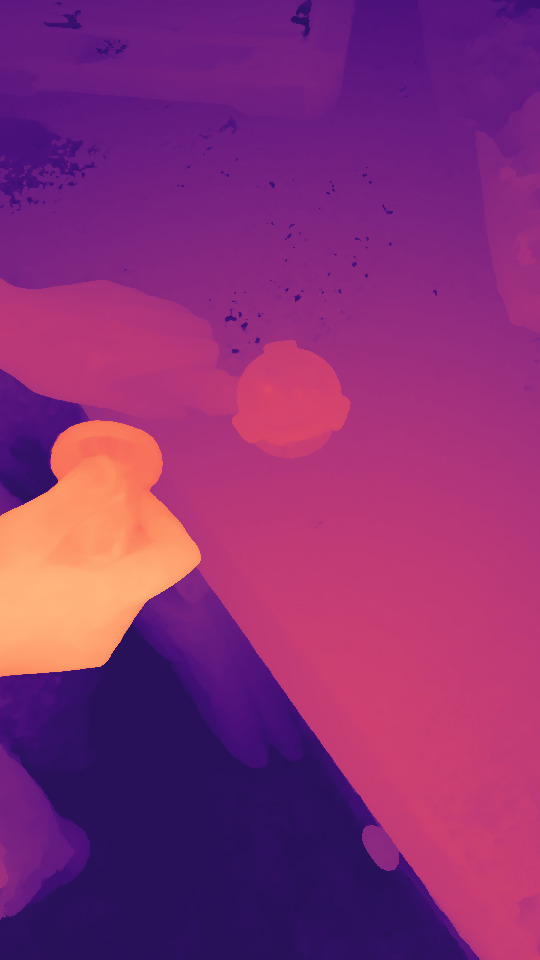} &
        \botcell{0 220 0 100}{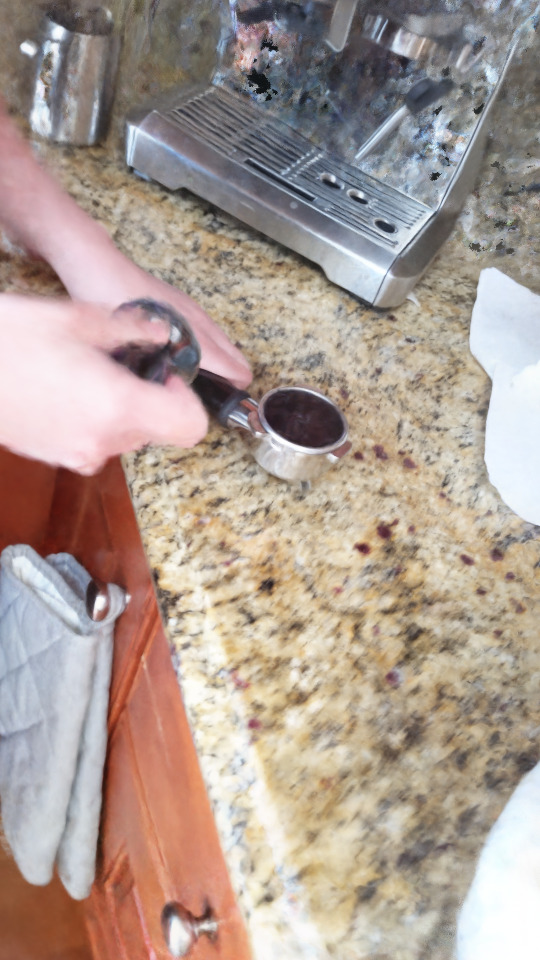} &
        \botcell{0 220 0 100}{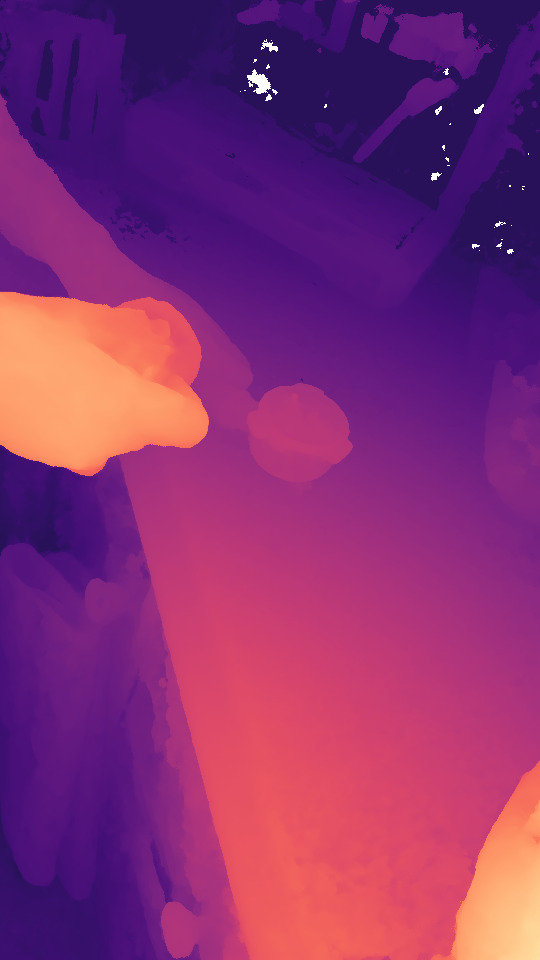} \\
        \\\\\\
        \multirow{2}{*}[0.34in]{\makebox[20pt]{\raisebox{25pt}{\rotatebox[origin=c]{90}{\textsc{Banana Slicer}}}} \hspace{-7pt}} &
        \gtcell{125 250 50 250}{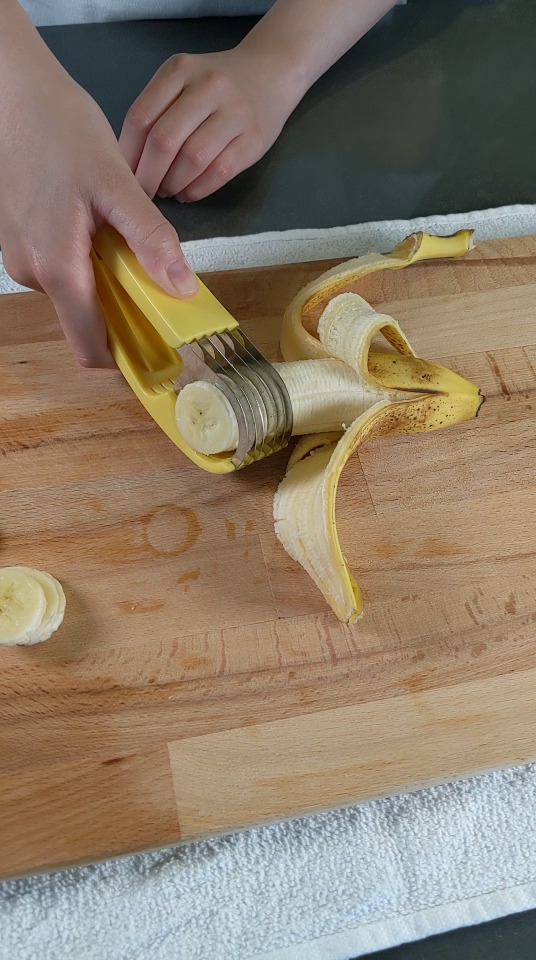} &
        \topcell{50 300 125 200}{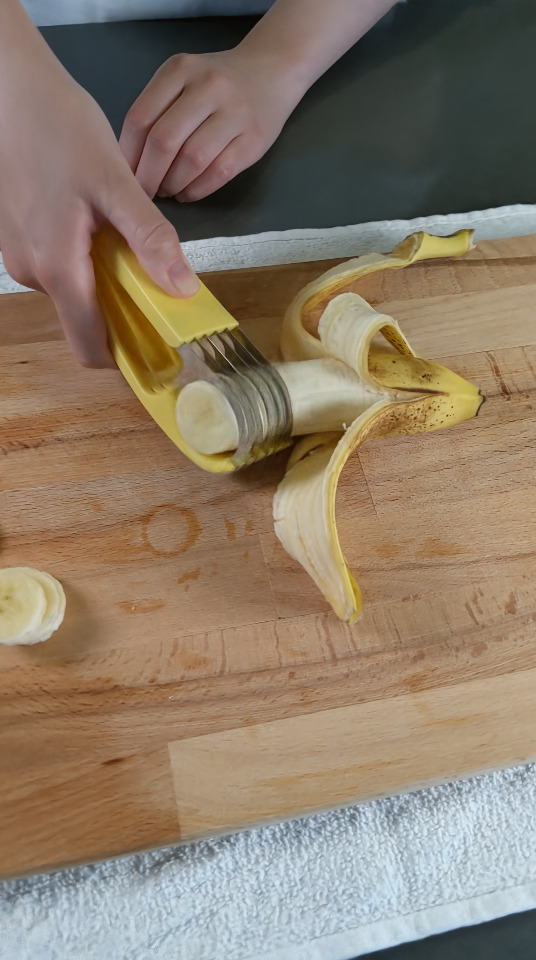} &
        \topcell{50 300 125 200}{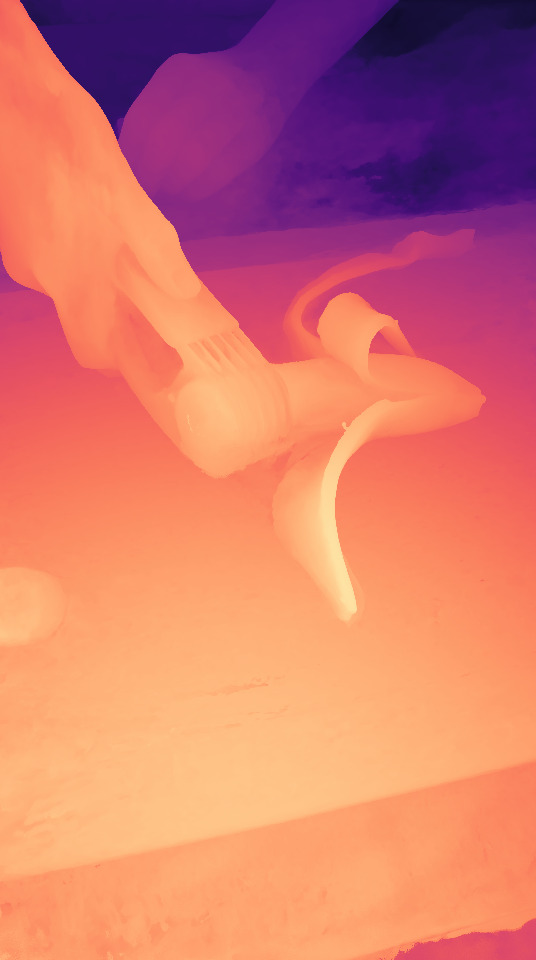} &
        \topcell{50 300 125 200}{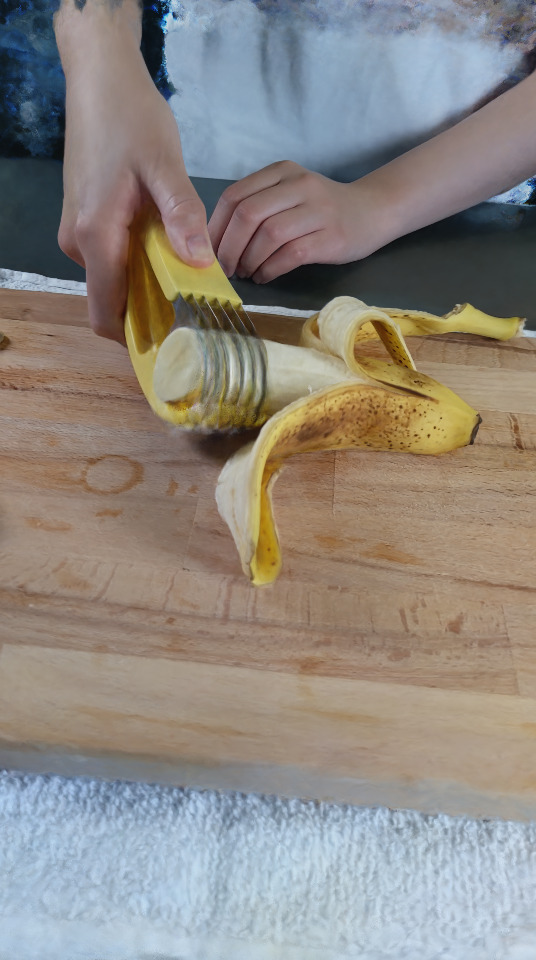} &
        \topcell{50 300 125 200}{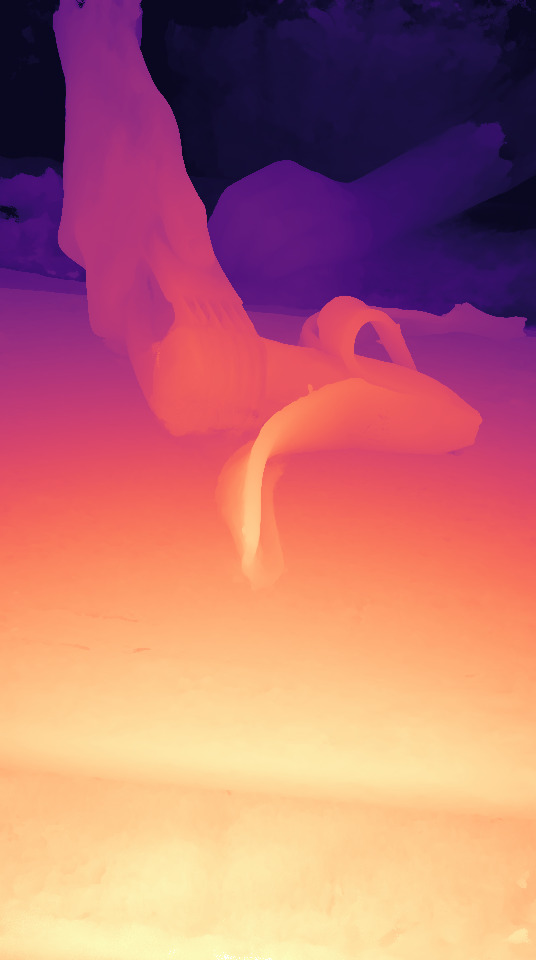} &
        \gtcell{50 250 125 250}{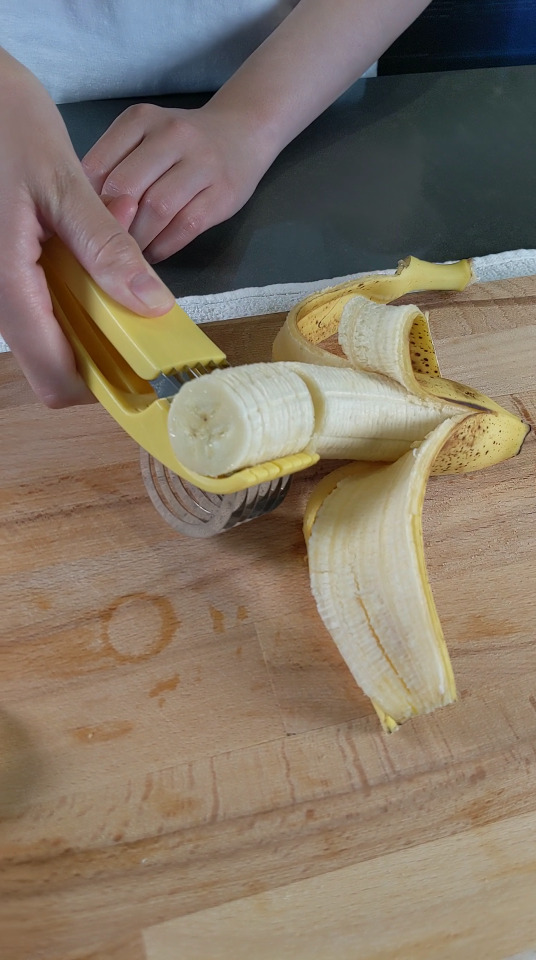} &
        \topcell{50 250 125 250}{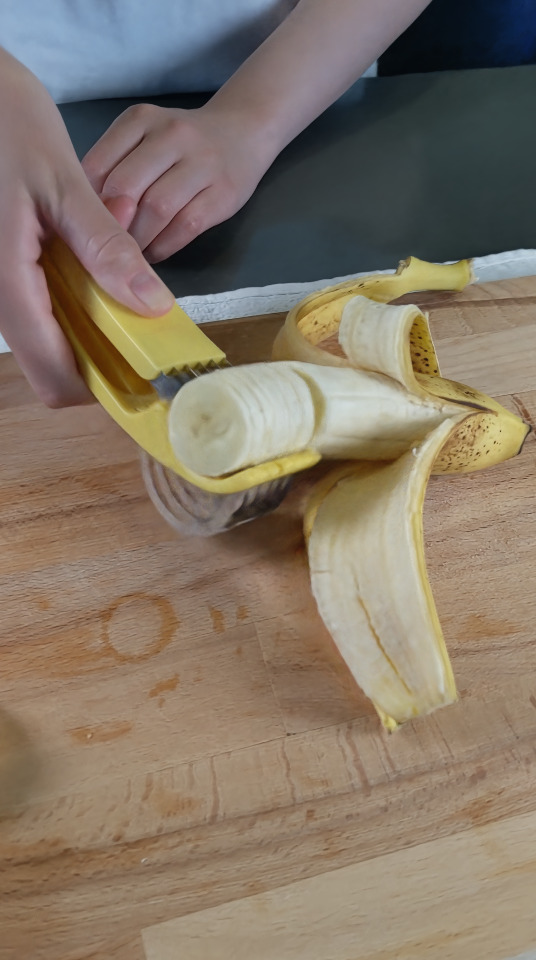} &
        \topcell{50 250 125 250}{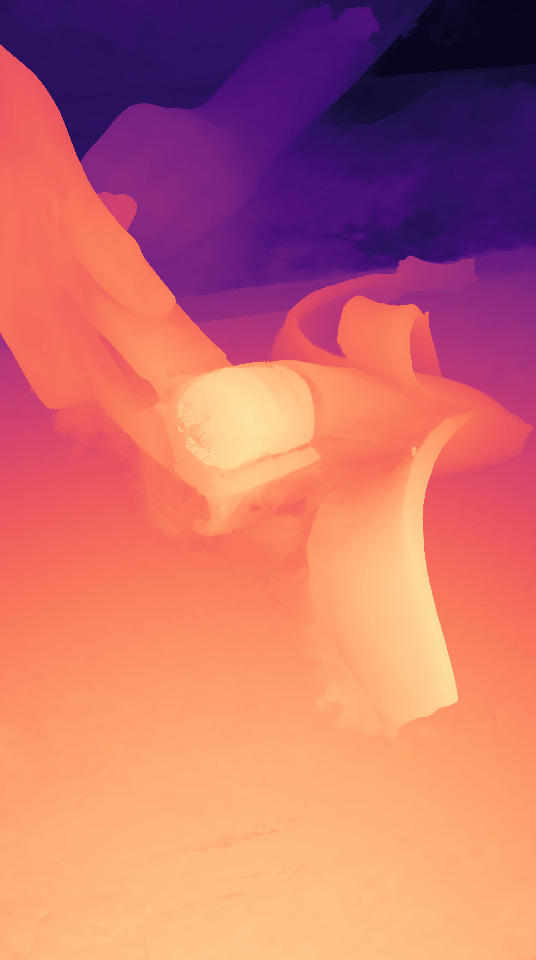} &
        \topcell{50 250 125 250}{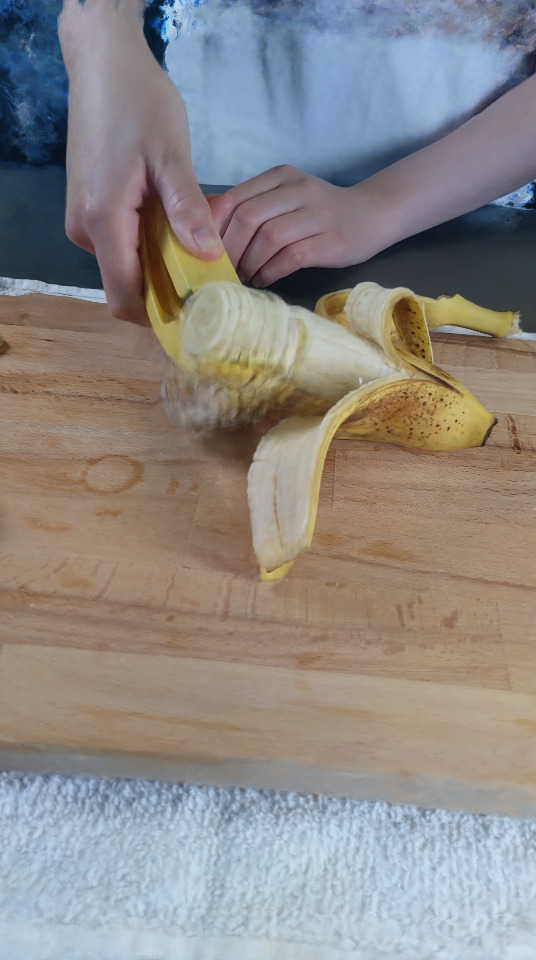}  &
        \topcell{50 250 125 250}{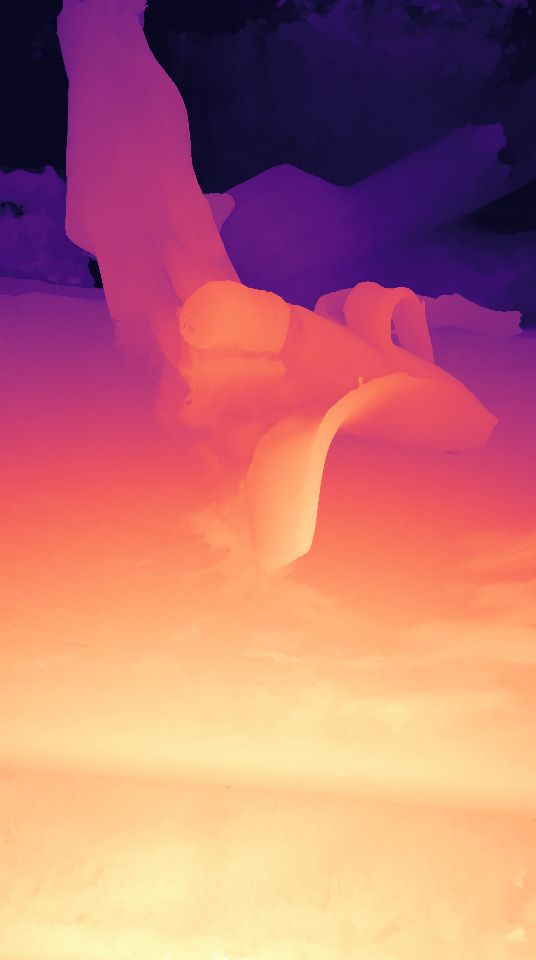}  \\
        &
        &
        \botcell{50 300 125 200}{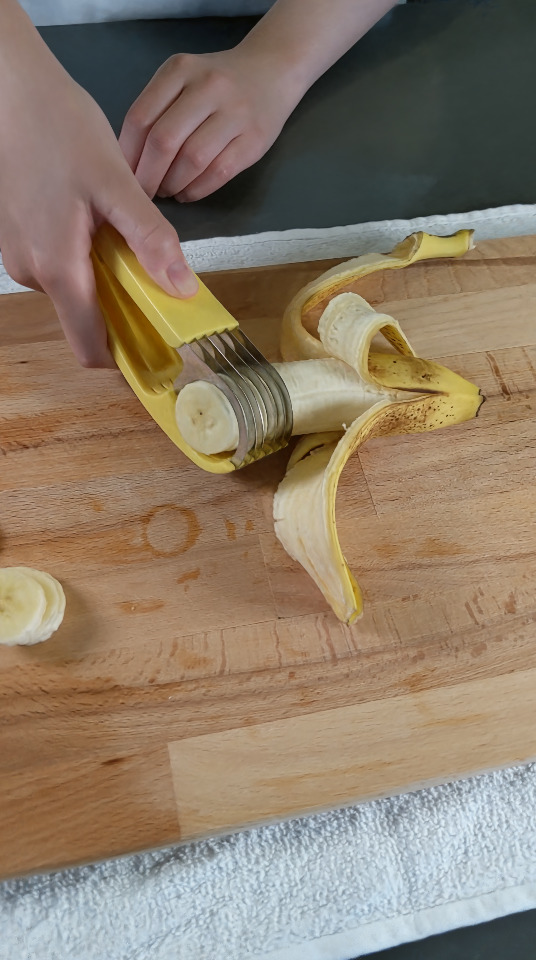} &
        \botcell{50 300 125 200}{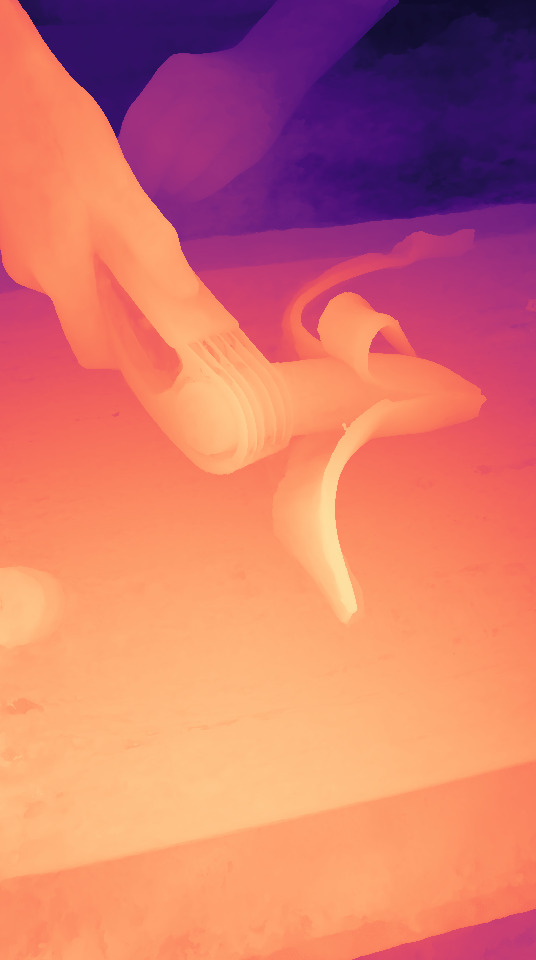} &
        \botcell{50 300 125 200}{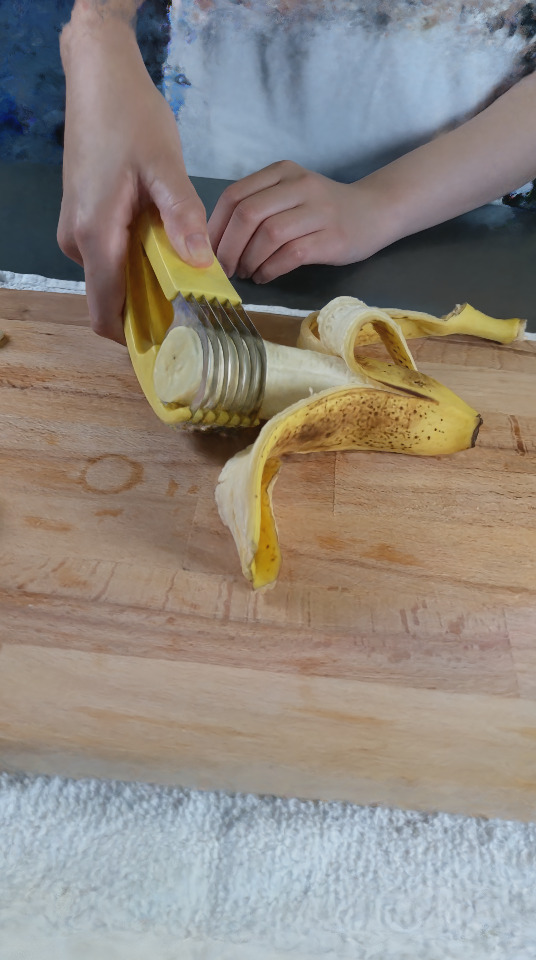} &
        \botcell{50 300 125 200}{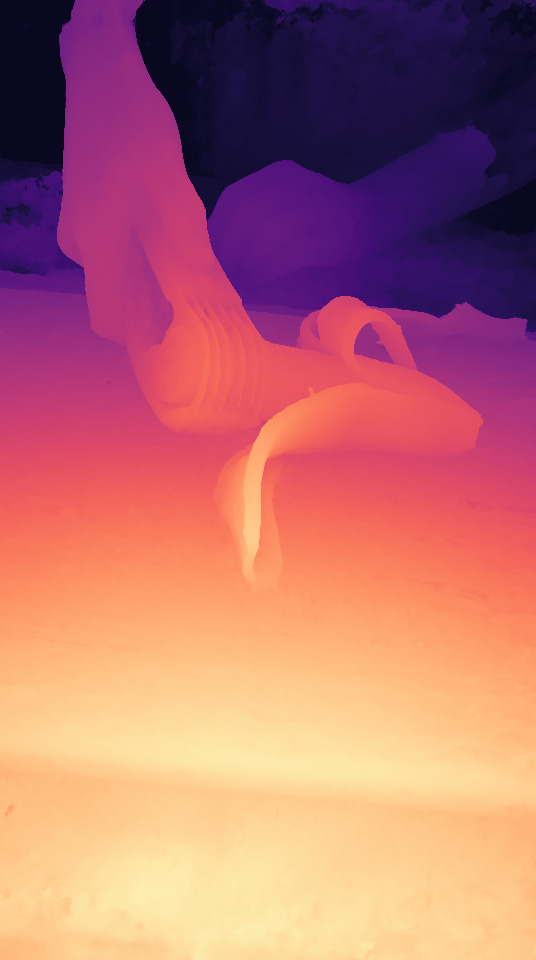} &
        &
        \botcell{50 250 125 250}{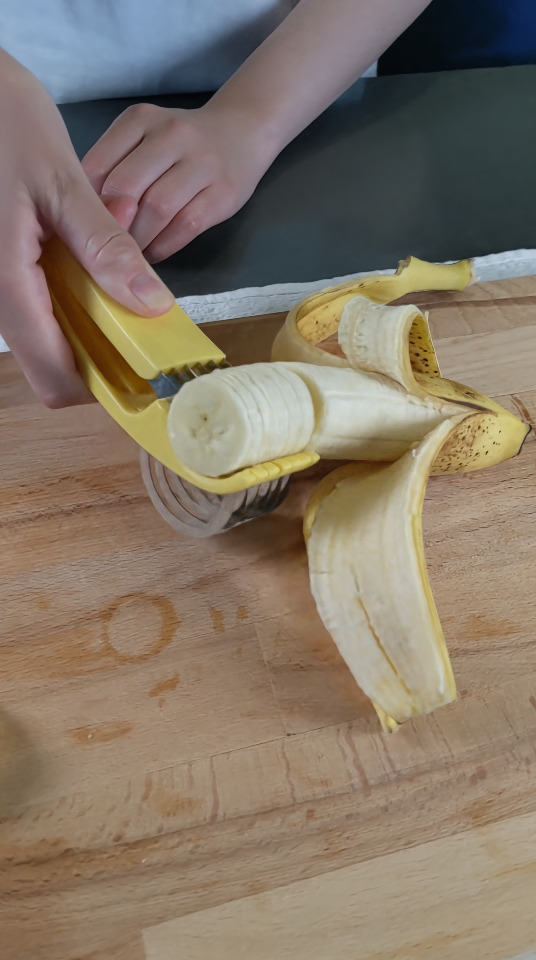} &
        \botcell{50 250 125 250}{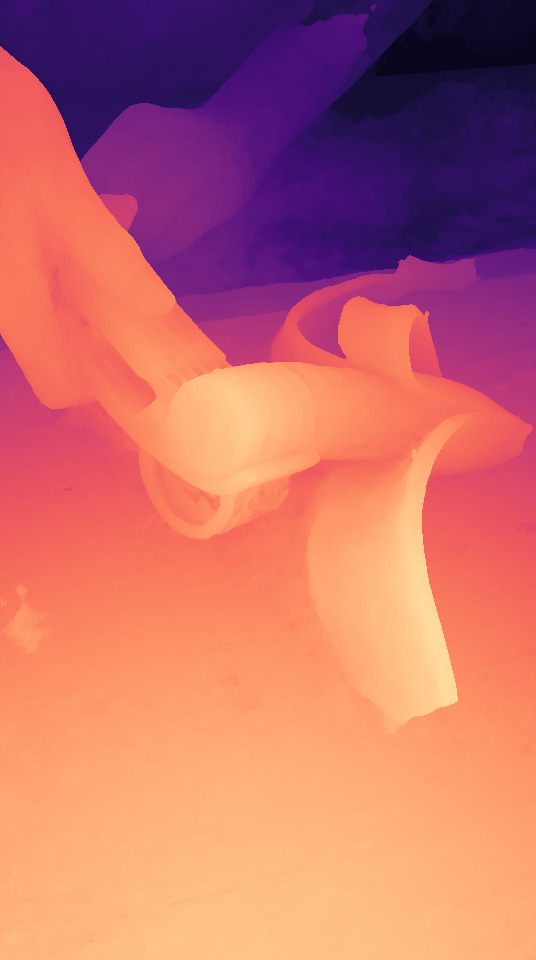} &
        \botcell{50 250 125 250}{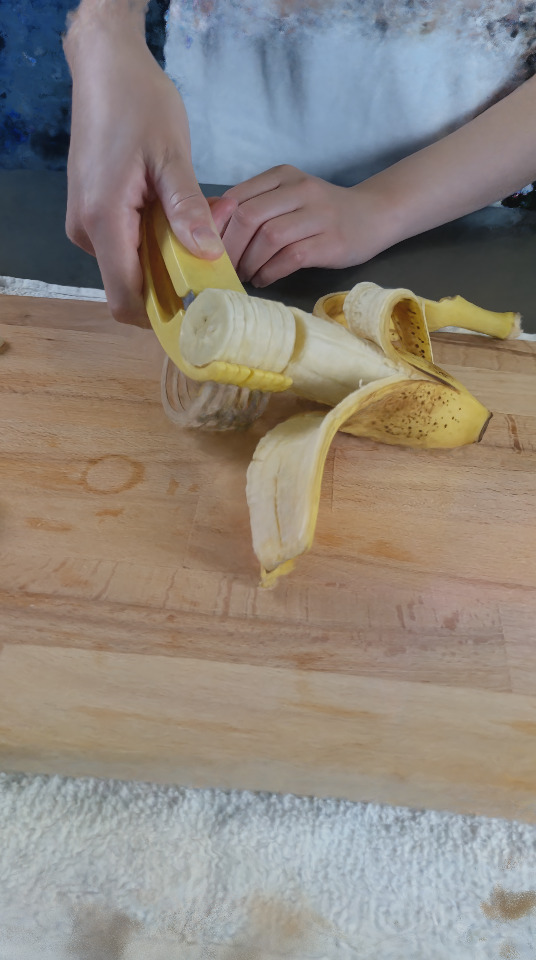} &
        \botcell{50 250 125 250}{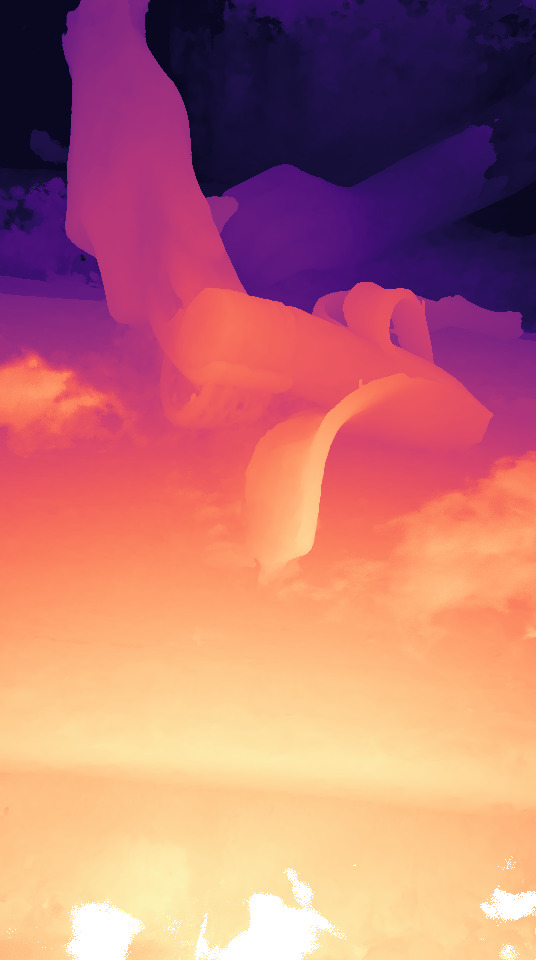} \\
        \\\\\\
        \multirow{2}{*}[0.34in]{\makebox[20pt]{\raisebox{25pt}{\rotatebox[origin=c]{90}{\textsc{Expressions 1}}}} \hspace{-7pt}} &
        \gtcell{250 500 100 500}{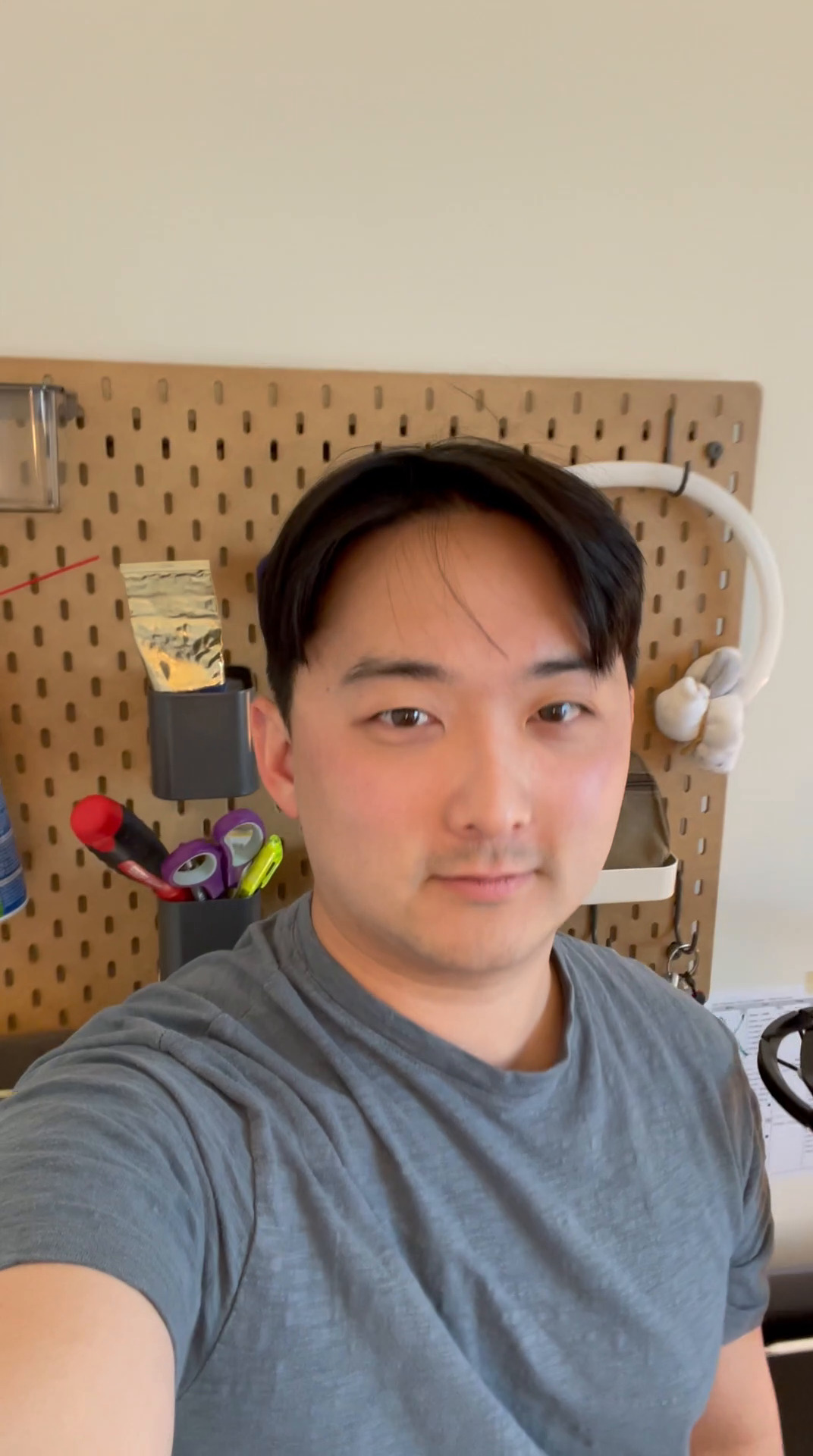} &
        \topcell{125 250 50 250}{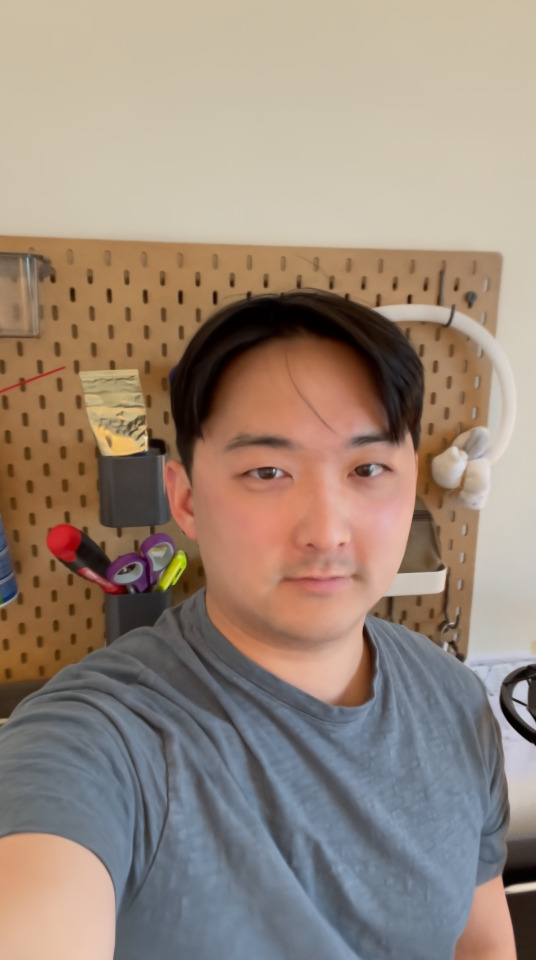} &
        \topcell{125 250 50 250}{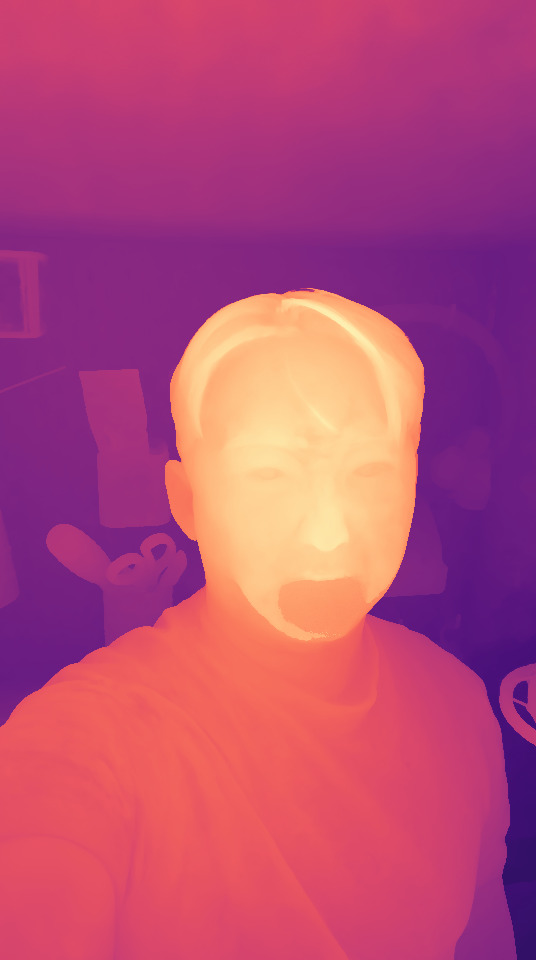} &
        \topcell{125 250 50 250}{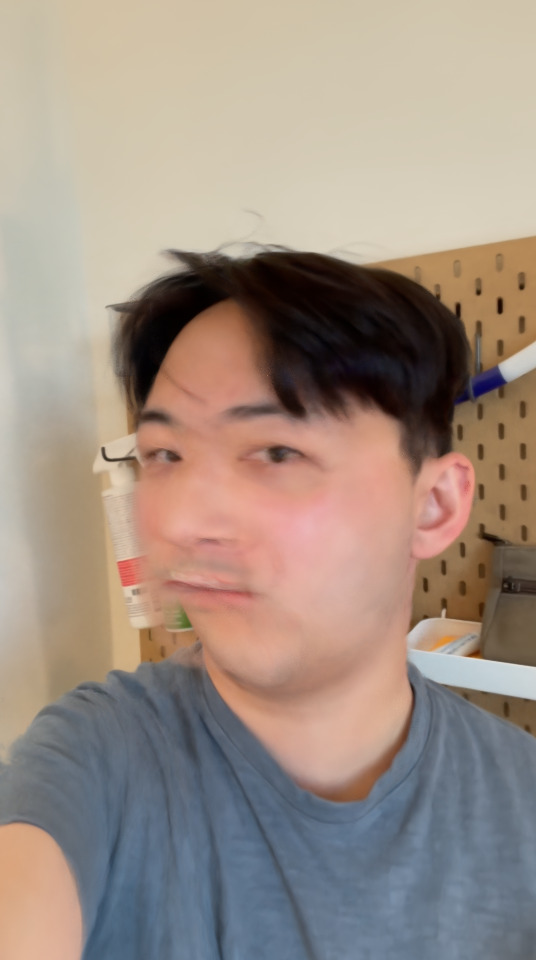} &
        \topcell{125 250 50 250}{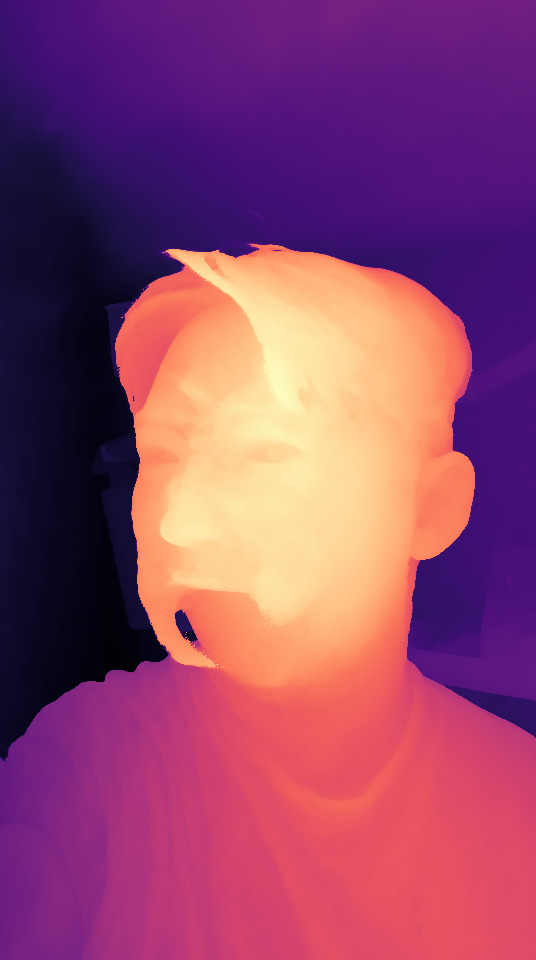} &
        \gtcell{125 250 50 250}{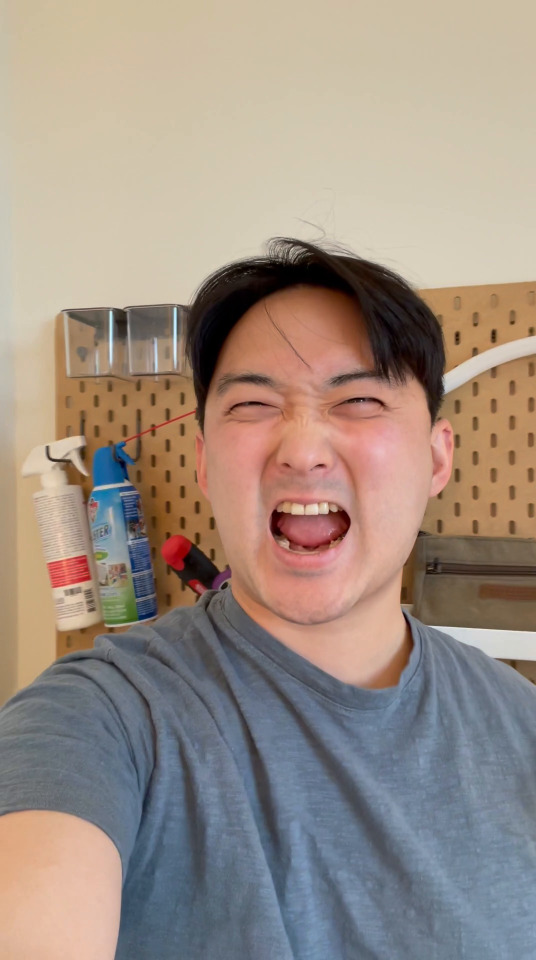} &
        \topcell{125 250 50 250}{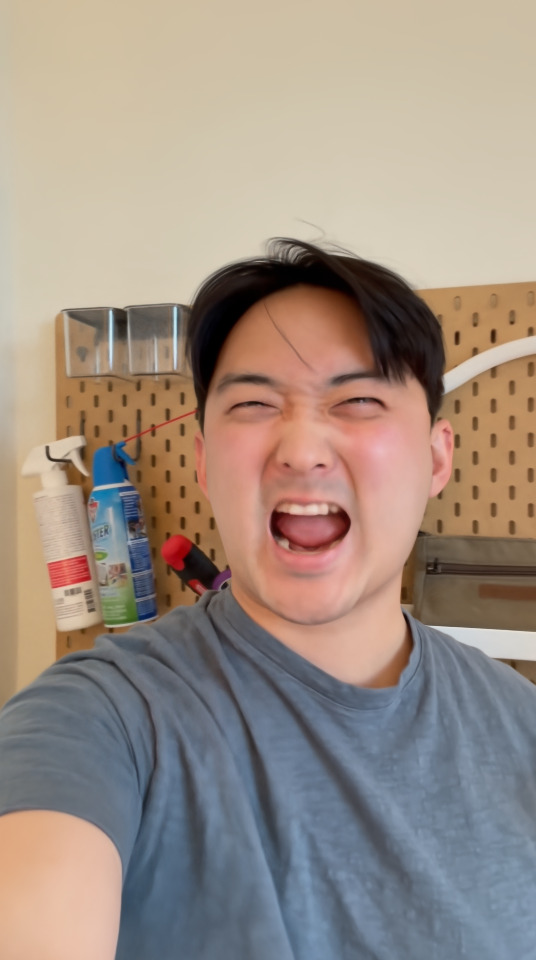} &
        \topcell{125 250 50 250}{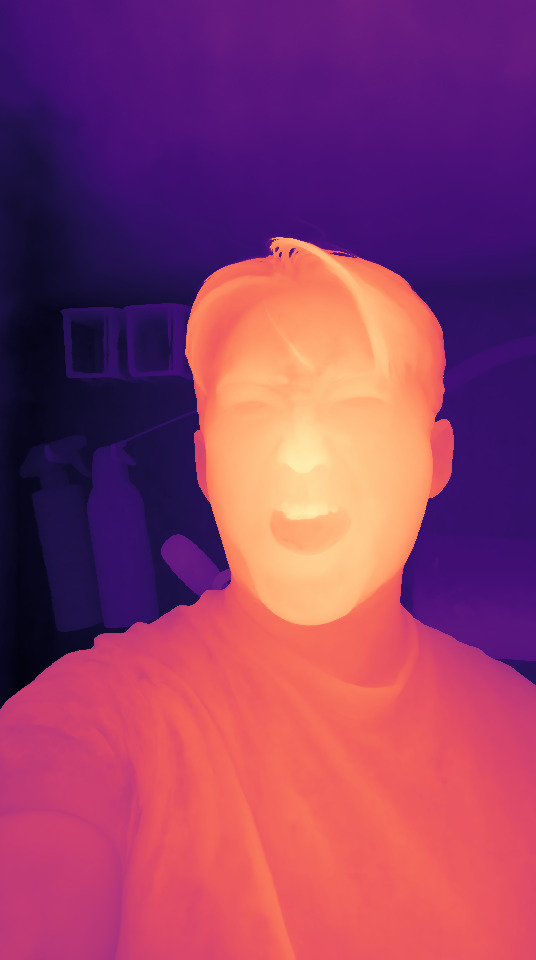} &
        \topcell{125 250 50 250}{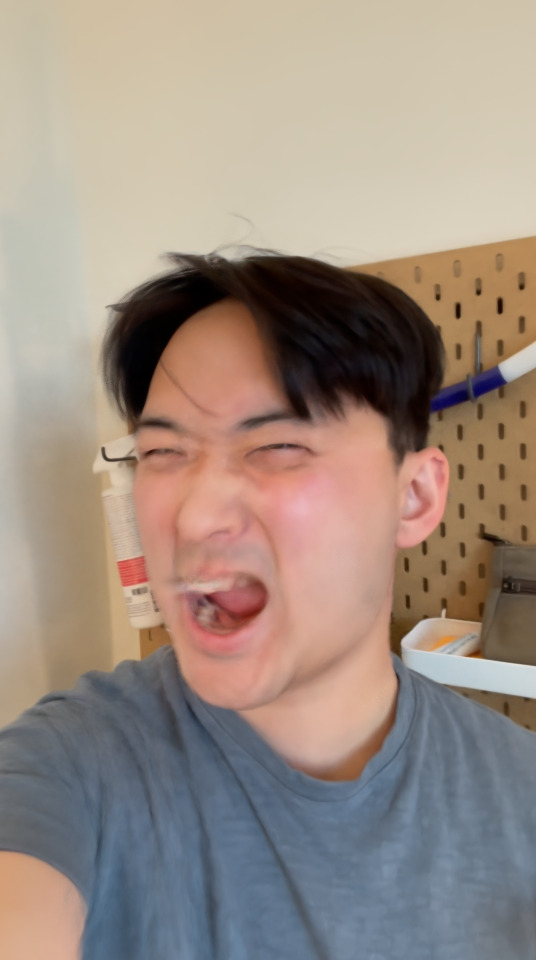}  &
        \topcell{125 250 50 250}{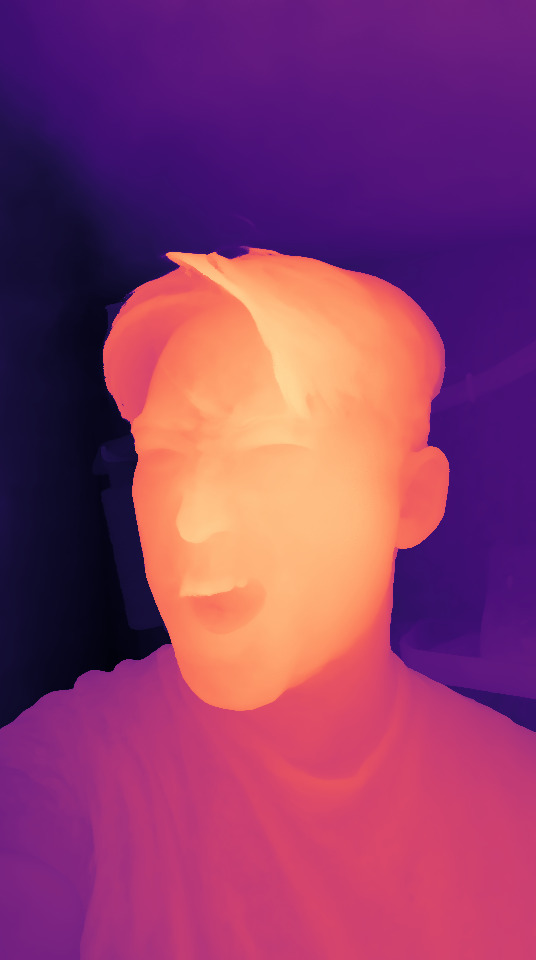}  \\
        &
        &
        \botcell{125 250 50 250}{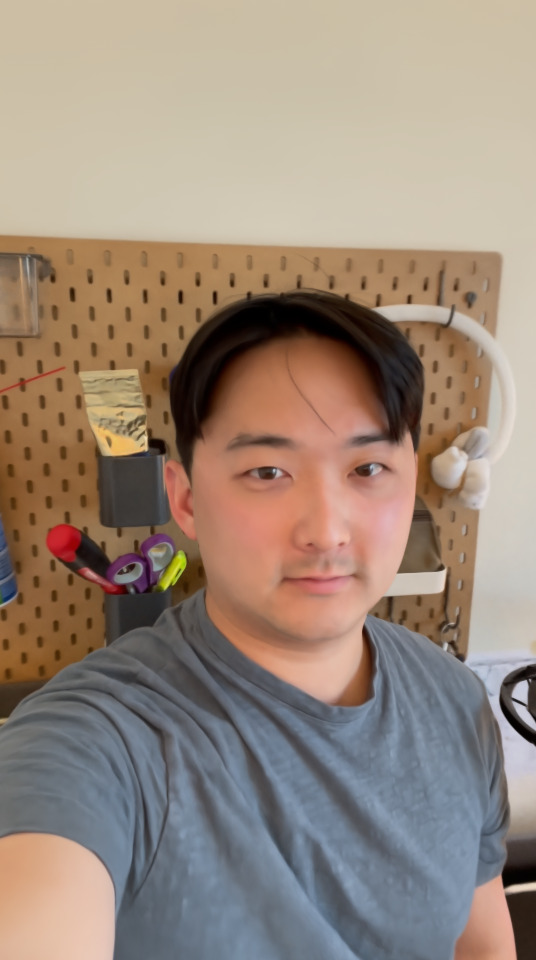} &
        \botcell{125 250 50 250}{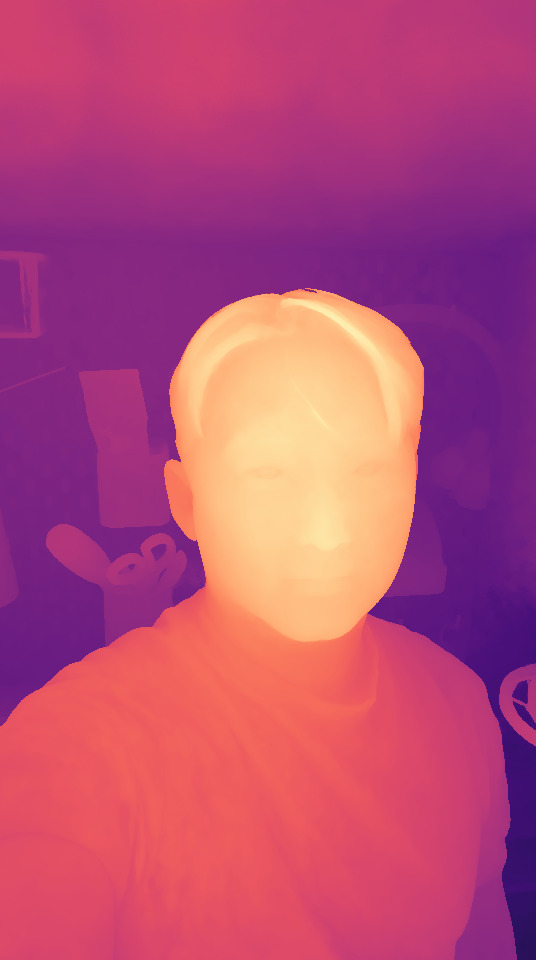} &
        \botcell{125 250 50 250}{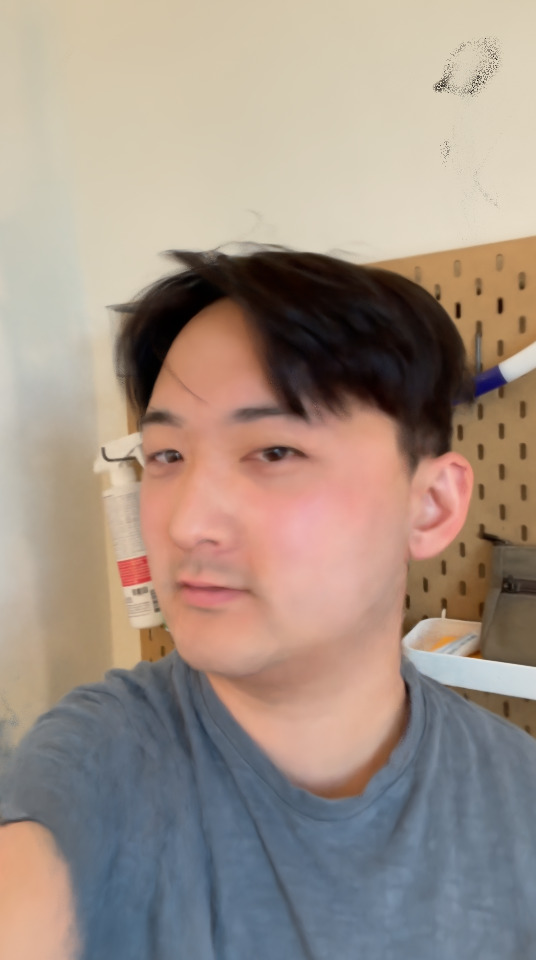} &
        \botcell{125 250 50 250}{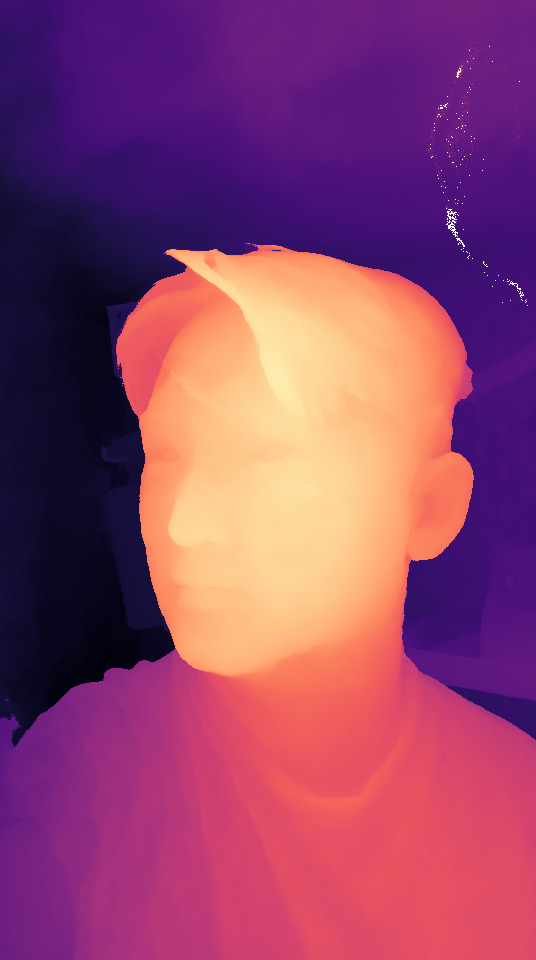} &
        &
        \botcell{125 250 50 250}{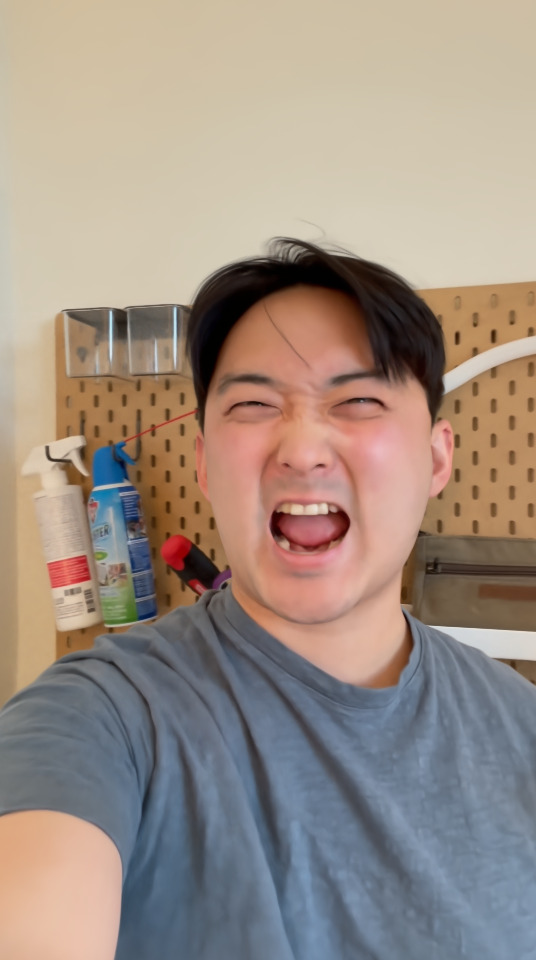} &
        \botcell{125 250 50 250}{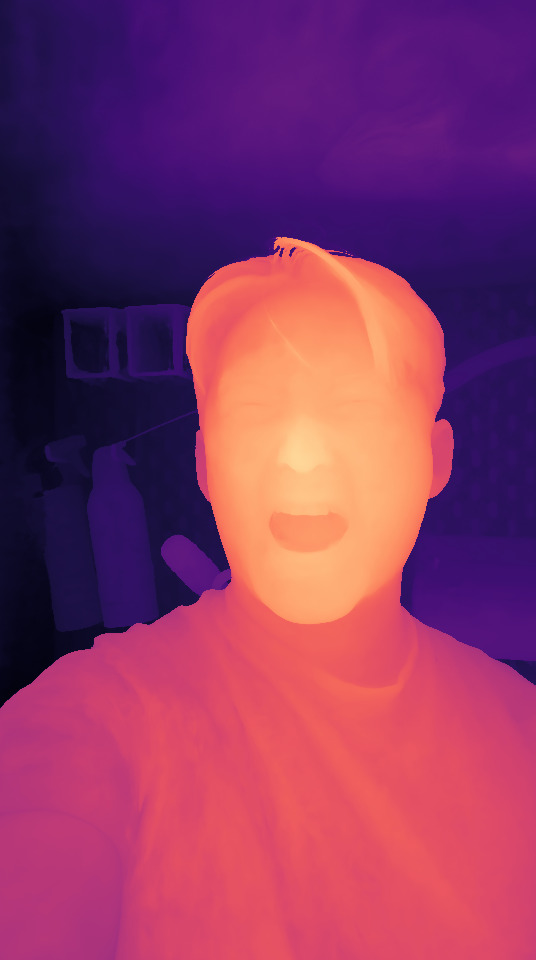} &
        \botcell{125 250 50 250}{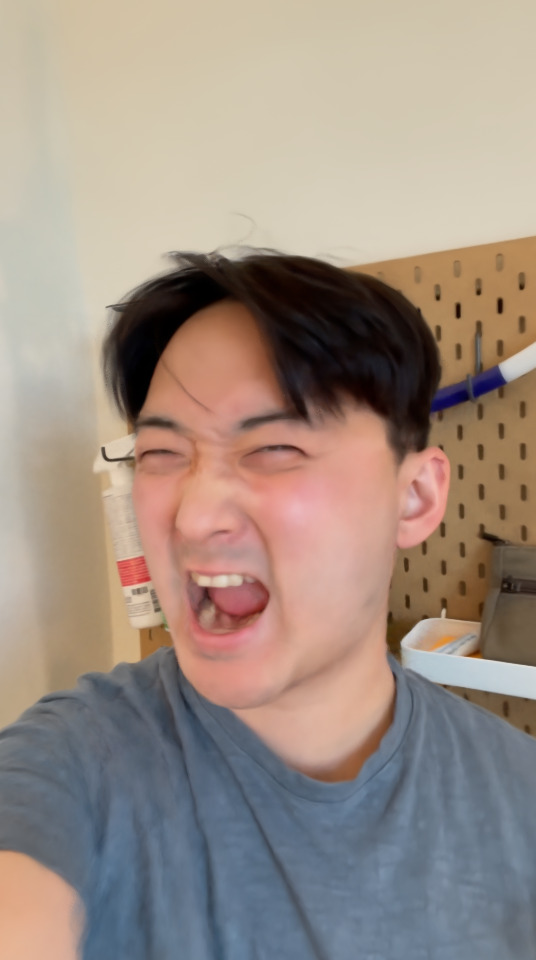} &
        \botcell{125 250 50 250}{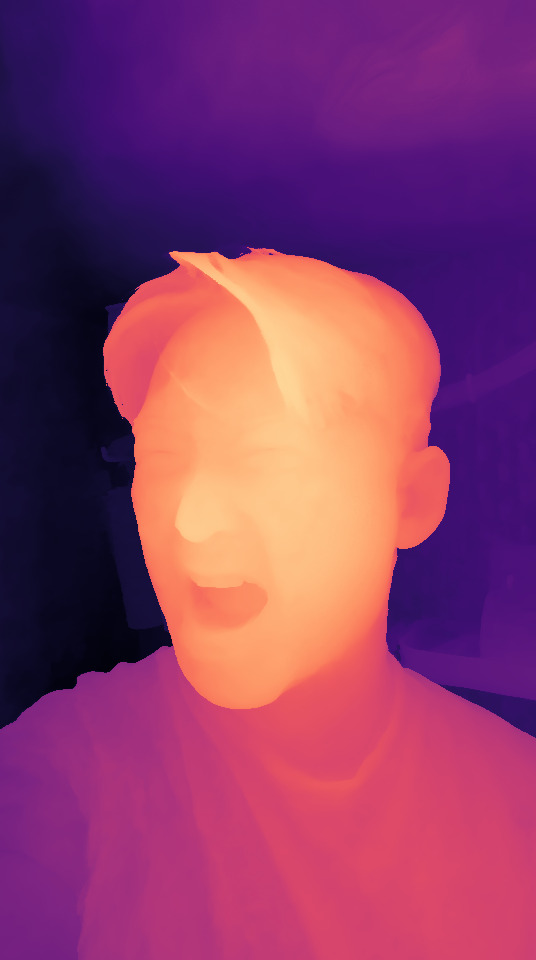} \\
        \\\\\\
        \multirow{2}{*}[0.34in]{\makebox[20pt]{\raisebox{25pt}{\rotatebox[origin=c]{90}{\textsc{Expressions 2}}}} \hspace{-7pt}} &
        \gtcell{125 250 50 250}{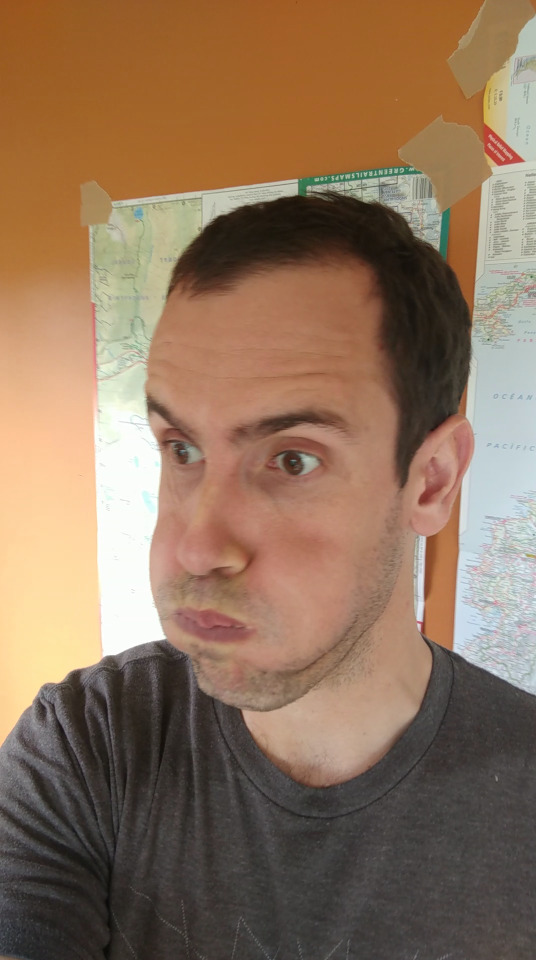} &
        \topcell{125 250 50 250}{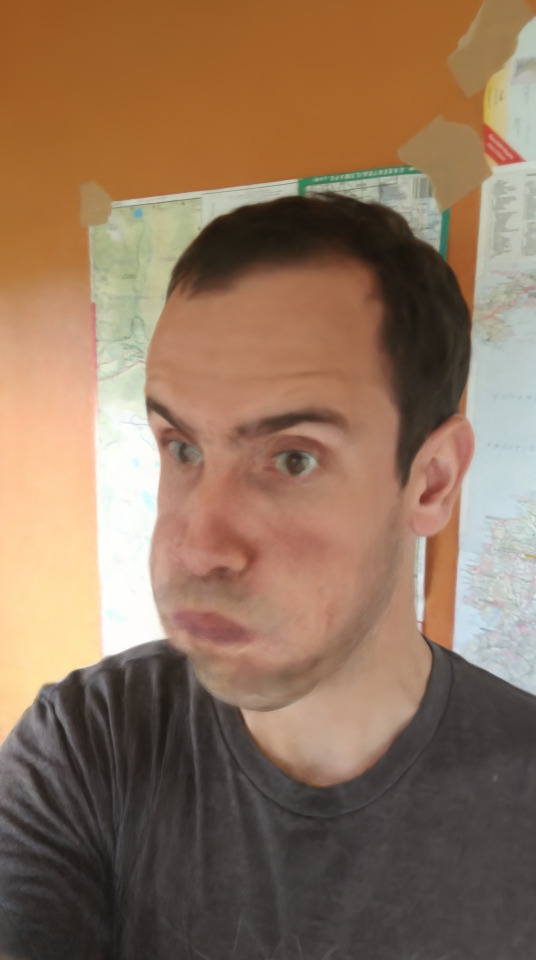} &
        \topcell{125 250 50 250}{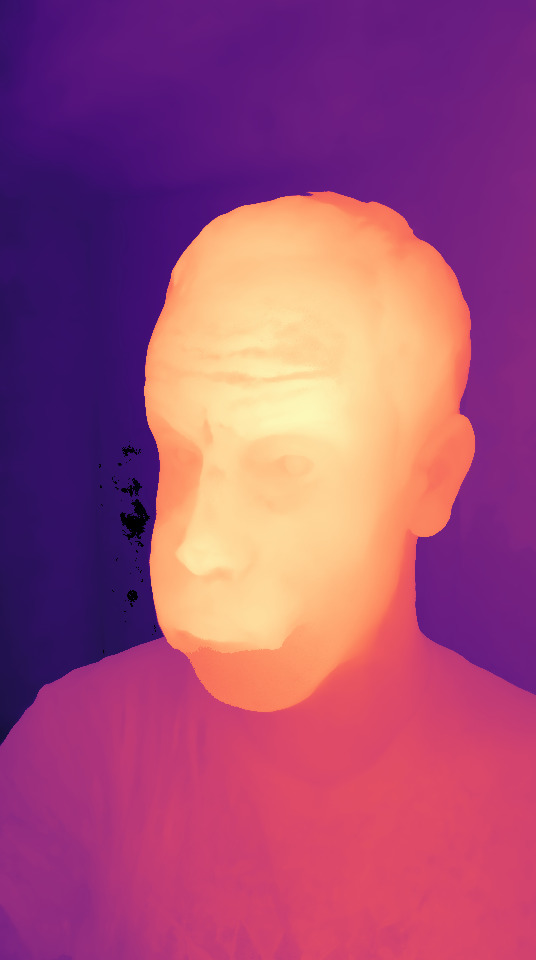} &
        \topcell{125 275 50 225}{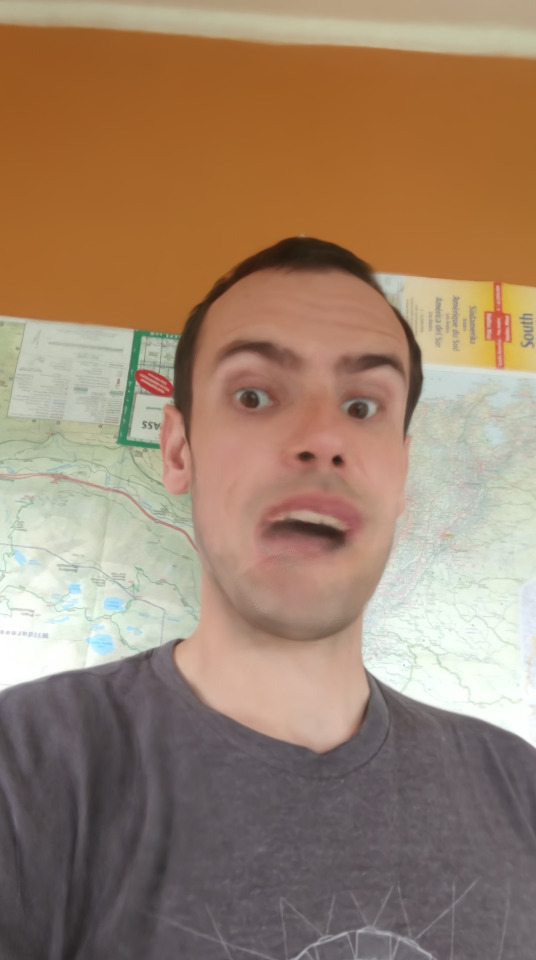} &
        \topcell{125 275 50 225}{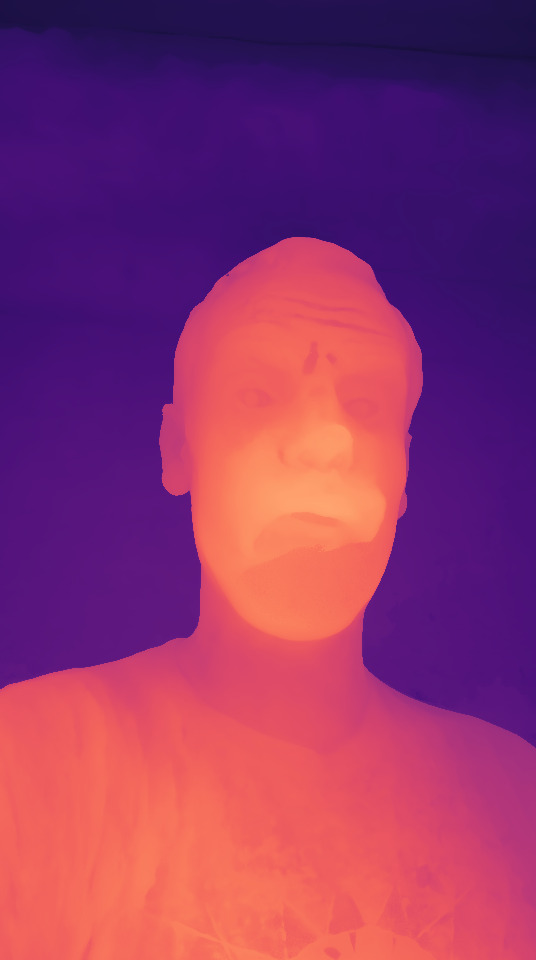} &
        \gtcell{50 250 125 250}{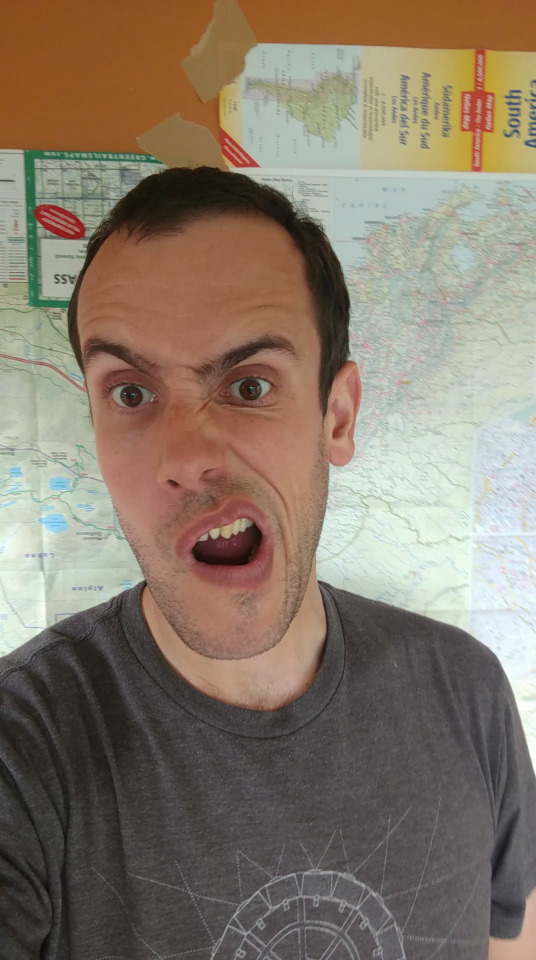} &
        \topcell{50 250 125 250}{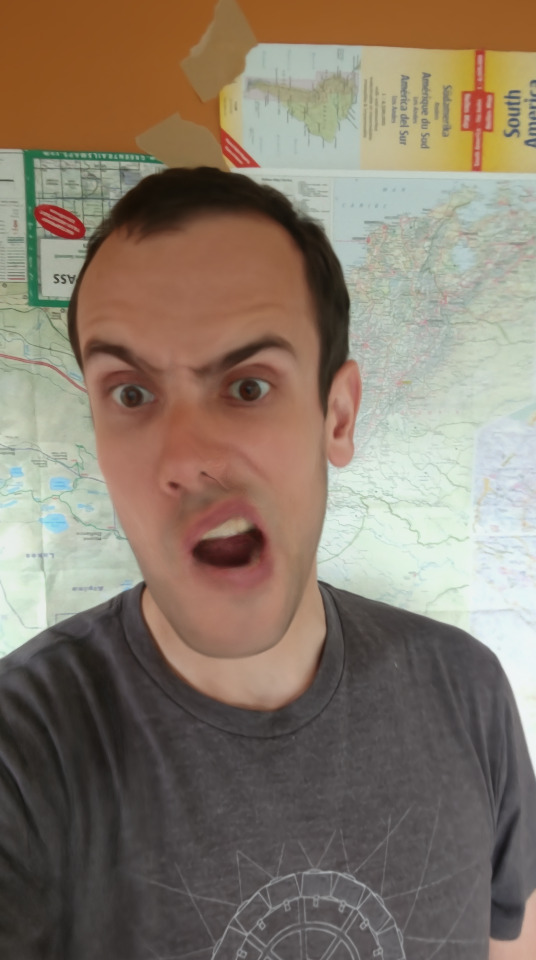} &
        \topcell{50 250 125 250}{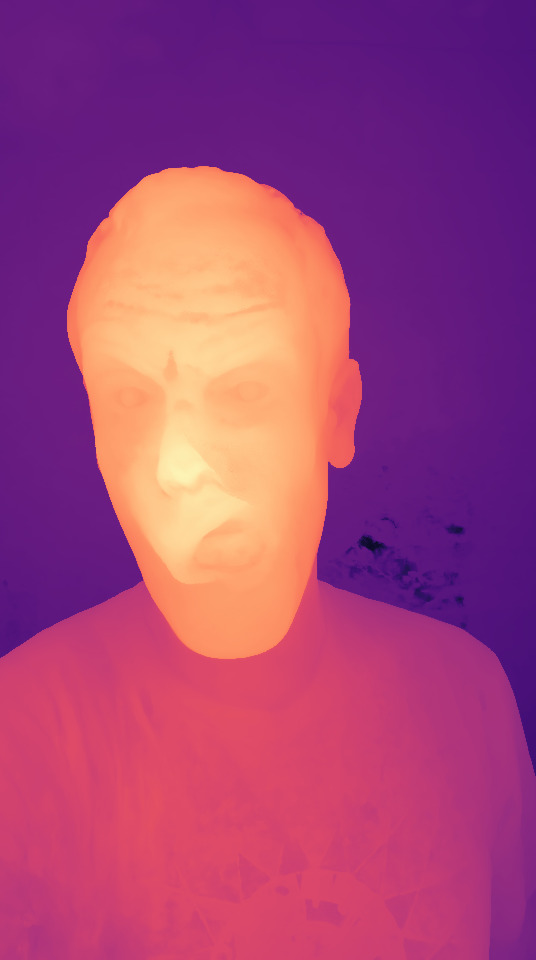} &
        \topcell{125 250 50 250}{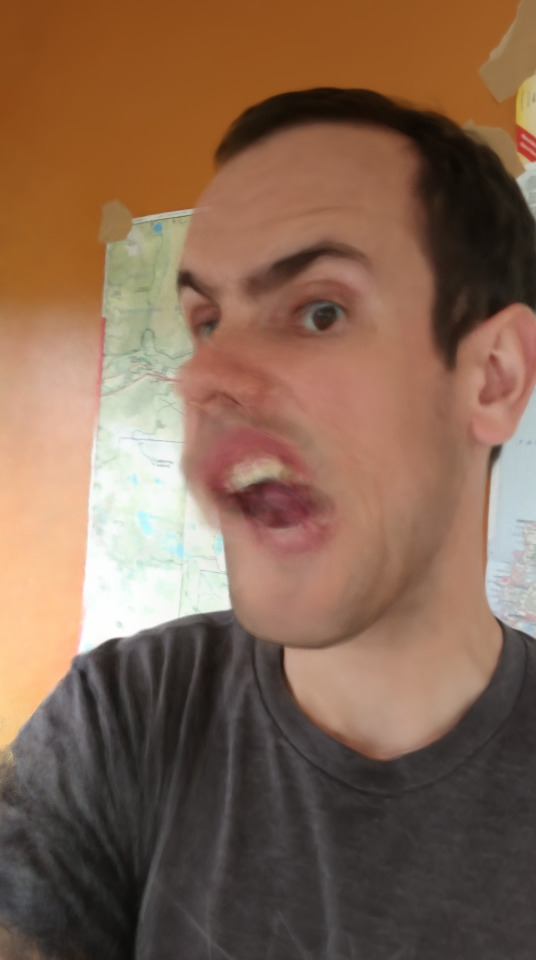}  &
        \topcell{125 250 50 250}{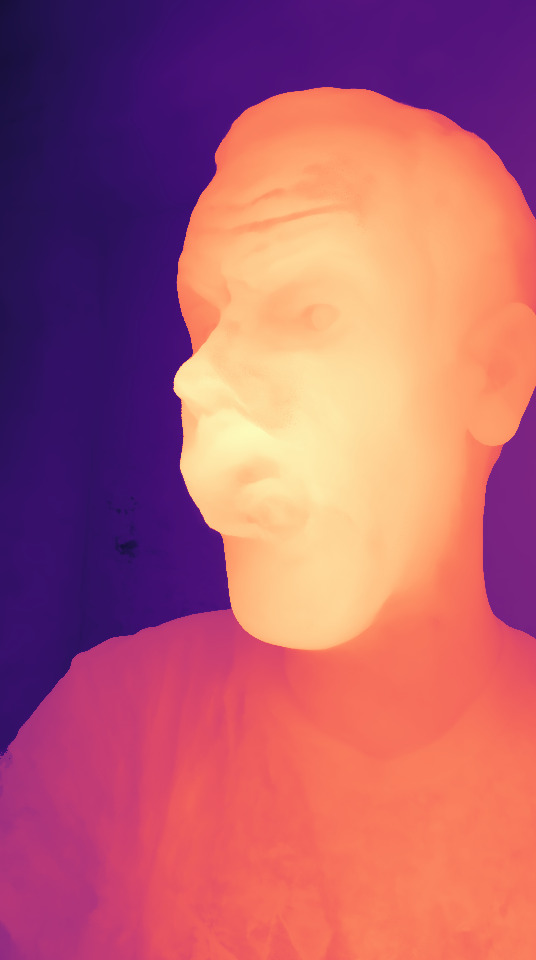}  \\
        &
        &
        \botcell{125 250 50 250}{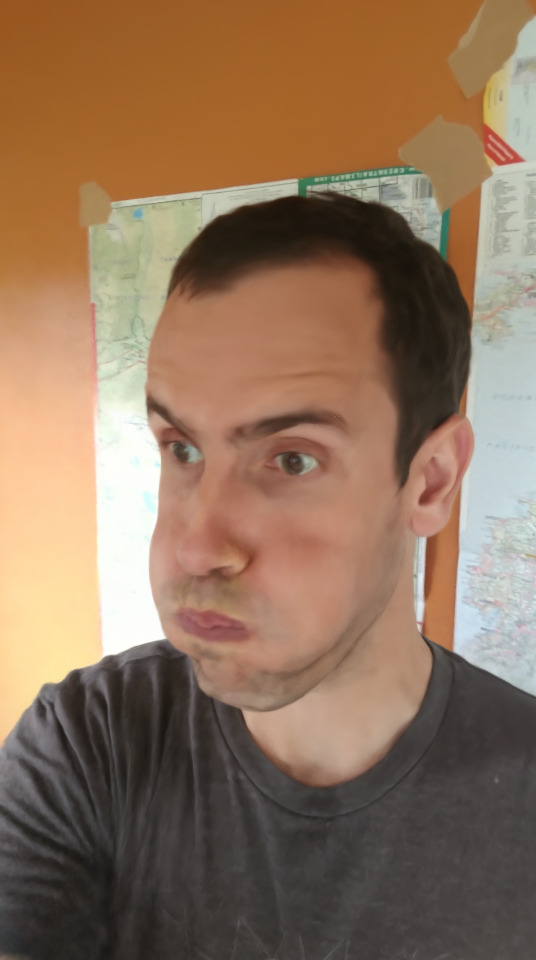} &
        \botcell{125 250 50 250}{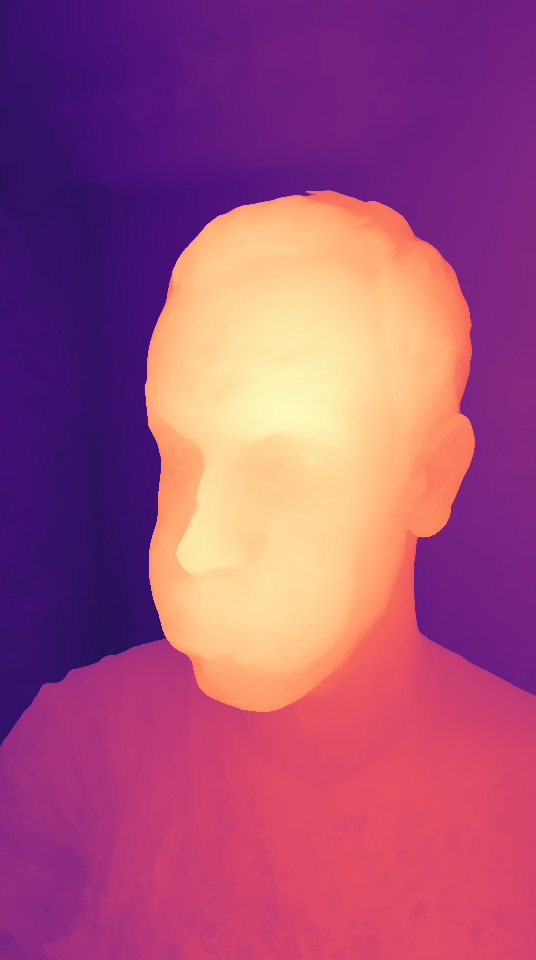} &
        \botcell{125 275 50 225}{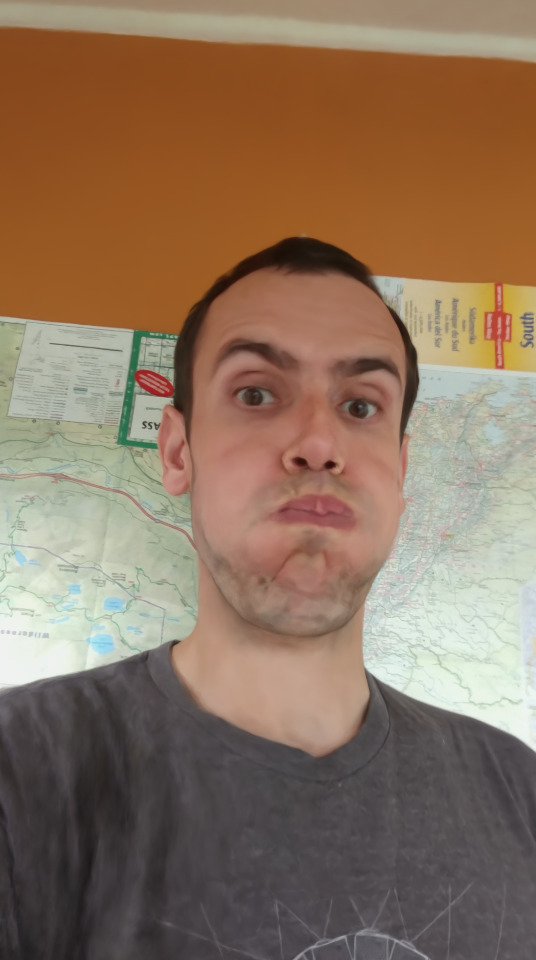} &
        \botcell{125 275 50 225}{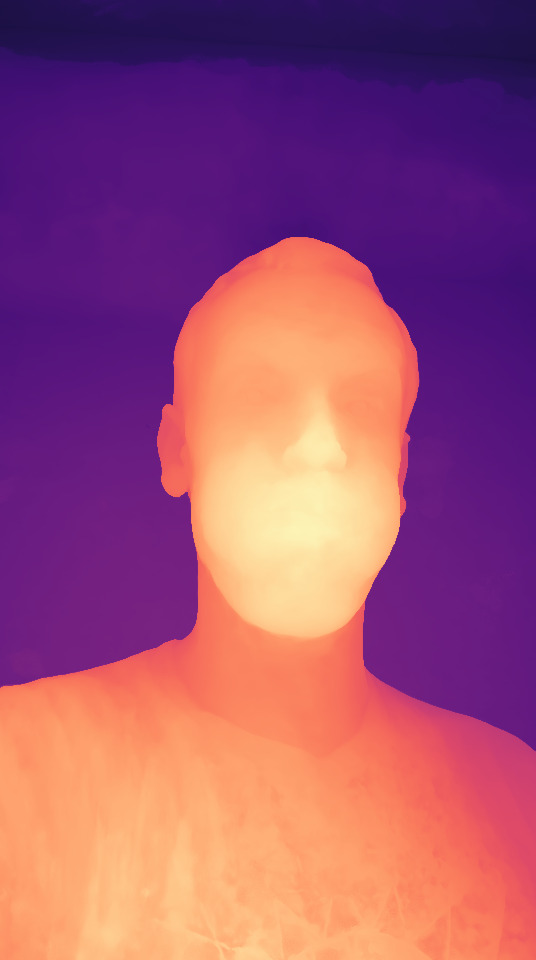} &
        &
        \botcell{50 250 125 250}{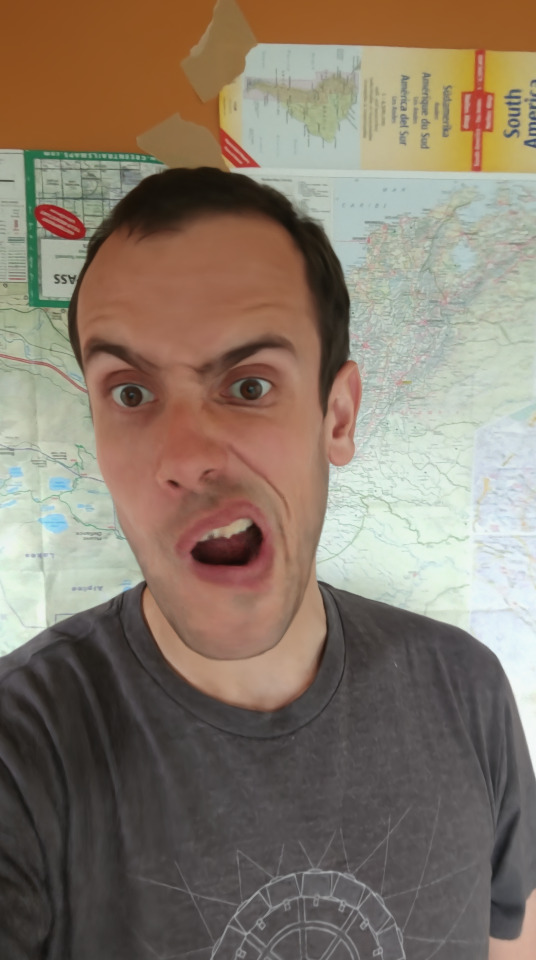} &
        \botcell{50 250 125 250}{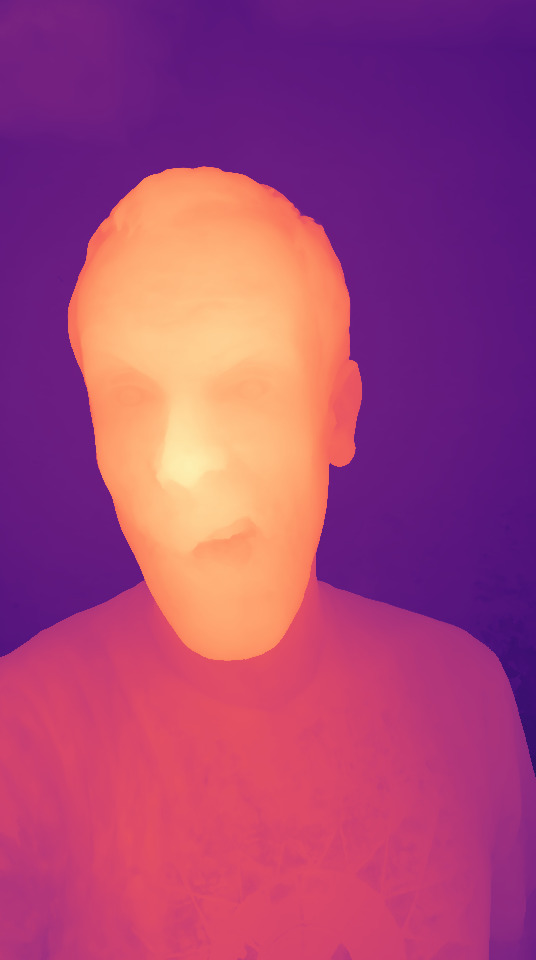} &
        \botcell{125 250 50 250}{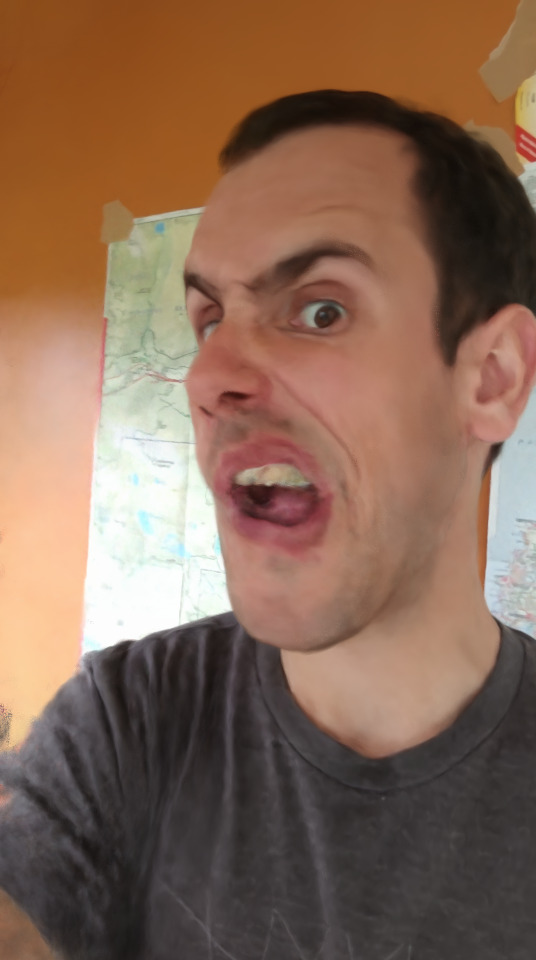} &
        \botcell{125 250 50 250}{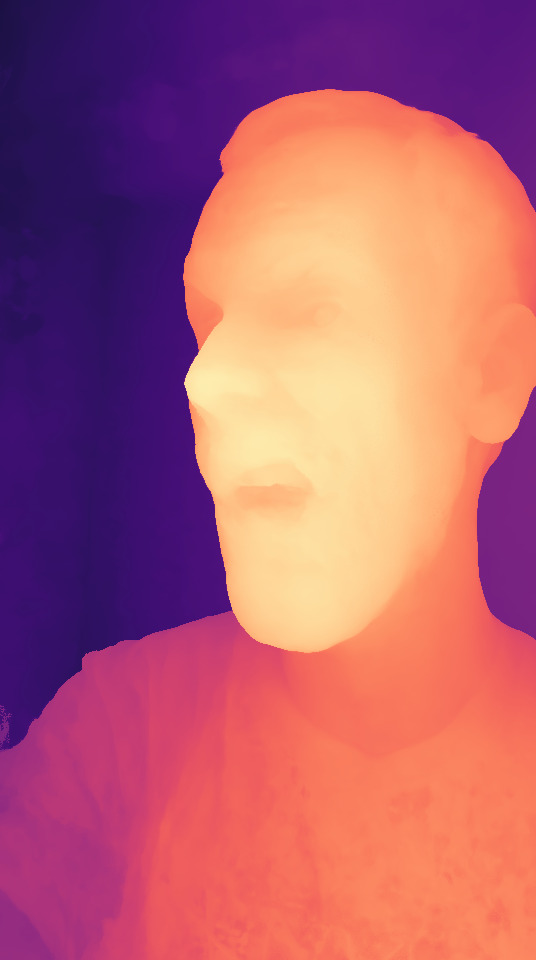}

        \\[2pt]
        
        &
        {\small\makecell{Ground Truth\\Color}} & 
        {\small\makecell{Train View\\Color}} & 
        {\small\makecell{Train View\\Depth}} &
        {\small\makecell{Novel View\\Color}} &
        {\small\makecell{Novel View\\Depth}} &
        {\small\makecell{Ground Truth\\Color}} & 
        {\small\makecell{Train View\\Color}} & 
        {\small\makecell{Train View\\Depth}} &
        {\small\makecell{Novel View\\Color}} &
        {\small\makecell{Novel View\\Depth}}
        \\
    \end{tabular}
}
    \vspace{-6pt}
    \caption{Here we show qualitative comparisons of \topbox{Nerfies}~\cite{park2020nerfies} (top rows) with \botbox{HyperNeRF (ours)} (bottom rows) on four different image sequences. For each sequence we visualize two different moments, each with a different apparent topology. Nerfies is unable to model the topological variation with its deformation field and therefore distorts the geometry in implausible ways in order to explain the training data, while HyperNeRF produces more plausible geometry estimates and more accurate renderings on novel views not seen during training.
    \vspace{0.4in} 
    }
    \label{fig:qualitative_results}
\end{figure*}


\begin{figure}[t]
	\centering
	\begin{subfigure}[b]{1.0\columnwidth}
    	\figcelltbwhite{0.32}{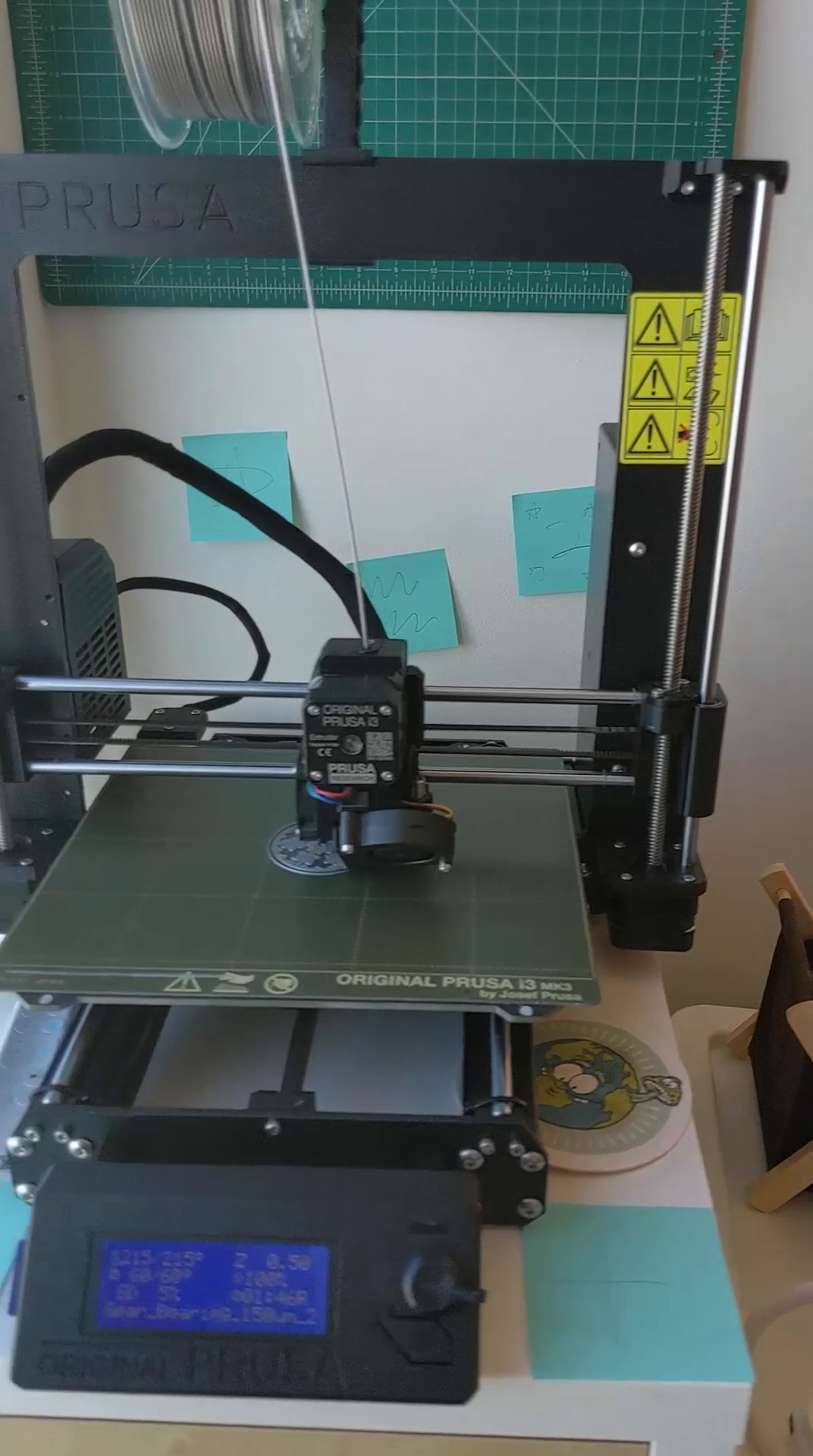}{(a) Input Color}{clip,trim=40 40 80 730}
    	\figcelltbwhite{0.32}{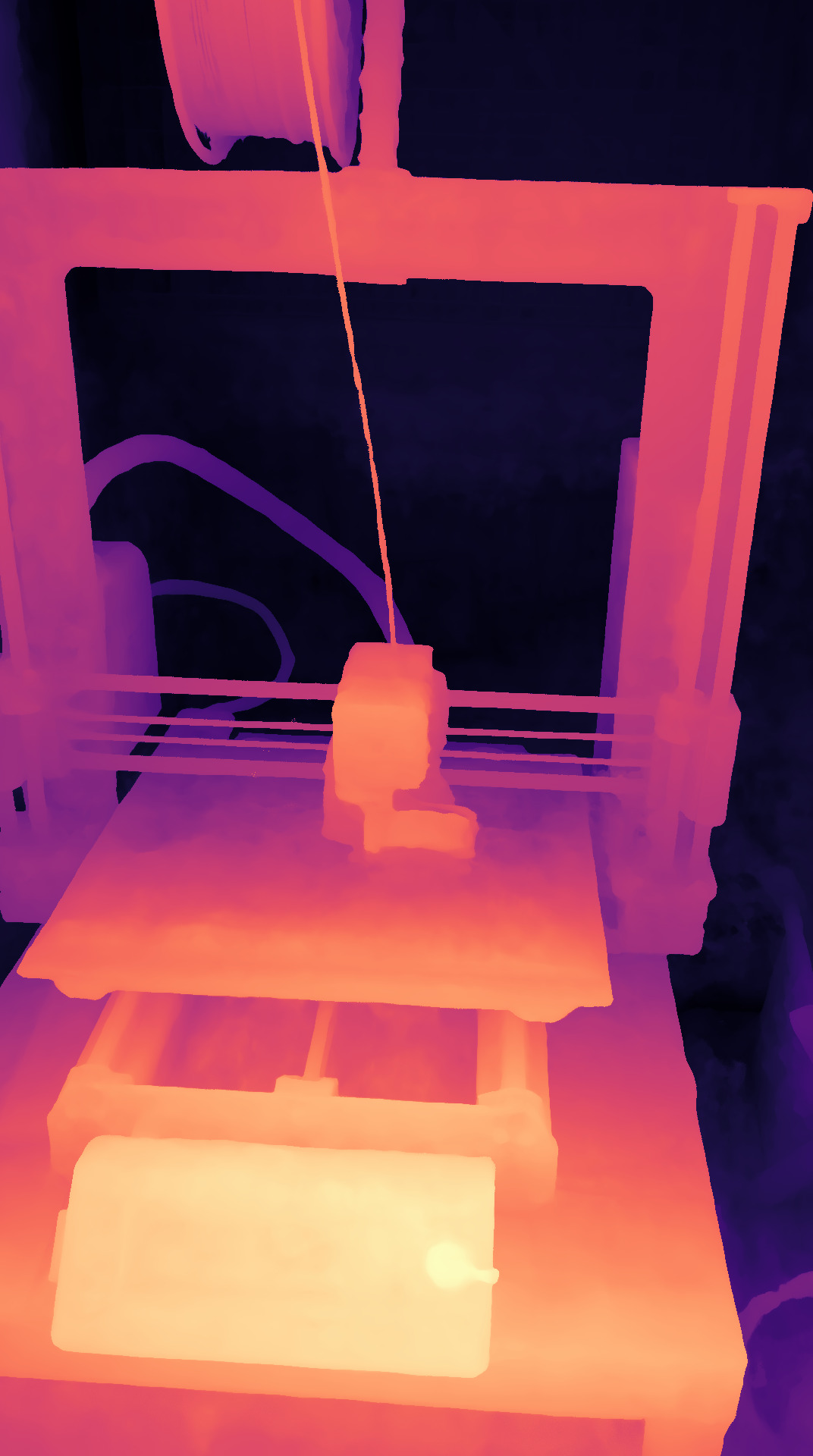}{(b) HyperNeRF (ours)}{clip,trim=40 40 80 730}
    	\figcelltbwhite{0.32}{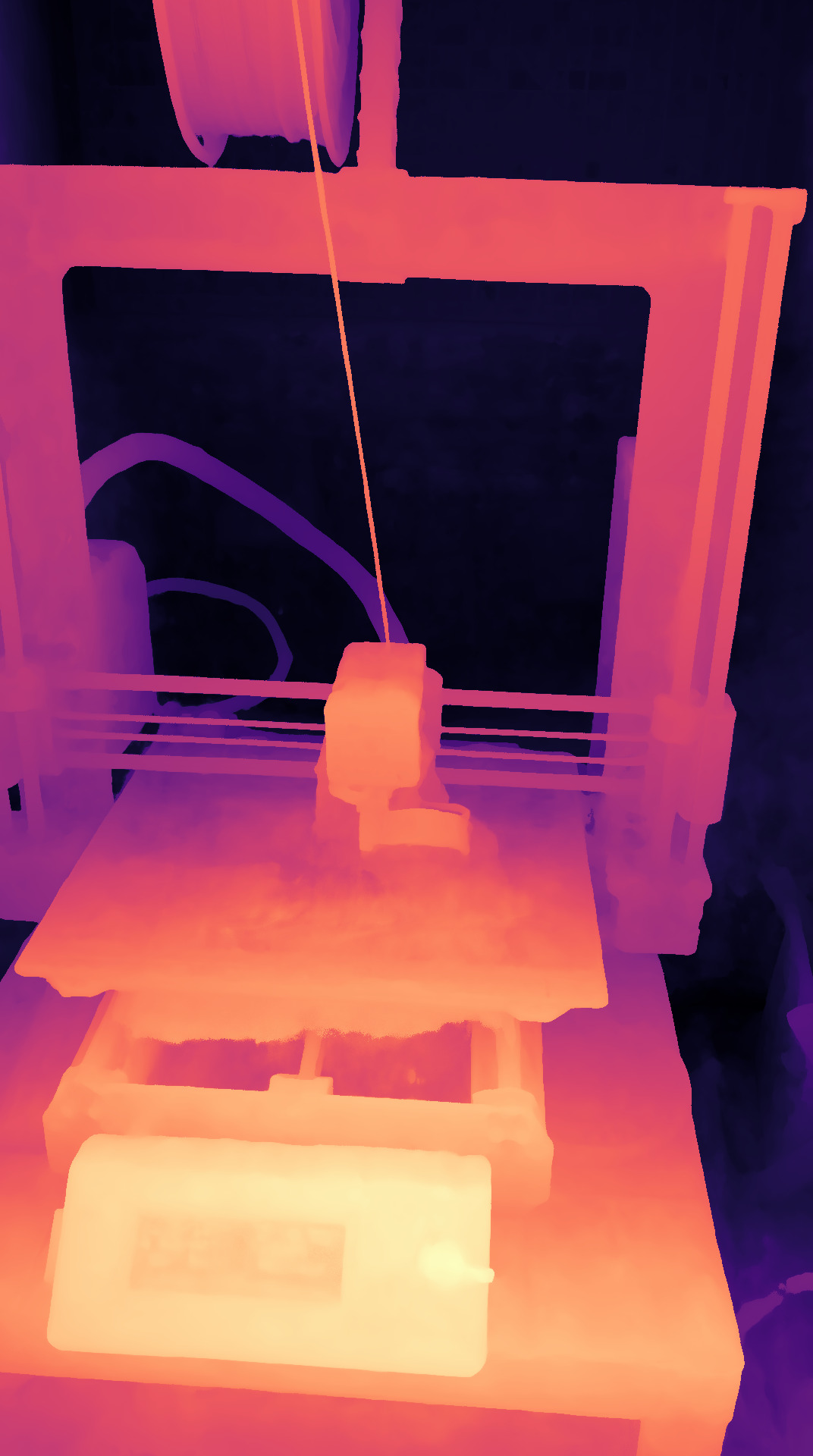}{(c) \cite{park2020nerfies}}{clip,trim=40 40 80 730}
	\end{subfigure}
    \vspace{-5pt}
    \caption{Although not technically a change in topology, the bed of the 3D printer moving relative to its base requires sharp changes in the deformation field, resulting in artifacts in the geometry. \emph{HyperNeRF} alleviates these.}
    \label{fig:qual_3dprinter}
\end{figure}

\section{Experiments}

\subsection{Implementation Details}
Our implementation of NeRF and the deformation field closely follows Nerfies~\cite{park2020nerfies}.
As in NeRF~\cite{mildenhall2020nerf}, we only use an L2 photometric loss.
We use 8 dimensions for the latent appearance and deformations codes.
We exponentially decay our learning rate from $10^{-3}$ to $10^{-4}$.
For the windowed positional encoding parameter $\alpha$ of the deformation field, we follow the same easing schedule as in \citet{park2020nerfies}.
We implement the deformable slicing surface as an MLP with depth 6 and width 64 with a skip connection at the 5th layer. The last layer is initialized with weights sampled from $\mathcal{N}(0,10^{-5})$. We use $m=1$ for the hyper-coordinate input $\hyper$ to the template $T$, fixing $\beta=0$ for 1000 iterations and then linearly increase it to $1$ over 10k iterations. We use $m=6$ for the spatial position input $\spatial$ to the slicing surface field $H$. We use 2 ambient dimensions ($W=2$) for all experiments, as increasing $W$ did not improve performance for our sequences.

We implement our method on top of JaxNeRF~\cite{jaxnerf2020github}, a JAX~\cite{jax2018github} implementation of NeRF. For our evaluation metrics, we train at half of 1080p resolution (960x540) for 250k iterations, using 128 samples per ray with a batch size of 6,144, which takes roughly 8 hours on 4 TPU v4s. For qualitative results, we train at full-HD (roughly 1920x1080) with 256 samples per ray for 1M iterations, which takes roughly 64 hours. 

\subsection{Evaluation}
\label{sec:evaluation}

Here we analyze the performance of our method both quantitatively and qualitatively. To best judge quality, we urge the reader to view the supplementary video which contains many visual results.

\subsubsection{Quantitative Evaluation}
\label{sec:evaluation_quantitative}

We evaluate our method on two tasks: (i) how well it interpolates between different moments seen during training while maintaining visual plausibility; and (ii) its ability to perform novel-view synthesis. We collected our own sequences for these tasks, as existing datasets do not focus on topologically varying scenes. While the dataset of \citet{yoon2020novel} contains two sequences exhibiting topological changes, these sequences have short capture baselines ($\sim$1.1m) and exaggerated frame-to-frame motion due to aggressive temporal sub-sampling.

We measure visual quality with LPIPS~\cite{zhang2018unreasonable}, MS-SSIM~\cite{wang2003multiscale}, and PSNR. Note that because we are reconstructing dynamic scenes with a single moving camera, there can be ambiguities which cause slight differences from the true geometry or appearance of the scene. Because of this, we find that quantitative metrics often do not reflect perceptual quality, as we show in \figref{fig:vrig_results}. In particular, PSNR is incredibly sensitive to small shifts, and will penalize sharp images over blurry results, while MS-SSIM may not pick up artifacts which are obvious to humans. Out of the three metrics, we find that LPIPS best reflects perceptual quality.

\paragraph{Ablations} We also compare with a couple variants of HyperNeRF: ``HyperNeRF (DS)'' uses the deformable slicing surface (\secref{sec:slicing_surfaces}), ``HyperNeRF (DS, w/ elastic)'' uses the elastic regularization loss from \citet{park2020nerfies}, ``HyperNeRF (AP)'' uses the axis-aligned slicing plane, and ``HyperNeRF (w/o deform)'' removes the deformation field and only uses the hyper-space formulation.

\paragraph{Interpolation}
\label{sec:evaluation_interpolation}
Like Nerfies~\cite{park2020nerfies}, our method can interpolate between moments in an input video by interpolating the input embeddings $\latent$ and $\appearance$. Here, we evaluate quality by leaving out intermediate frames from the input video and comparing them with corresponding interpolated frames. Specifically, we collect a set of videos, each 30-60s long, sub-sampled to 15fps and register the frames using COLMAP~\cite{schoenberger2016sfm}. 

We build our dataset by using every 4th frame as a training frame, and taking the middle frame between each pair of training frames as a validation frame.
Since certain frames may have been dropped due failed registration in COLMAP, we use the timestamp of the validation frame to compute its relative position between its corresponding training frames. 

We compare our method with Nerfies~\cite{park2020nerfies}, Neural Volumes~\cite{lombardi2019neural}, and NSFF~\cite{li2020nsff}. For Neural Volumes, we render the interpolated frame by encoding the two reference frames with the encoder and linearly interpolating the latent codes. For NSFF, we linearly interpolate the input time variable between the two reference frames.
\tabref{tab:evaluation_interpolation} shows that our method outperforms all three baselines in most cases. 
We show qualitative results for the interpolation sequences in \figref{fig:interp_results1} and \figref{fig:interp_results2}, with additional results in the supplementary materials.

\newcommand{\tablefirst}[0]{\cellcolor{myred}}
\newcommand{\tablesecond}[0]{\cellcolor{myorange}}
\newcommand{\tablethird}[0]{\cellcolor{myyellow}}

\begin{table}[t]
\centering

\caption{
Metrics on the interpolation task, averaged across all sequences. We report PSNR, MS-SSIM, LPIPS, and the ``average'' metric of \citet{barron2021mipnerf}. We color each row as \colorbox{myred}{\textbf{best}}, \colorbox{myorange}{\textbf{second best}}, and \colorbox{myyellow}{\textbf{third best}}. See Sec.~\ref{sec:evaluation_interpolation} for details, and the supplement for metrics for each sequence.
\label{tab:evaluation_interpolation}
}

\resizebox{0.8\linewidth}{!}{
\setlength{\tabcolsep}{2pt}


\begin{tabular}{l||ccc|c}

\toprule

& \multicolumn{1}{c}{ \footnotesize PSNR$\uparrow$ }
& \multicolumn{1}{c}{ \footnotesize MS-SSIM$\uparrow$ }
& \multicolumn{1}{c|}{ \footnotesize LPIPS$\downarrow$ }
& \multicolumn{1}{c}{ \footnotesize Avg$\downarrow$ }
\\
\hline

  NeRF~\cite{mildenhall2020nerf}
  &$21.3$
  &$.745$
  &$.490$
  &$.126$
  
  \\   NV~\cite{lombardi2019neural}
  &$25.3$
  &$.880$
  &$.214$
  &$.0630$
  
  \\   NSFF~\cite{li2020nsff}
  &$25.5$
  &$.863$
  &$.242$
  &$.0663$
  
  \\   Nerfies~\cite{park2020nerfies}
  &$27.7$
  &$.908$
  &\tablesecond$.193$
  &$.0487$
  
  \\   Nerfies (w/o elastic)
  &$28.0$
  &$.909$
  &\tablethird$.193$
  &\tablethird$.0476$
  
  \\ \hline  Hyper-NeRF (DS)
  &\tablefirst$28.3$
  &\tablefirst$.914$
  &\tablefirst$.185$
  &\tablefirst$.0456$
  
  \\   Hyper-NeRF (DS, w/ elastic)
  &\tablethird$28.1$
  &\tablethird$.912$
  &$.195$
  &\tablesecond$.0471$
  
  \\   Hyper-NeRF (AP)
  &\tablesecond$28.3$
  &\tablesecond$.912$
  &$.208$
  &$.0476$
  
  \\   Hyper-NeRF (w/o deform)
  &$27.4$
  &$.895$
  &$.253$
  &$.0557$
  
  \\ \bottomrule

\end{tabular}

}


\end{table}

\paragraph{Novel-view Synthesis}
\label{sec:evaluation_nvs}
We evaluate the novel-view synthesis quality of our method on topologically varying scenes. We render the scene at fixed moments from unseen viewpoints and compare how well the predicted images match the corresponding ground truth image. To allow for a fair comparison with prior work, we closely follow the evaluation protocol of \citet{park2020nerfies}. We create a capture rig comprised of a pole with two Pixel 3 phones rigidly attached roughly 16cm apart. Please see \citet{park2020nerfies} for details on dataset processing.
We evaluate on the \textsc{Broom} sequence from \citet{park2020nerfies} which exhibits a topological change when the broom contacts the ground, and augment the sequences of \citet{park2020nerfies} with 4 additional sequences exhibiting topological changes: \textsc{3D Printer}, \textsc{Chicken}, \textsc{Expressions}, and \textsc{Peel Banana}. \tabref{tab:evaluation_vrig} reports the metrics computed on unseen validation views.

\paragraph{Baselines}
\label{sec:evaluation_baselines}
For view-synthesis, we compare with NeRF~\cite{mildenhall2020nerf}, Neural Volumes~\cite{lombardi2019neural} and two recent dynamic NeRF methods: NSFF~\cite{li2020nsff} and Nerfies~\cite{park2020nerfies}.
For interpolation, we compare with NeRF, Nerfies, and Neural Volumes. To evaluate Neural Volumes on the interpolation task, we compute the latent codes of the two training images corresponding to each validation image with the image encoder, and linearly interpolate the encoding vector using the identical procedure mentioned in \secref{sec:evaluation_interpolation}. As Neural Volumes works with fixed size inputs, we only evaluate error metrics on a central crop, resulting in slightly improved results for this baseline.
We do not compare with \citet{yoon2020novel} as their code was not available. Instead, we compare with NSFF~\cite{li2020nsff} which achieves higher quality on the dataset of \citet{yoon2020novel}.
Note that the default hyper-parameters for NSFF~\cite{li2020nsff} provided with the code release performs poorly on our sequences --- we therefore contacted the authors to help us tune the hyper-parameters (see appendix).

\subsubsection{Qualitative Results}
\label{sec:evaluation_qualitative}

\figref{fig:qualitative_results} shows qualitative results, comparing our method with \citet{park2020nerfies}. We also show visual results from the validation rig dataset in \figref{fig:vrig_results}. Note how some image quality metrics sometimes prefer blurry results (e.g. \textsc{Expressions}).

Like our method, Nerfies~\cite{park2020nerfies} is able to produce sharp results with few artifacts. However, our method better reconstructs the poses of objects in scenes such as \textsc{Peel Banana} and \textsc{Expressions} in \figref{fig:vrig_results}. This is also shown by the warped faces in \figref{fig:qualitative_results}, where Nerfies often does not model geometry for the chin.

\renewcommand{\tablefirst}[0]{\cellcolor{myred}}
\renewcommand{\tablesecond}[0]{\cellcolor{myorange}}
\renewcommand{\tablethird}[0]{\cellcolor{myyellow}}

\begin{table*}[t]
\centering

\caption{
Quantitative evaluation on validation rig captures. We compare with baselines and ablations of our method. We color each row as \colorbox{myred}{\textbf{best}}, \colorbox{myorange}{\textbf{second best}}, and \colorbox{myyellow}{\textbf{third best}}. Note that traditional metrics like PSNR and SSIM are sensitive to small shifts, penalizing sharp images over blurry results (see \figref{fig:vrig_results} below). See \secref{sec:evaluation_nvs} for more details on these experiments.
\label{tab:evaluation_vrig}
}

\resizebox{1.0\linewidth}{!}{
\setlength{\tabcolsep}{2pt}


\begin{tabular}{l||ccc|ccc|ccc|ccc|ccc|ccc}

\toprule
& \multicolumn{ 3 }{c}{
  \makecell{
  \textsc{\small Broom }
\\(197 images)
  }
}
& \multicolumn{ 3 }{c}{
  \makecell{
  \textsc{\small 3D printer }
\\(207 images)
  }
}
& \multicolumn{ 3 }{c}{
  \makecell{
  \textsc{\small Chicken }
\\(164 images)
  }
}
& \multicolumn{ 3 }{c}{
  \makecell{
  \textsc{\small Expressions }
\\(259 images)
  }
}
& \multicolumn{ 3 }{c|}{
  \makecell{
  \textsc{\small Peel Banana }
\\(513 images)
  }
}
& \multicolumn{ 3 }{c}{
  \makecell{
  \textsc{\small Mean }
  }
}
\\

& \multicolumn{1}{c}{ \footnotesize PSNR$\uparrow$ }
& \multicolumn{1}{c}{ \footnotesize MS-SSIM$\uparrow$ }
& \multicolumn{1}{c}{ \footnotesize LPIPS$\downarrow$ }
& \multicolumn{1}{c}{ \footnotesize PSNR$\uparrow$ }
& \multicolumn{1}{c}{ \footnotesize MS-SSIM$\uparrow$ }
& \multicolumn{1}{c}{ \footnotesize LPIPS$\downarrow$ }
& \multicolumn{1}{c}{ \footnotesize PSNR$\uparrow$ }
& \multicolumn{1}{c}{ \footnotesize MS-SSIM$\uparrow$ }
& \multicolumn{1}{c}{ \footnotesize LPIPS$\downarrow$ }
& \multicolumn{1}{c}{ \footnotesize PSNR$\uparrow$ }
& \multicolumn{1}{c}{ \footnotesize MS-SSIM$\uparrow$ }
& \multicolumn{1}{c}{ \footnotesize LPIPS$\downarrow$ }
& \multicolumn{1}{c}{ \footnotesize PSNR$\uparrow$ }
& \multicolumn{1}{c}{ \footnotesize MS-SSIM$\uparrow$ }
& \multicolumn{1}{c|}{ \footnotesize LPIPS$\downarrow$ }
& \multicolumn{1}{c}{ \footnotesize PSNR$\uparrow$ }
& \multicolumn{1}{c}{ \footnotesize MS-SSIM$\uparrow$ }
& \multicolumn{1}{c}{ \footnotesize LPIPS$\downarrow$ }
\\
\hline

  NeRF~\cite{mildenhall2020nerf}
  &\tablethird$19.9$
  &\tablethird$.653$
  &$.692$
  
  &\tablethird$20.7$
  &$.780$
  &$.357$
  
  &$19.9$
  &$.777$
  &$.325$
  
  &$20.1$
  &$.697$
  &$.394$
  
  &$20.0$
  &$.769$
  &$.352$
  
  &$20.1$
  &$.735$
  &$.424$
  
  \\   NV~\cite{lombardi2019neural}
  &$17.7$
  &$.623$
  &$.360$
  
  &$16.2$
  &$.665$
  &$.330$
  
  &$17.6$
  &$.615$
  &$.336$
  
  &$14.6$
  &$.672$
  &$.276$
  
  &$15.9$
  &$.380$
  &$.413$
  
  &$16.4$
  &$.591$
  &$.343$
  
  \\   NSFF~\cite{li2020nsff}\textsuperscript{\textdagger}
  &\tablefirst$26.1$
  &\tablefirst$.871$
  &\tablesecond$.284$
  
  &\tablefirst$27.7$
  &\tablefirst$.947$
  &$.125$
  
  &$26.9$
  &$.944$
  &$.106$
  
  &\tablefirst$26.7$
  &\tablefirst$.922$
  &$.157$
  
  &\tablefirst$24.6$
  &\tablesecond$.902$
  &$.198$
  
  &\tablefirst$26.4$
  &\tablefirst$.917$
  &$.174$
  
  \\   Nerfies~\cite{park2020nerfies}
  &$19.2$
  &$.567$
  &$.325$
  
  &$20.6$
  &\tablethird$.830$
  &\tablefirst$.108$
  
  &$26.7$
  &$.943$
  &\tablesecond$.0777$
  
  &$21.8$
  &$.802$
  &$.150$
  
  &$22.4$
  &$.872$
  &$.147$
  
  &$22.1$
  &$.803$
  &\tablethird$.162$
  
  \\   Nerfies (w/o elastic)
  &$19.4$
  &$.581$
  &$.323$
  
  &$20.2$
  &$.820$
  &$.115$
  
  &$26.0$
  &$.935$
  &$.0837$
  
  &$21.8$
  &$.800$
  &$.149$
  
  &$21.7$
  &$.852$
  &$.157$
  
  &$21.8$
  &$.798$
  &$.165$
  
  \\ \hline  Hyper-NeRF (DS)
  &$19.3$
  &$.591$
  &\tablethird$.296$
  
  &$20.0$
  &$.821$
  &\tablethird$.111$
  
  &$26.9$
  &$.948$
  &\tablethird$.0787$
  
  &$21.6$
  &$.800$
  &\tablesecond$.148$
  
  &\tablethird$23.3$
  &\tablethird$.896$
  &\tablefirst$.133$
  
  &$22.2$
  &$.811$
  &\tablesecond$.153$
  
  \\   Hyper-NeRF (DS, w/ elastic)
  &$19.5$
  &$.605$
  &\tablefirst$.277$
  
  &$20.2$
  &$.823$
  &\tablesecond$.109$
  
  &\tablesecond$27.5$
  &\tablefirst$.954$
  &\tablefirst$.0756$
  
  &$21.9$
  &\tablethird$.806$
  &\tablefirst$.144$
  
  &$22.7$
  &$.882$
  &\tablesecond$.133$
  
  &\tablethird$22.3$
  &\tablethird$.814$
  &\tablefirst$.148$
  
  \\   Hyper-NeRF (AP)
  &$19.6$
  &$.596$
  &$.319$
  
  &$20.0$
  &$.814$
  &$.131$
  
  &\tablethird$27.2$
  &\tablethird$.950$
  &$.0941$
  
  &\tablesecond$22.2$
  &\tablesecond$.817$
  &\tablethird$.149$
  
  &$22.4$
  &$.874$
  &\tablethird$.142$
  
  &$22.2$
  &$.810$
  &$.167$
  
  \\   Hyper-NeRF (w/o deform)
  &\tablesecond$20.6$
  &\tablesecond$.714$
  &$.613$
  
  &\tablesecond$21.4$
  &\tablesecond$.846$
  &$.212$
  
  &\tablefirst$27.6$
  &\tablesecond$.950$
  &$.108$
  
  &\tablethird$22.0$
  &$.793$
  &$.196$
  
  &\tablesecond$24.3$
  &\tablefirst$.914$
  &$.170$
  
  &\tablesecond$23.2$
  &\tablesecond$.843$
  &$.260$
  
  \\ \bottomrule

\end{tabular}

}


\end{table*}

\definecolor{bestcolor}{rgb}{0.85, 0.113, 0.188}

\renewcommand{\textimage}[4]{
	\begin{overpic}[width=2.05cm,unit=1mm,clip,trim=#1]{images/vrig_images_v2/#2}
	\put (9.5,0.7) {\sethlcolor{white}\footnotesize\hl{$#3 / #4$}}
    \end{overpic}
}

\newcommand{\textimagew}[4]{
	\begin{overpic}[width=2.05cm,unit=1mm,clip,trim=#1]{images/vrig_images_v2/#2}
	\put (8.5,0.7) {\sethlcolor{white}\footnotesize\hl{$#3 / #4$}}
    \end{overpic}
}

\renewcommand{\animage}[2]{\includegraphics[width=2.05cm,clip,trim=#1]{images/vrig_images_v2/#2}}
\renewcommand{\bestnum}[1]{\textcolor{bestcolor}{#1}}

\begin{figure*}[t!]
    \setlength{\tabcolsep}{0.5pt}
    \renewcommand{\arraystretch}{0.25}
    \begin{tabular}{cccccccccc}
                    
        \makebox[20pt]{\raisebox{40pt}{\rotatebox[origin=c]{90}{\textsc{Expressions}}}} \hspace{-7pt} &
        \animage{20 70 30 100}{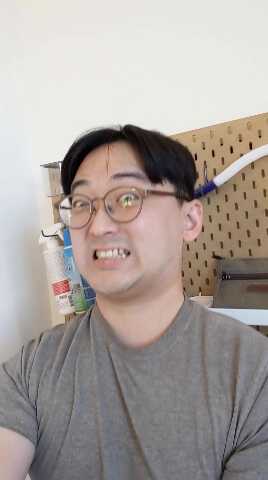} &
        \animage{20 70 30 100}{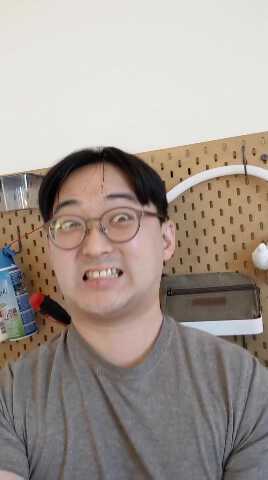} &
        \textimage{20 70 30 100}{vrig-kexp.hypernerf-bs-2d-elastic.pred.jpg}{22.0}{\bestnum{0.106}} &
        \textimage{20 70 30 100}{vrig-kexp.nerfies.pred.jpg}{21.9}{0.113} &
        \textimage{20 70 30 100}{vrig-kexp.no-deform.pred.jpg}{20.5}{0.189} &
        \textimage{20 70 30 100}{vrig-kexp.nsff.pred.jpg}{\bestnum{27.7}}{0.150} &
        \textimage{20 70 30 100}{vrig-kexp.neuralvolumes.pred.jpg}{15.1}{0.261} &
        \textimage{20 70 30 100}{vrig-kexp.nerf.pred.jpg}{19.5}{0.445} \\
         
        \makebox[20pt]{\raisebox{40pt}{\rotatebox[origin=c]{90}{\textsc{Broom}}}} \hspace{-7pt} &
        \animage{10 130 10 50}{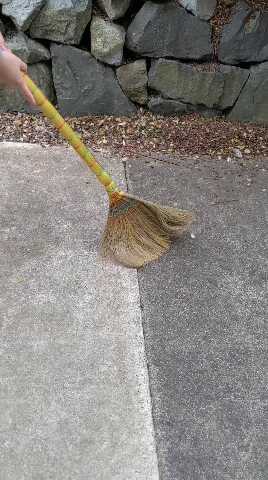} &
        \animage{10 130 10 50}{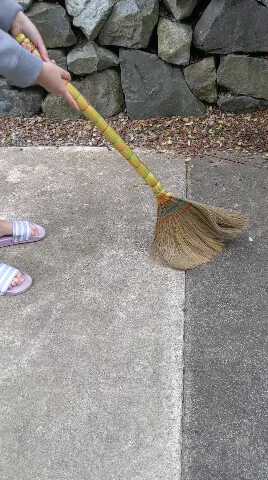} &
        \textimage{10 130 10 50}{broom2.hypernerf-bs-2d-elastic.pred.jpg}{20.0}{\bestnum{0.214}} &
        \textimage{10 130 10 50}{broom2.nerfies.pred.jpg}{20.9}{0.237} &
        \textimage{10 130 10 50}{broom2.no-deform.pred.jpg}{{21.4}}{0.563} &
        \textimage{10 130 10 50}{broom2.nsff.pred.jpg}{\bestnum{25.4}}{0.269} &
        \textimage{10 130 10 50}{broom2.neuralvolumes.pred.jpg}{18.8}{0.335} &
        \textimage{10 130 10 50}{broom2.nerf.pred.jpg}{20.7}{0.661} \\
        
        \makebox[20pt]{\raisebox{40pt}{\rotatebox[origin=c]{90}{\textsc{Peel Banana}}}} \hspace{-7pt} &
        \animage{12 100 12 55}{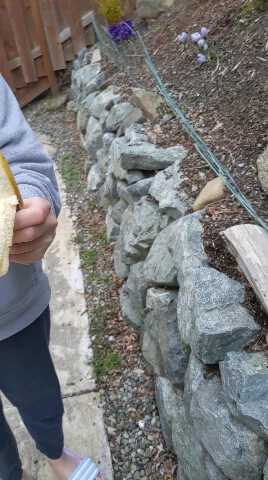} &
        \animage{12 100 12 55}{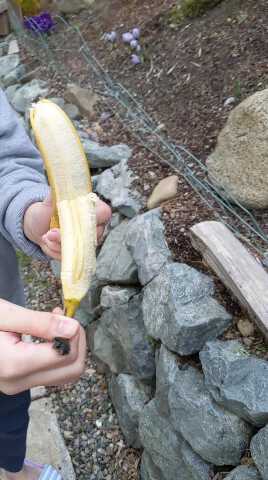} &
        \textimage{12 100 12 55}{vrig-peel-banana.hypernerf-bs-2d.pred.jpg}{{20.8}}{\bestnum{0.127}} &
        \textimage{12 100 12 55}{vrig-peel-banana.nerfies-elastic.pred.jpg}{20.3}{0.146} &
        \textimage{12 100 12 55}{vrig-peel-banana.no-deform.pred.jpg}{17.9}{0.192} &
        \textimage{12 100 12 55}{vrig-peel-banana.nsff.pred.jpg}{\bestnum{23.3}}{0.192} &
        \textimage{12 100 12 55}{vrig-peel-banana.neuralvolumes.pred.jpg}{15.0}{0.484} &
        \textimage{12 100 12 55}{vrig-peel-banana.nerf.pred.jpg}{18.4}{0.247} \\
        
        \makebox[20pt]{\raisebox{40pt}{\rotatebox[origin=c]{90}{\textsc{Chicken}}}} \hspace{-7pt} &
        \animage{12 150 30 50}{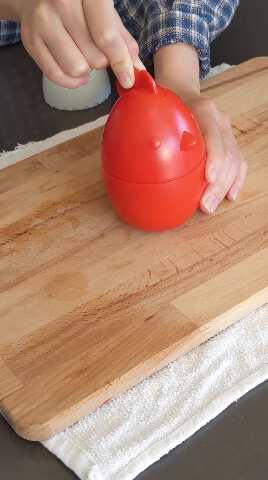} &
        \animage{12 150 30 50}{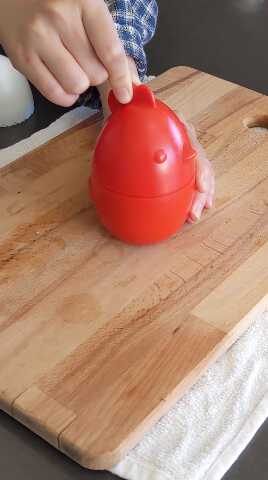} &
        \textimagew{12 150 30 50}{vrig-chicken.hypernerf-bs-2d-elastic.pred.jpg}{{28.7}}{\bestnum{0.0631}} &
        \textimagew{12 150 30 50}{vrig-chicken.nerfies-elastic.pred.jpg}{26.9}{0.0677} &
        \textimagew{12 150 30 50}{vrig-chicken.no-deform.pred.jpg}{\bestnum{29.1}}{0.0870} &
        \textimage{12 150 30 50}{vrig-chicken.nsff.pred.jpg}{25.3}{0.106} &
        \textimage{12 150 30 50}{vrig-chicken.neuralvolumes.pred.jpg}{18.1}{0.344} &
        \textimage{12 150 30 50}{vrig-chicken.nerf.pred.jpg}{18.9}{0.306} \\

        \makebox[20pt]{\raisebox{40pt}{\rotatebox[origin=c]{90}{\textsc{3D Printer}}}} \hspace{-7pt} &
        \animage{15 120 15 0}{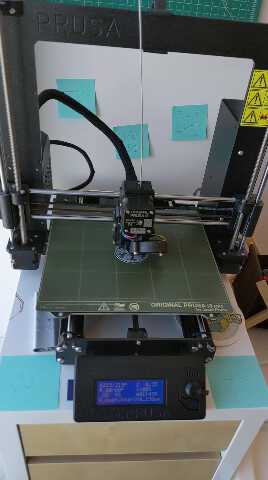} &
        \animage{15 120 15 0}{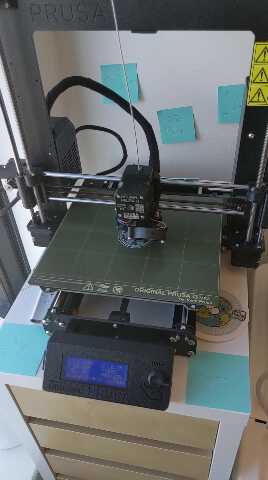} &
        \textimage{15 120 15 0}{vrig-3dprinter.hypernerf-bs-2d.pred.jpg}{21.1}{\bestnum{0.118}} &
        \textimage{15 120 15 0}{vrig-3dprinter.nerfies-elastic.pred.jpg}{21.5}{0.119} &
        \textimage{15 120 15 0}{vrig-3dprinter.no-deform.pred.jpg}{23.7}{0.172} &
        \textimage{15 120 15 0}{vrig-3dprinter.nsff.pred.jpg}{\bestnum{28.9}}{{0.122}} &
        \textimage{15 120 15 0}{vrig-3dprinter.neuralvolumes.pred.jpg}{14.2}{0.432} &
        \textimage{15 120 15 0}{vrig-3dprinter.nerf.pred.jpg}{{24.3}}{0.286} \\
        
        \\[2pt]
        
        & {\small Training view} & 
        {\small Novel view (GT)} & 
        {\small HyperNeRF (Ours)} &
        {\small Nerfies} &
        {\small No Deform} &
        {\small NSFF} &
        {\small Neural Volumes} &
        {\small NeRF}
        \\
    \end{tabular}
    \vspace{-6pt}
    \caption{Comparisons of baselines and our method on validation rig scenes. {\footnotesize PSNR / LPIPS} metrics on bottom right with best colored \bestnum{red}. Baselines shown are: Nerfies~\cite{park2020nerfies}, HyperNeRF without deformations, Neural Volumes~\cite{lombardi2019neural}, NSFF~\cite{li2020nsff}, and NeRF~\cite{mildenhall2020nerf}. Note how PSNR often prefers blurry results.}
    \label{fig:vrig_results}
    \vspace{-12pt}
\end{figure*}

\definecolor{bestcolor}{rgb}{0.85, 0.113, 0.188}

\renewcommand{\animage}[2]{\includegraphics[width=1.65cm,clip,trim=#1]{images/interp_images_v3/#2}}

\begin{figure*}[t!]
    \setlength{\tabcolsep}{0.5pt}
    \renewcommand{\arraystretch}{0.25}
    \begin{tabular}{ccccccccccc}
            
        \multirow{2}{*}[0.35in]{\makebox[20pt]{\raisebox{25pt}{\rotatebox[origin=c]{90}{\textsc{Expressions 1}}}} \hspace{-7pt}} &
        \animage{30 120 30 60}{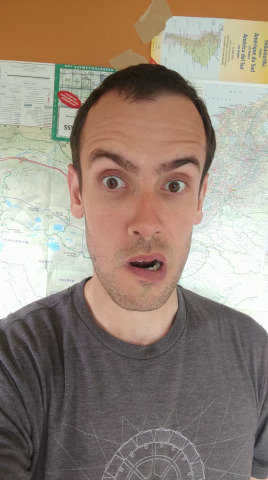} &
        \animage{30 120 30 60}{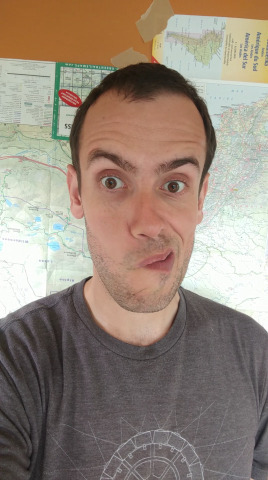} &
        \animage{30 120 30 60}{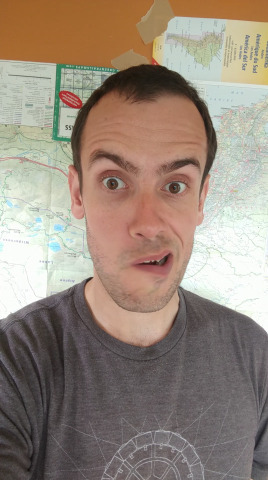} &
        \animage{30 120 30 60}{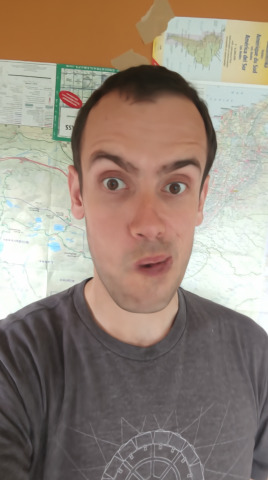} &
        \animage{30 120 30 60}{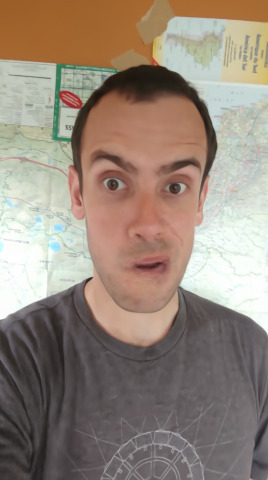} &
        \animage{30 120 30 60}{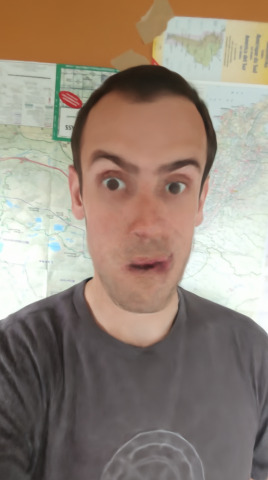} &
        \animage{30 120 30 60}{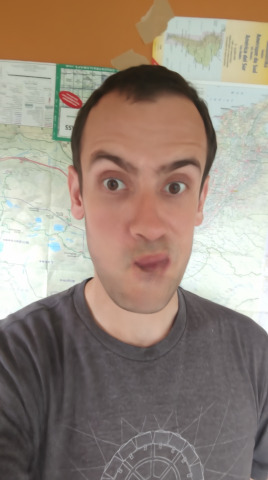} &
        \animage{30 120 30 60}{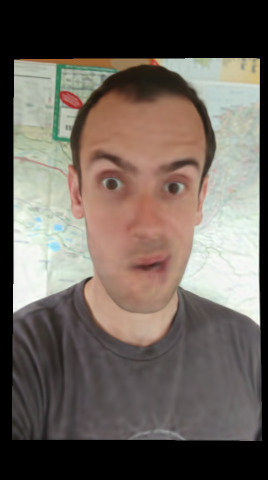} &
        \animage{30 120 30 58}{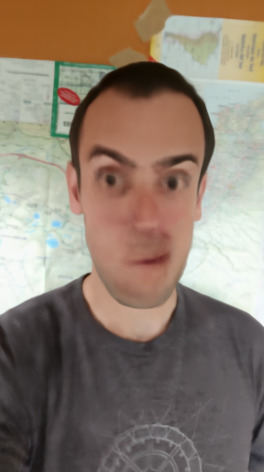} &
        \animage{30 120 30 60}{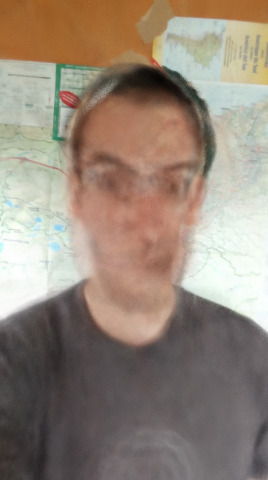} \\
        & & & &
        \animage{30 120 30 60}{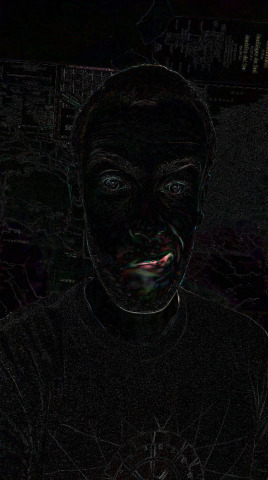} &
        \animage{30 120 30 60}{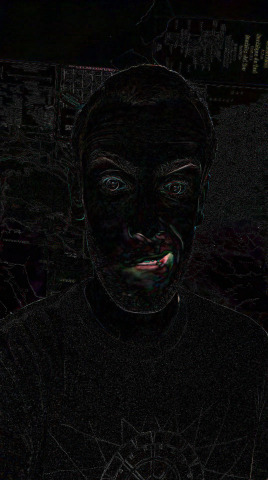} &
        \animage{30 120 30 60}{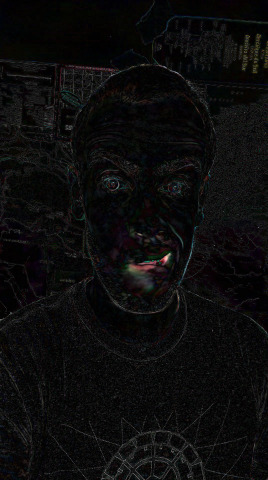} &
        \animage{30 120 30 60}{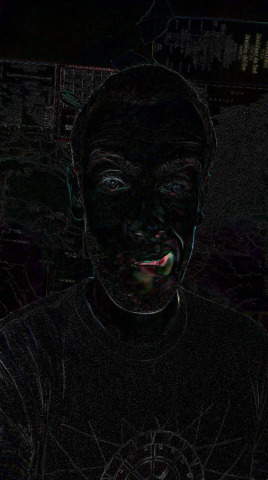} &
        \animage{30 120 30 60}{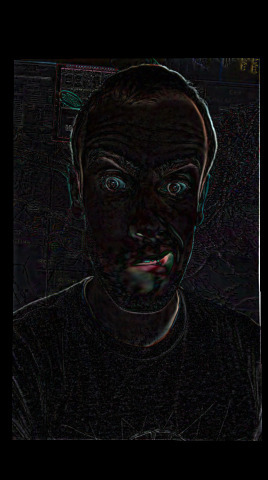} &
        \animage{30 120 30 58}{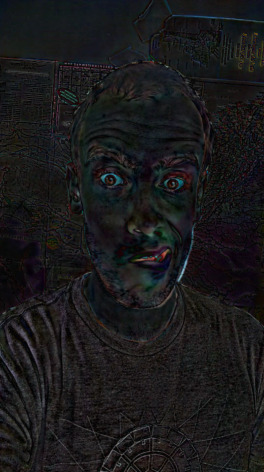} &
        \animage{30 120 30 60}{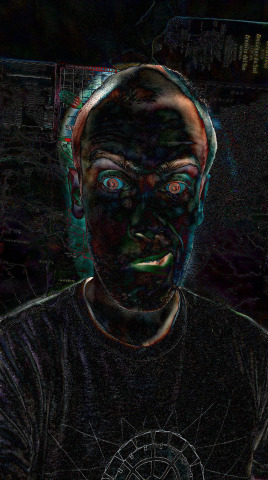} \\
        
        \multirow{2}{*}[0.35in]{\makebox[20pt]{\raisebox{25pt}{\rotatebox[origin=c]{90}{\textsc{Expressions 2}}}} \hspace{-7pt}} &
        \animage{75 140 40 110}{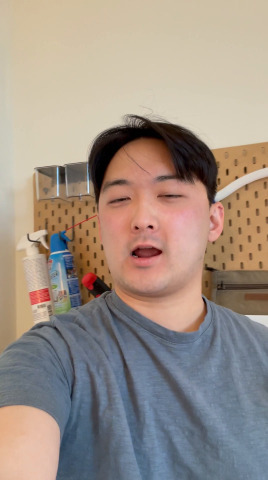} &
        \animage{75 140 40 110}{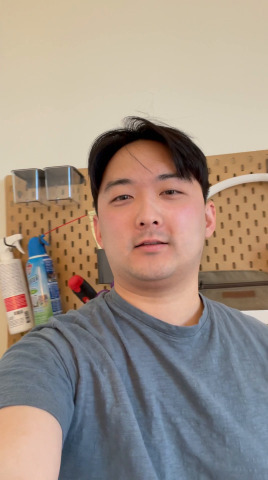} &
        \animage{75 140 40 110}{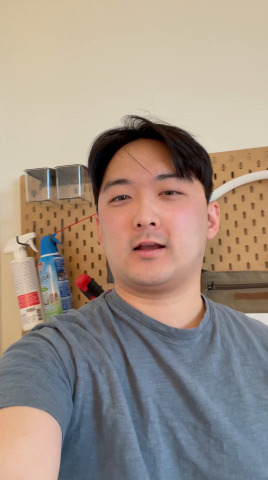} &
        \animage{75 140 40 110}{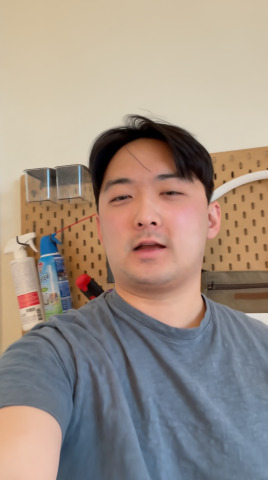} &
        \animage{75 140 40 110}{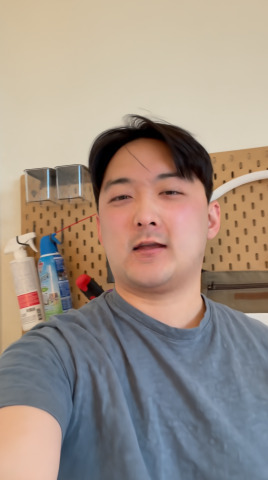} &
        \animage{75 140 40 110}{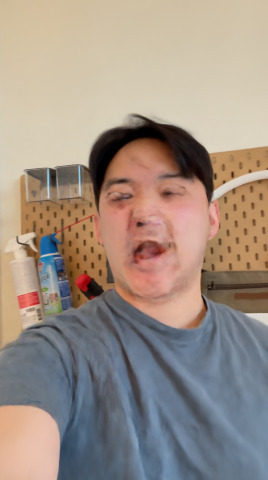} &
        \animage{75 140 40 110}{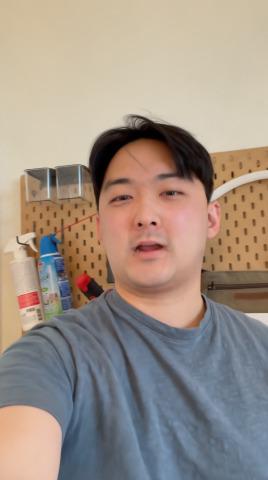} &
        \animage{75 140 40 110}{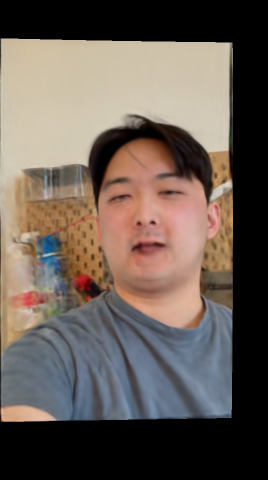} &
        \animage{67 128 36 102}{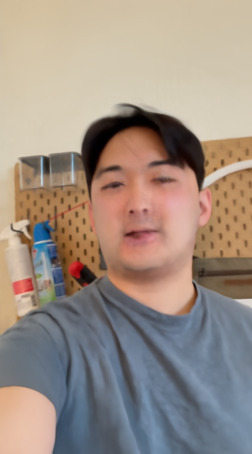} &
        \animage{75 140 40 110}{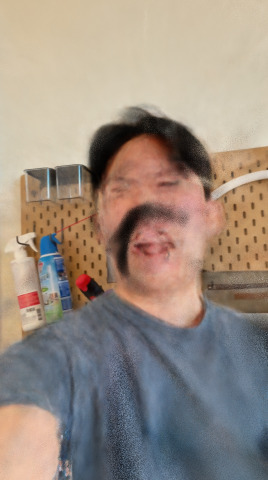} \\
        & & & &
        \animage{75 140 40 110}{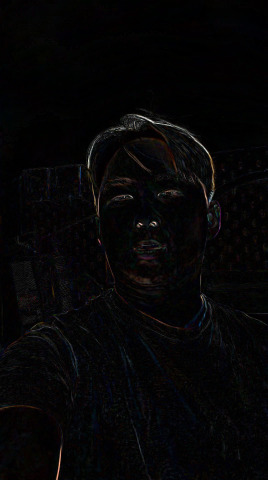} &
        \animage{75 140 40 110}{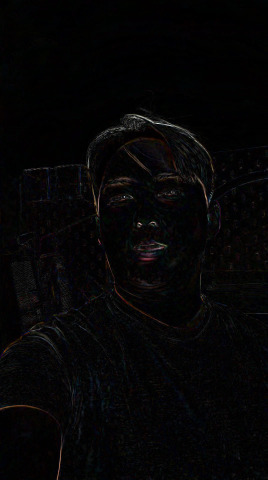} &
        \animage{75 140 40 110}{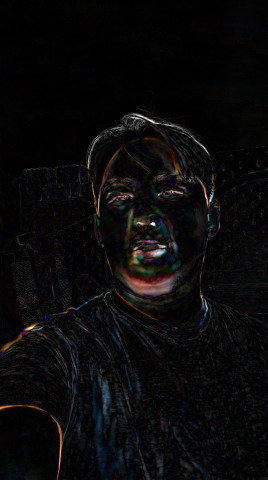} &
        \animage{75 140 40 110}{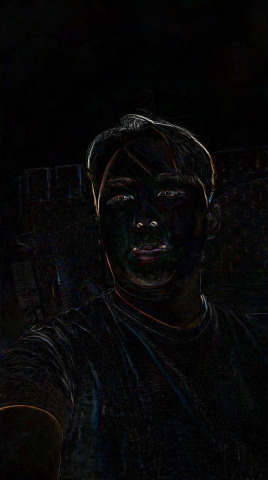} &
        \animage{75 140 40 110}{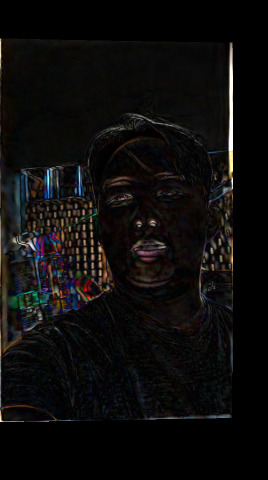} &
        \animage{67 128 36 102}{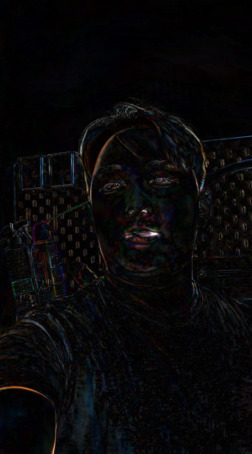} &
        \animage{75 140 40 110}{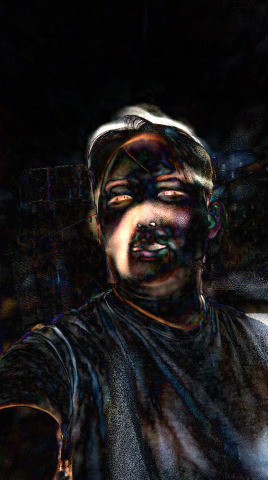} \\

        \multirow{2}{*}[0.35in]{\makebox[20pt]{\raisebox{25pt}{\rotatebox[origin=c]{90}{\textsc{Expressions 3}}}} \hspace{-7pt}} &
        \animage{25 80 30 81}{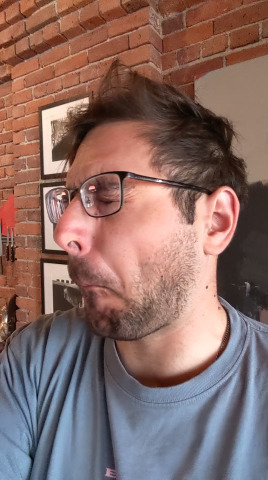} &
        \animage{25 80 30 81}{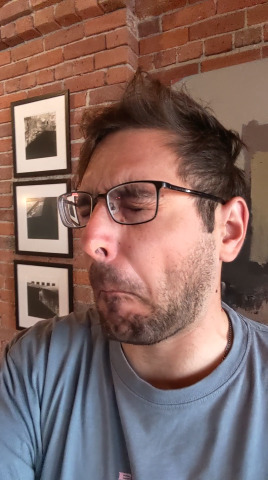} &
        \animage{25 80 30 81}{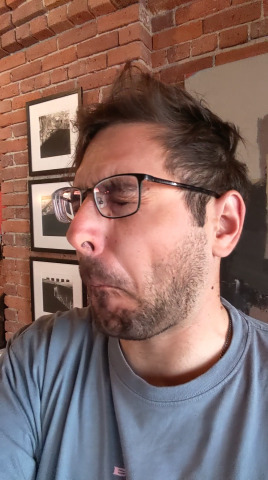} &
        \animage{25 80 30 81}{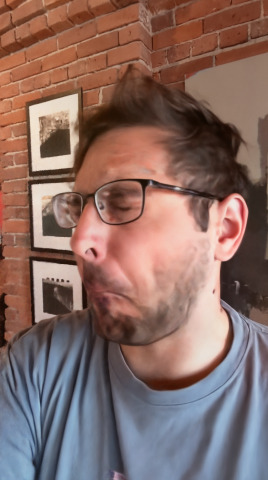} &
        \animage{25 80 30 81}{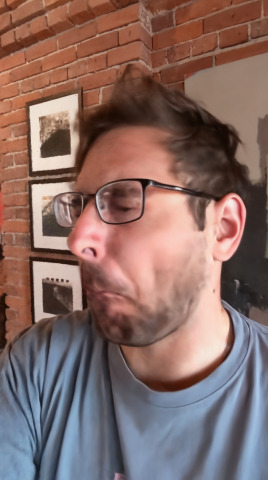} &
        \animage{25 80 30 81}{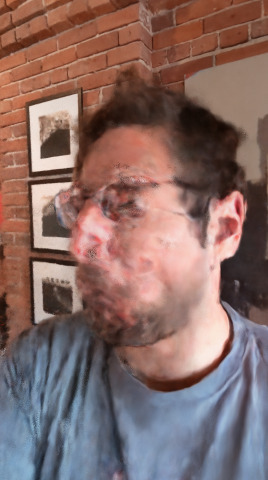} &
        \animage{25 80 30 81}{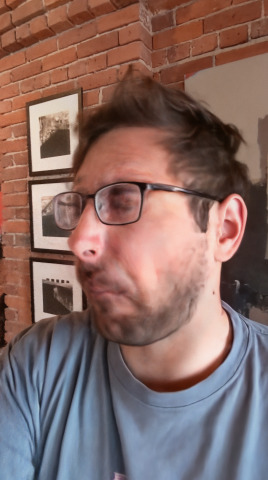} &
        \animage{25 80 30 81}{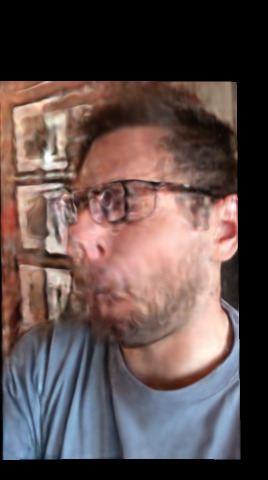} &
        \animage{25 80 30 81}{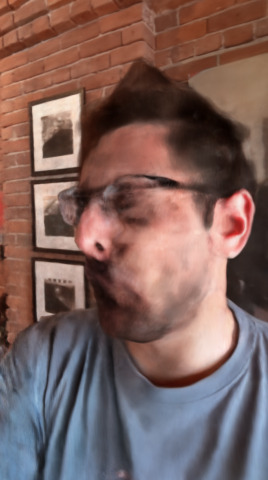} &
        \animage{25 80 30 81}{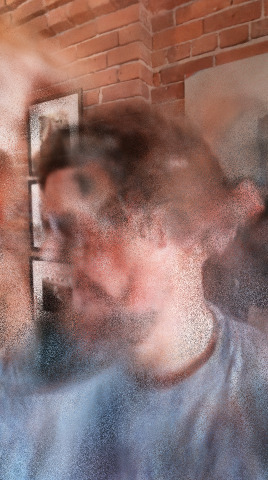} \\
        & & & &
        \animage{25 80 30 81}{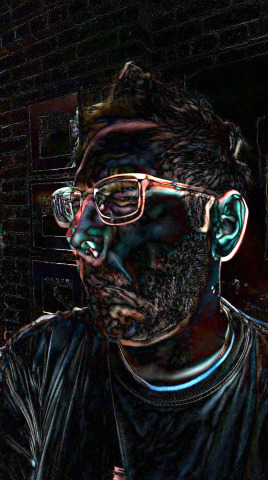} &
        \animage{25 80 30 81}{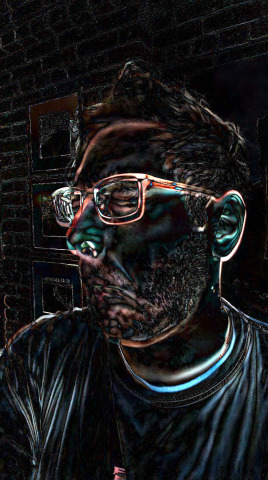} &
        \animage{25 80 30 81}{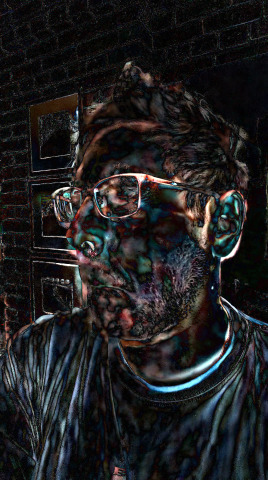} &
        \animage{25 80 30 81}{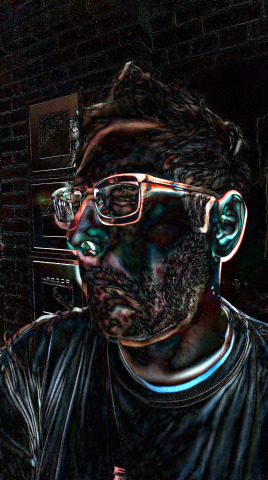} &
        \animage{25 80 30 81}{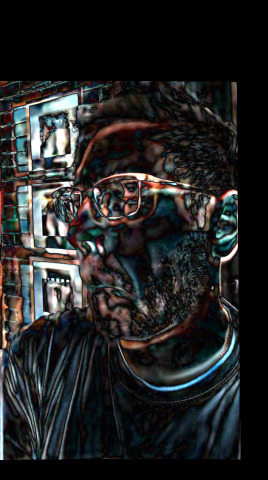} &
        \animage{25 80 30 81}{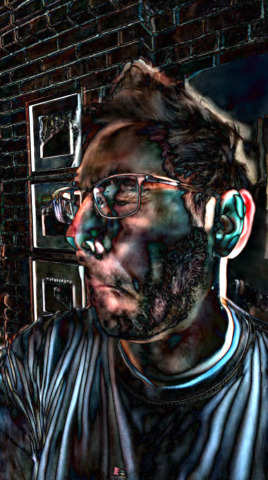} &
        \animage{25 80 30 81}{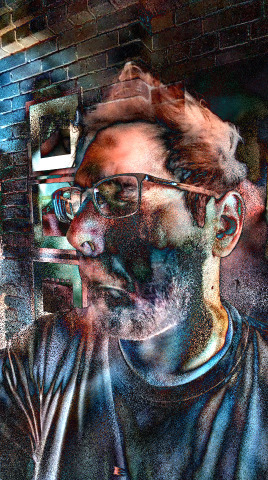} \\
        
        \multirow{2}{*}[0.2in]{\makebox[10pt]{\raisebox{25pt}{\rotatebox[origin=c]{90}{\textsc{Teapots}}}} \hspace{-7pt}} &
        \animage{40 65 80 200}{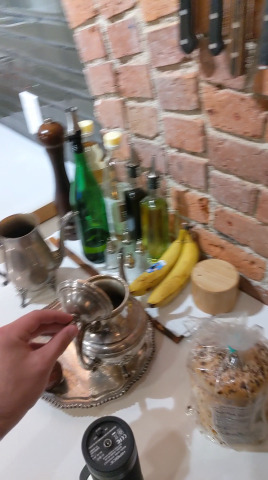} &
        \animage{40 65 80 200}{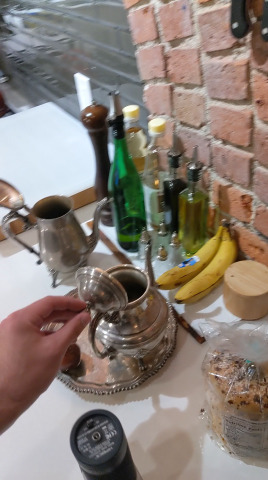} &
        \animage{40 65 80 200}{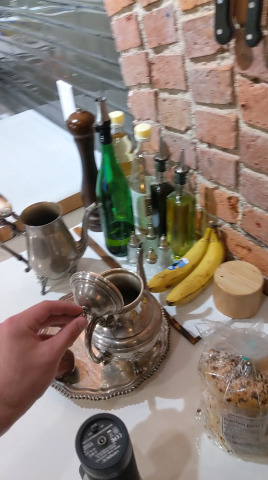} &
        \animage{40 65 80 200}{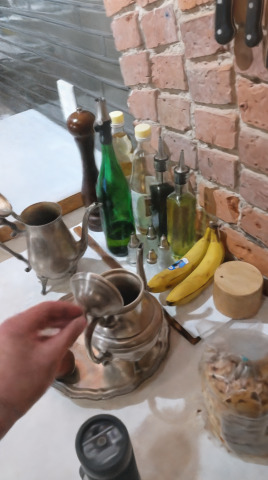} &
        \animage{40 65 80 200}{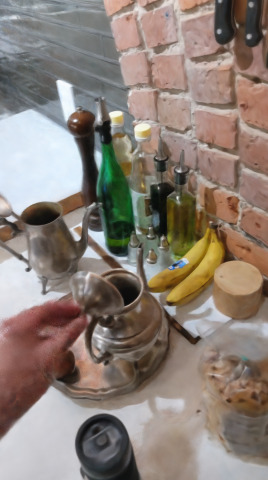} &
        \animage{40 65 80 200}{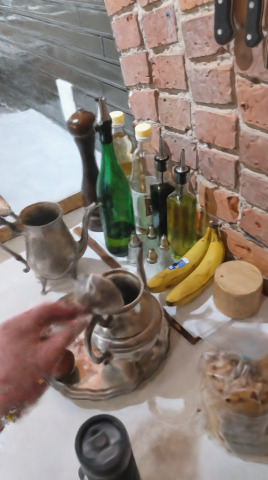} &
        \animage{40 65 80 200}{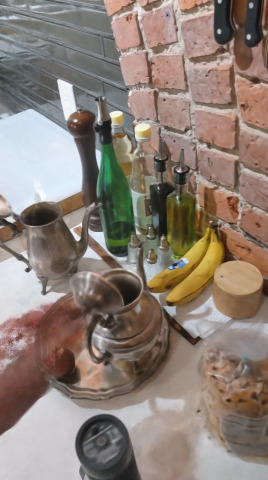} &
        \animage{40 65 80 200}{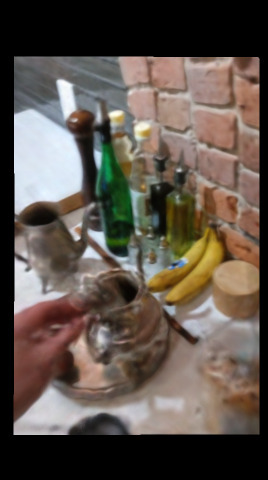} &
        \animage{40 65 80 200}{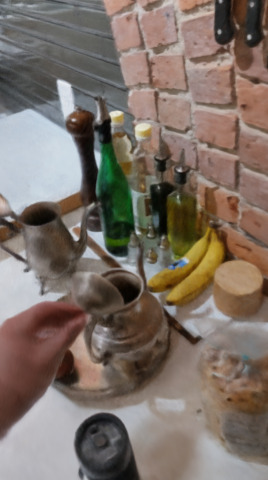} &
        \animage{40 65 80 200}{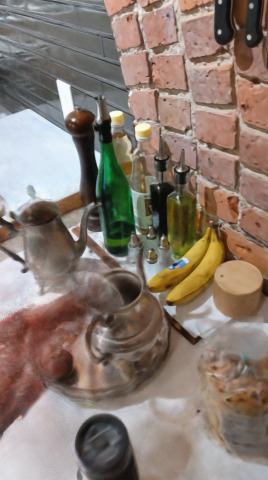} \\
        & & & &
        \animage{40 65 80 200}{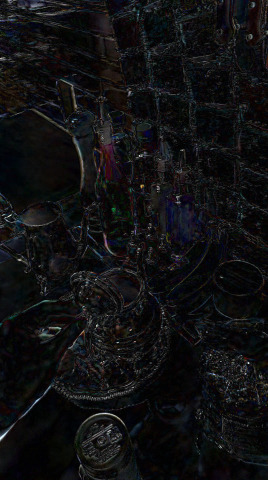} &
        \animage{40 65 80 200}{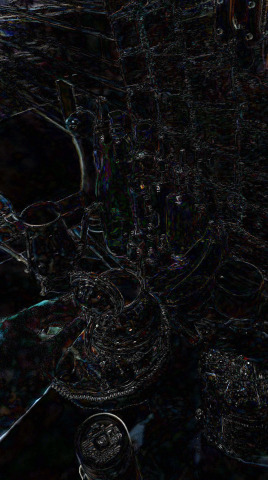} &
        \animage{40 65 80 200}{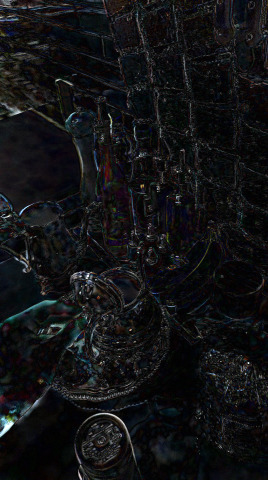} &
        \animage{40 65 80 200}{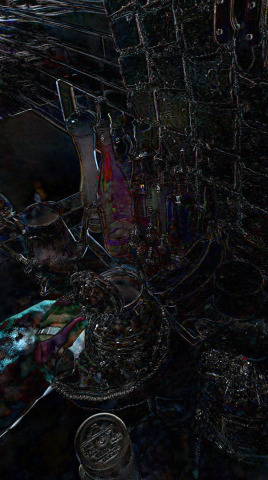} &
        \animage{40 65 80 200}{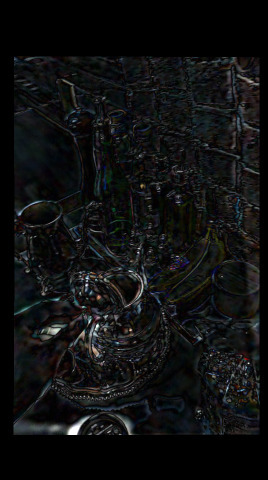} &
        \animage{40 65 80 200}{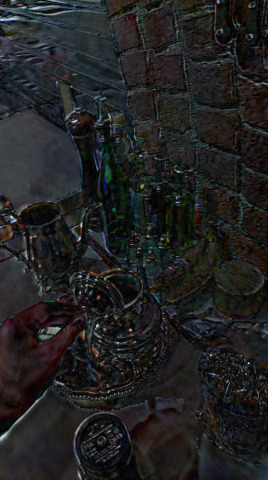} &
        \animage{40 65 80 200}{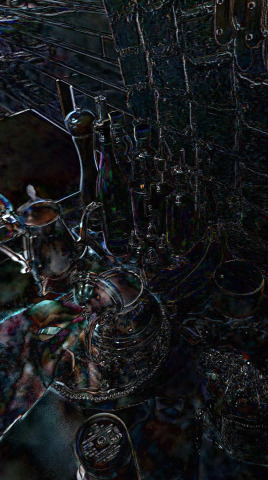} \\
        
        \\[2pt]
        
        & 
        {\smaller Reference 1} & 
        {\smaller Reference 2} & 
        {\smaller \makecell{Interpolated\\(GT)}} & 
        {\smaller \makecell{HyperNeRF\\(DS)}} &
        {\smaller \makecell{HyperNeRF\\(AP)}} &
        {\smaller No Deform} &
        {\smaller Nerfies} &
        {\smaller \makecell{Neural\\Volumes}} &
        {\smaller NSFF} &
        {\smaller NeRF}
        \\
    \end{tabular}
    \vspace{-6pt}
    \caption{Qualitative comparisons of our method, ablations, and baselines on the interpolation dataset (See \secref{sec:evaluation_quantitative}). The interpolated frame is rendered by linearly interpolating between the latent codes of the two reference frames. The bottom row of each sequence shows the absolute error. Baselines shown are: Nerfies~\cite{park2020nerfies}, Neural Volumes~\cite{lombardi2019neural}, NSFF~\cite{li2020nsff}, and NeRF~\cite{mildenhall2020nerf}.}
    \label{fig:interp_results1}
    \vspace{-20pt}
\end{figure*}

\definecolor{bestcolor}{rgb}{0.85, 0.113, 0.188}

\renewcommand{\animage}[2]{\includegraphics[width=1.65cm,clip,trim=#1]{images/interp_images_v3/#2}}

\begin{figure*}[t!]
    \setlength{\tabcolsep}{0.5pt}
    \renewcommand{\arraystretch}{0.25}
    \begin{tabular}{ccccccccccc}
            
        \multirow{2}{*}[0.2in]{\makebox[20pt]{\raisebox{25pt}{\rotatebox[origin=c]{90}{\textsc{Chicken}}}} \hspace{-7pt}} &
        \animage{45 130 45 90}{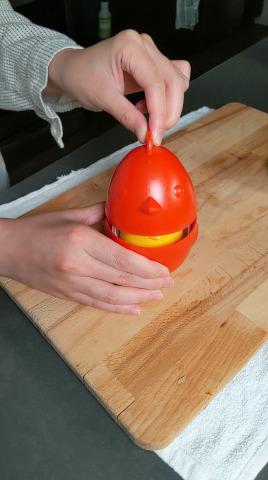} &
        \animage{45 130 45 90}{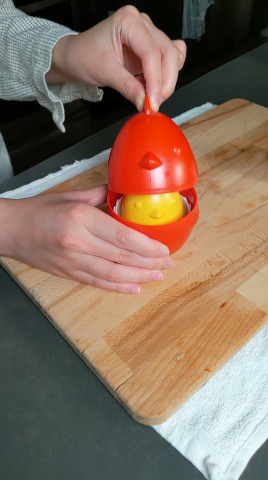} &
        \animage{45 130 45 90}{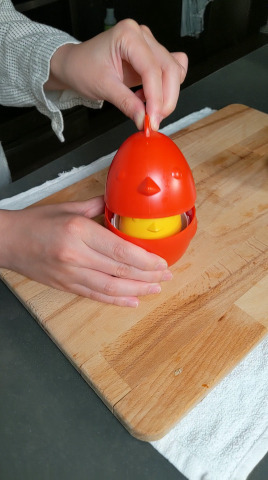} &
        \animage{45 130 45 90}{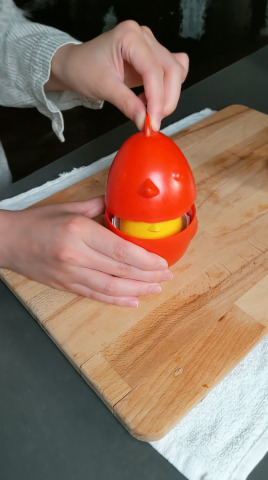} &
        \animage{45 130 45 90}{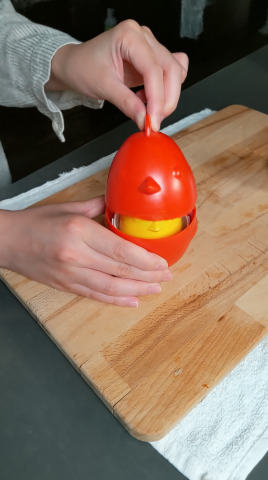} &
        \animage{45 130 45 90}{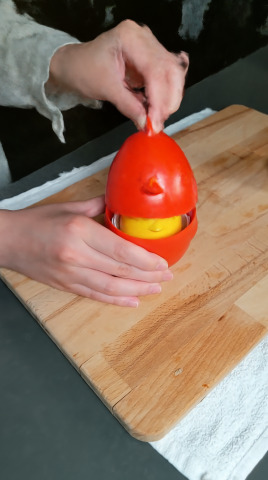} &
        \animage{45 130 45 90}{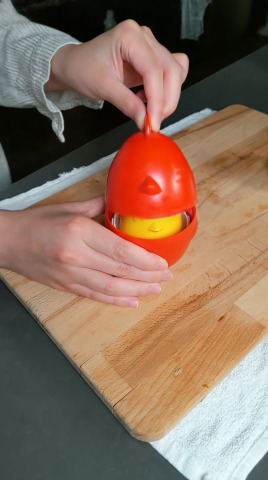} &
        \animage{45 130 45 90}{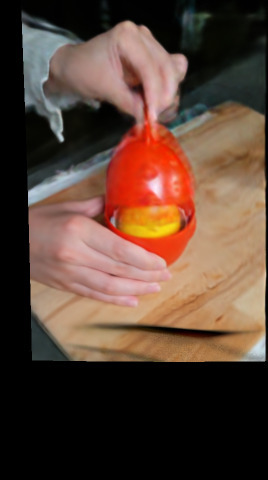} &
        \animage{45 130 45 94}{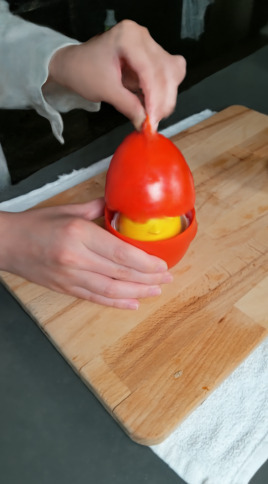} &
        \animage{45 130 45 90}{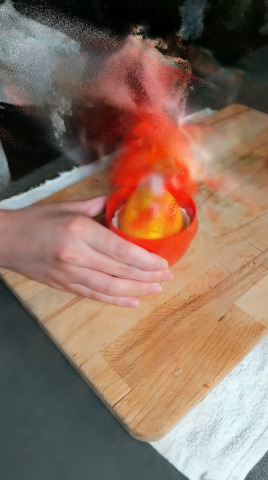} \\
        & & & &
        \animage{45 130 45 90}{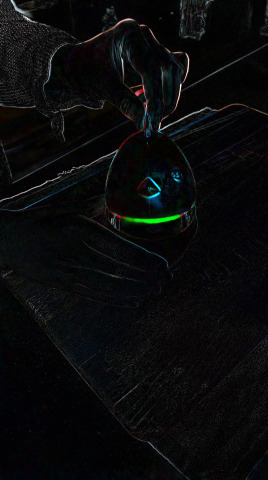} &
        \animage{45 130 45 90}{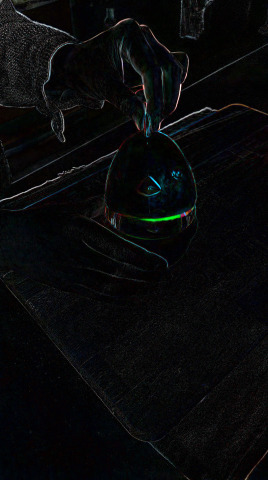} &
        \animage{45 130 45 90}{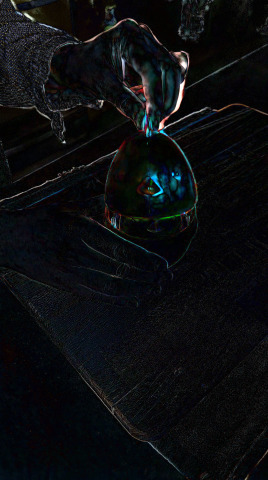} &
        \animage{45 130 45 90}{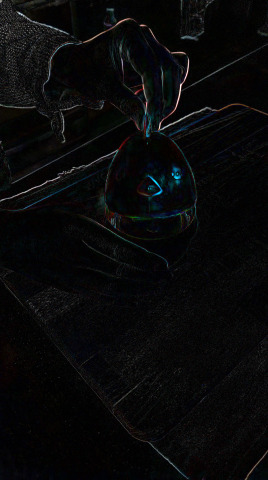} &
        \animage{45 130 45 90}{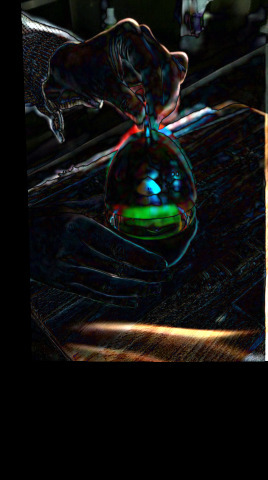} &
        \animage{45 130 45 94}{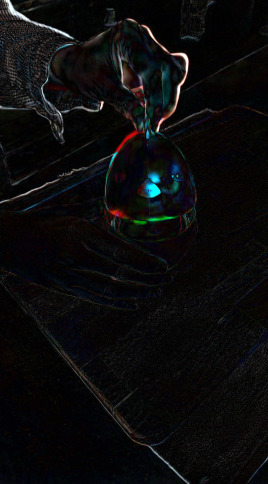} &
        \animage{45 130 45 90}{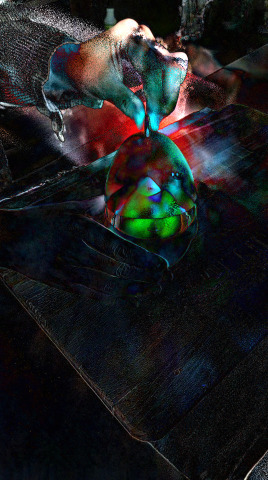} \\
        
        \multirow{2}{*}[0.1in]{\makebox[20pt]{\raisebox{25pt}{\rotatebox[origin=c]{90}{\textsc{Fist}}}} \hspace{-7pt}} &
        \animage{20 150 80 120}{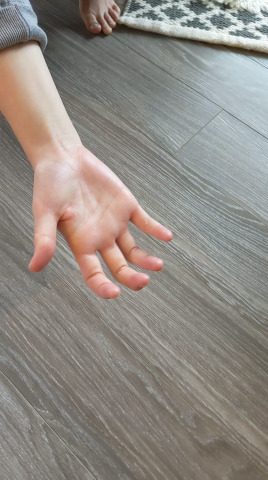} &
        \animage{20 150 80 120}{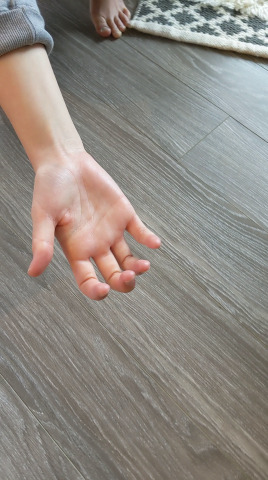} &
        \animage{20 150 80 120}{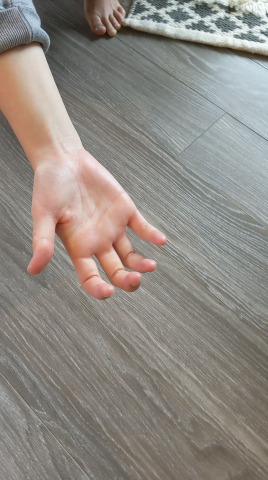} &
        \animage{20 150 80 120}{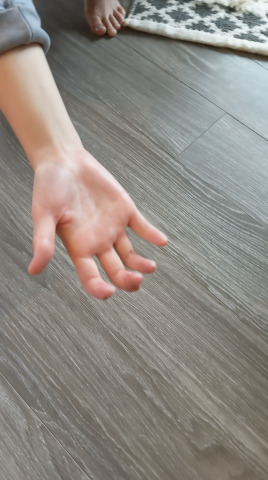} &
        \animage{20 150 80 120}{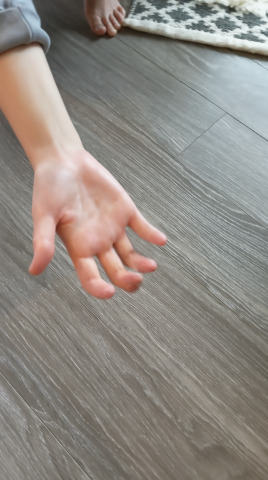} &
        \animage{20 150 80 120}{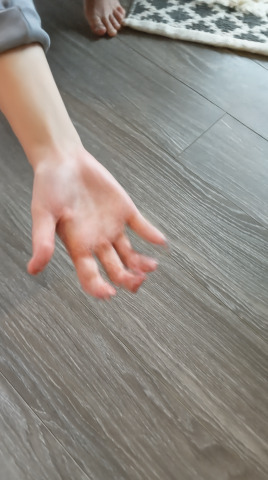} &
        \animage{20 150 80 120}{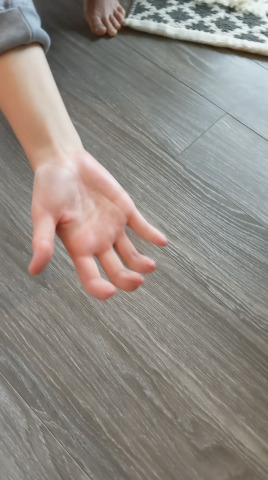} &
        \animage{20 150 80 120}{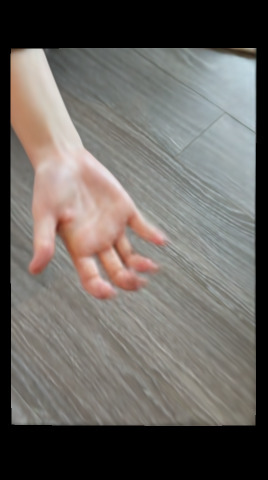} &
        \animage{20 150 80 124}{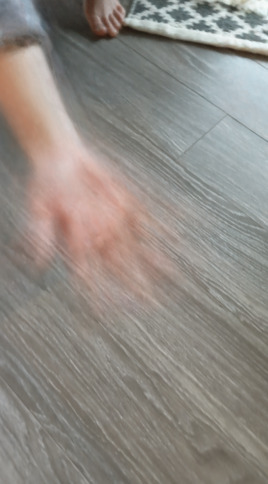} &
        \animage{20 150 80 120}{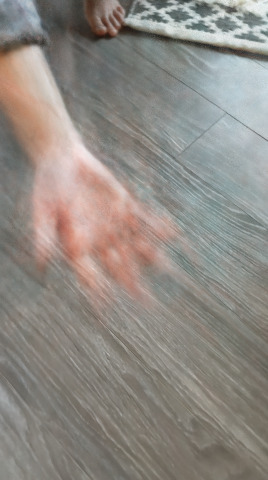} \\
        & & & &
        \animage{20 150 80 120}{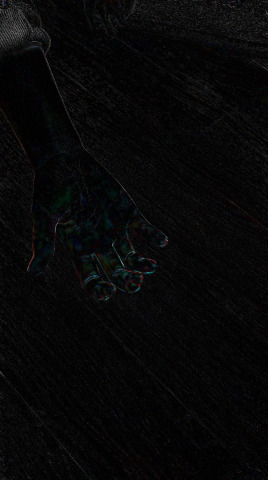} &
        \animage{20 150 80 120}{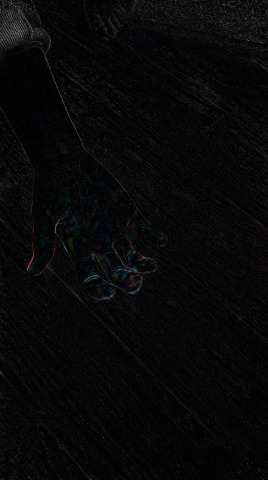} &
        \animage{20 150 80 120}{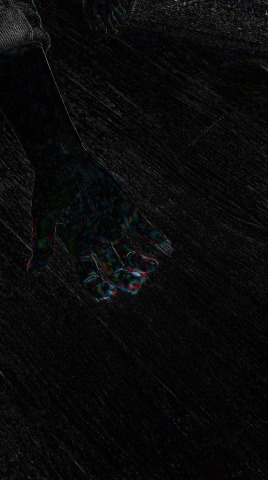} &
        \animage{20 150 80 120}{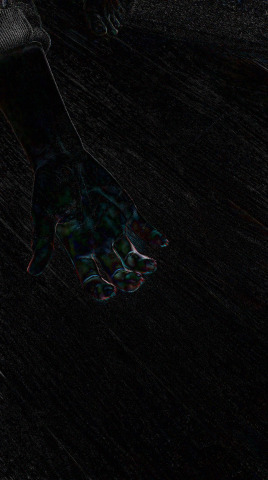} &
        \animage{20 150 80 120}{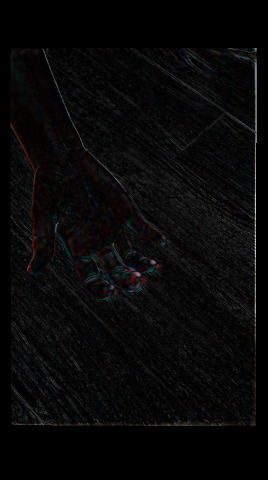} &
        \animage{20 150 80 124}{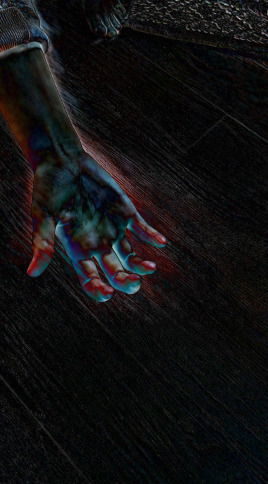} &
        \animage{20 150 80 120}{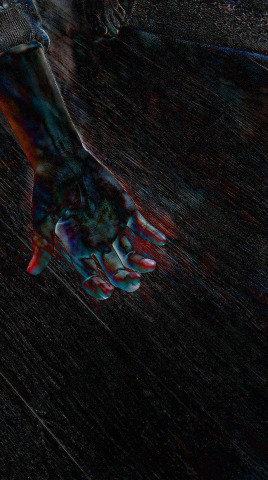} \\

        \multirow{2}{*}[0.35in]{\makebox[20pt]{\raisebox{25pt}{\rotatebox[origin=c]{90}{\textsc{Slice Banana}}}} \hspace{-7pt}} &
        \animage{35 150 25 100}{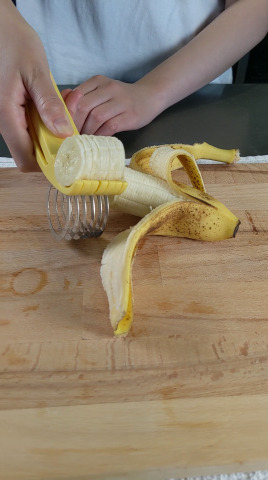} &
        \animage{35 150 25 100}{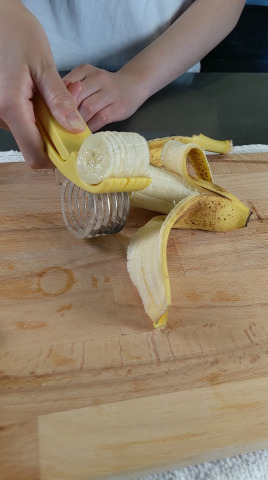} &
        \animage{35 150 25 100}{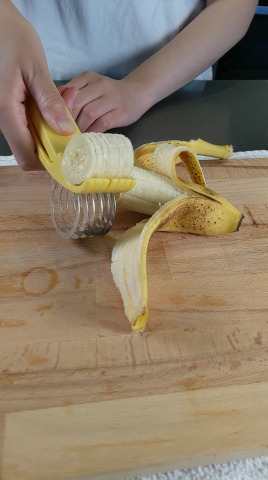} &
        \animage{35 150 25 100}{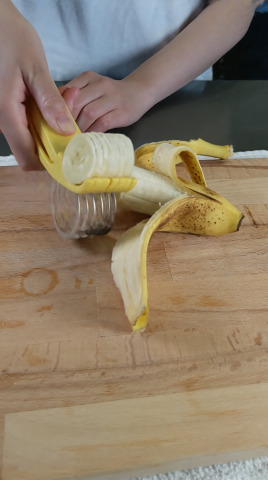} &
        \animage{35 150 25 100}{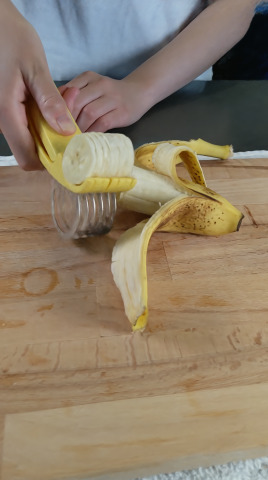} &
        \animage{35 150 25 100}{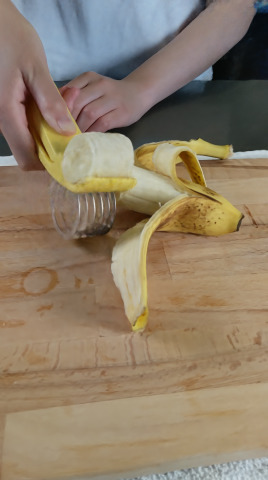} &
        \animage{35 150 25 100}{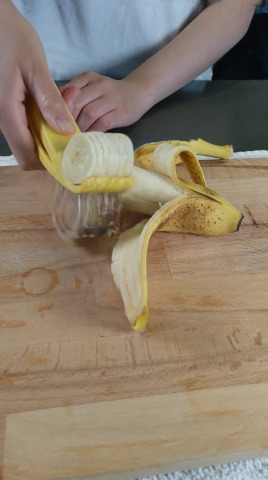} &
        \animage{35 150 25 100}{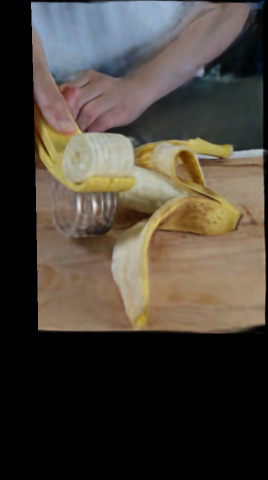} &
        \animage{35 150 25 100}{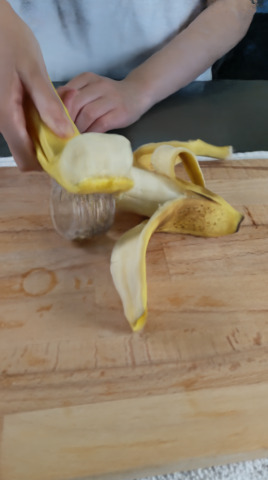} &
        \animage{35 150 25 100}{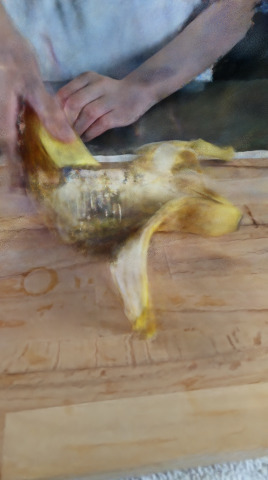} \\
        & & & &
        \animage{35 150 25 100}{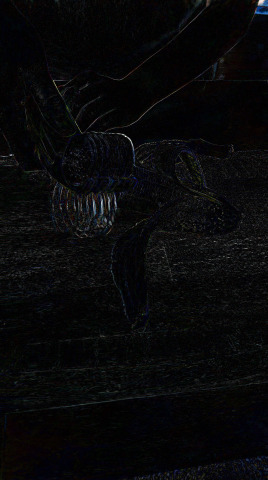} &
        \animage{35 150 25 100}{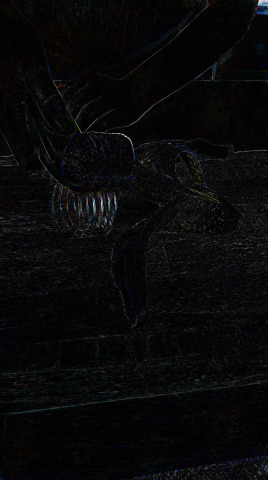} &
        \animage{35 150 25 100}{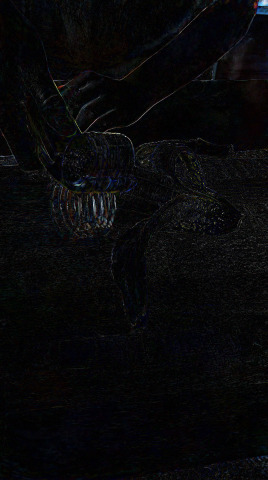} &
        \animage{35 150 25 100}{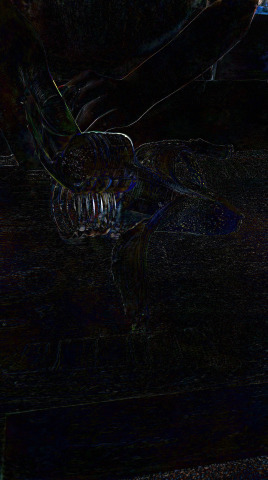} &
        \animage{35 150 25 100}{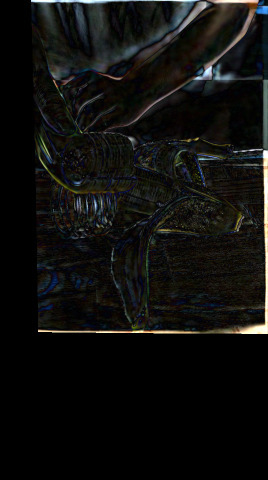} &
        \animage{35 150 25 100}{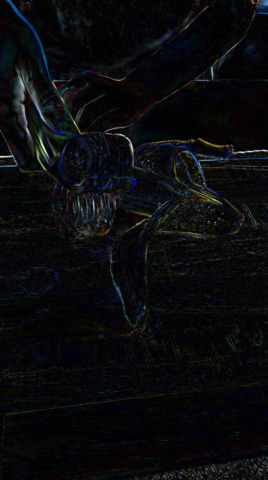} &
        \animage{35 150 25 100}{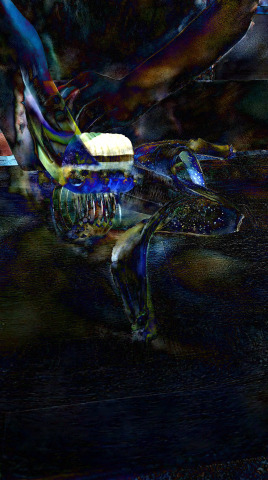} \\
        
        \multirow{2}{*}[0.13in]{\makebox[20pt]{\raisebox{25pt}{\rotatebox[origin=c]{90}{\textsc{Torch}}}} \hspace{-7pt}} &
        \animage{80 90 0 140}{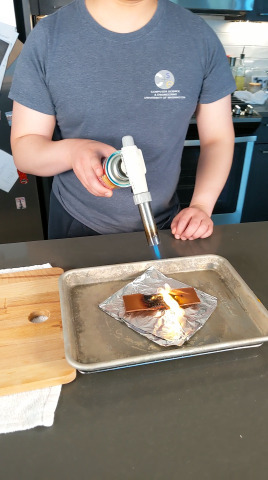} &
        \animage{80 90 0 140}{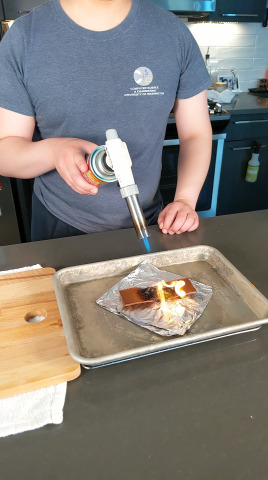} &
        \animage{80 90 0 140}{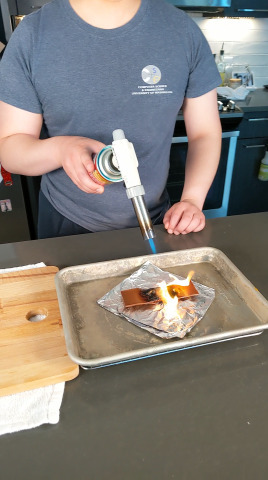} &
        \animage{80 90 0 140}{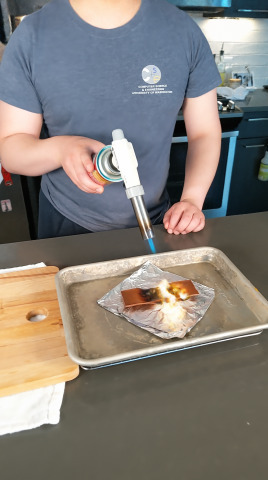} &
        \animage{80 90 0 140}{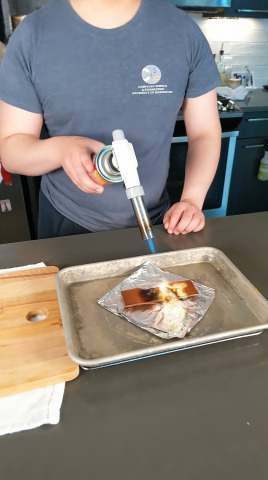} &
        \animage{80 90 0 140}{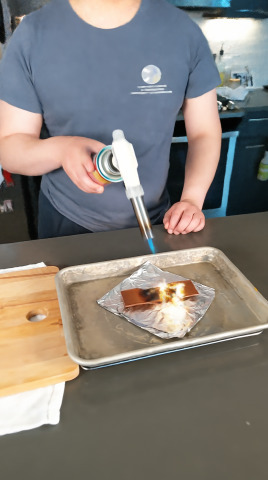} &
        \animage{80 90 0 140}{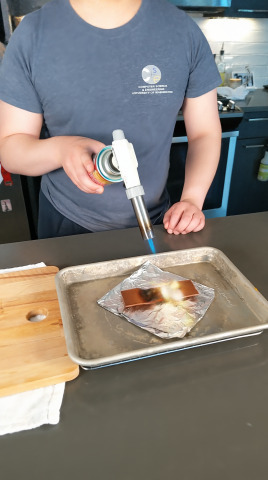} &
        \animage{80 90 0 140}{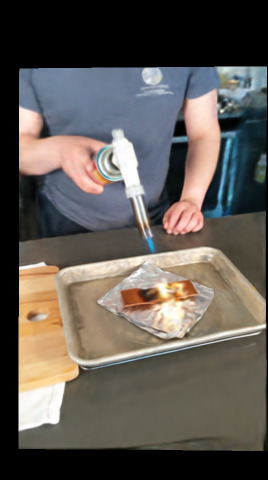} &
        \animage{80 90 0 140}{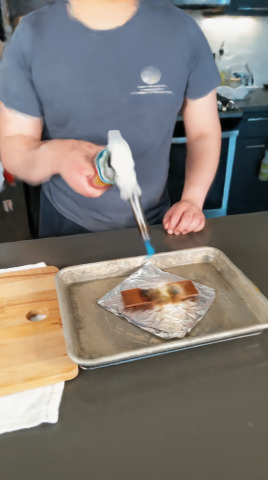} &
        \animage{80 90 0 140}{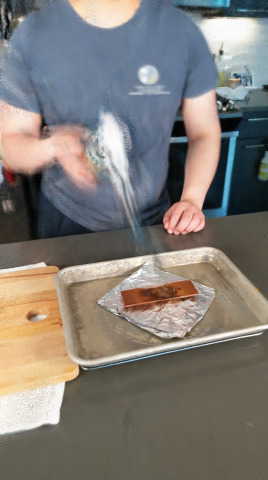} \\
        & & & &
        \animage{80 90 0 140}{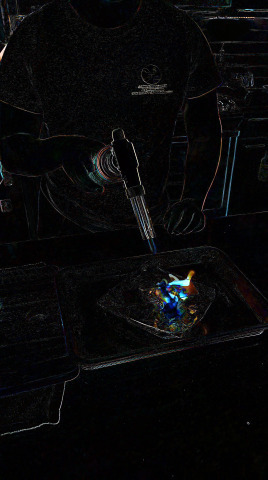} &
        \animage{80 90 0 140}{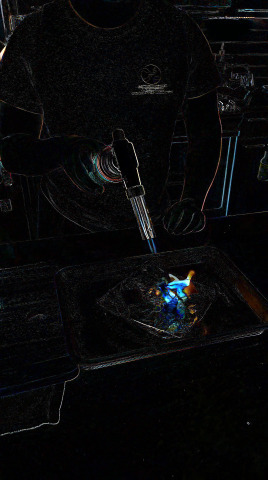} &
        \animage{80 90 0 140}{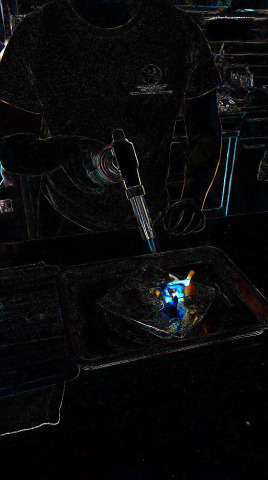} &
        \animage{80 90 0 140}{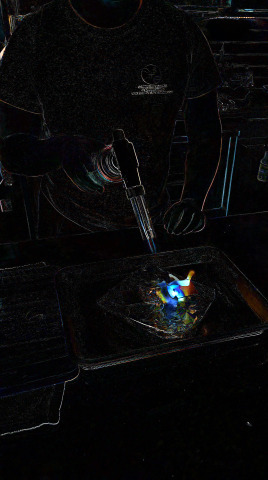} &
        \animage{80 90 0 140}{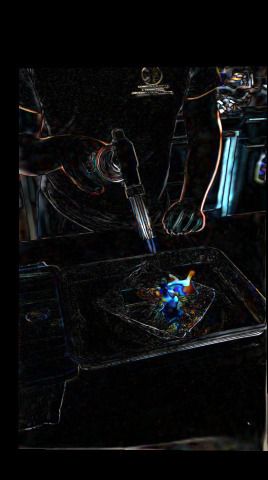} &
        \animage{80 90 0 140}{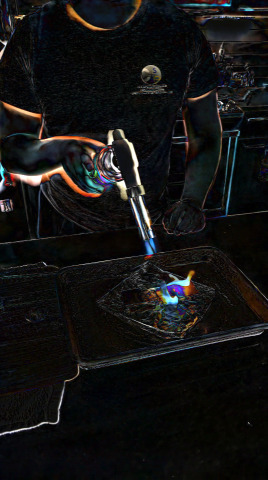} &
        \animage{80 90 0 140}{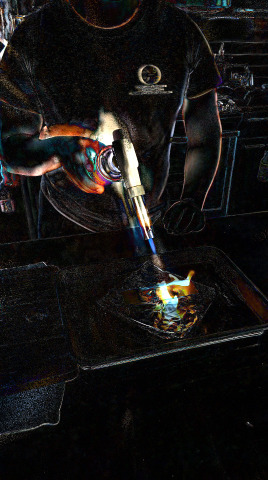} \\
        
         \multirow{2}{*}[0.13in]{\makebox[20pt]{\raisebox{25pt}{\rotatebox[origin=c]{90}{\textsc{Lemon}}}} \hspace{-7pt}} &
        \animage{120 40 90 20}{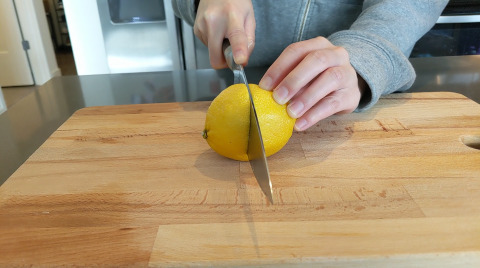} &
        \animage{80 40 130 20}{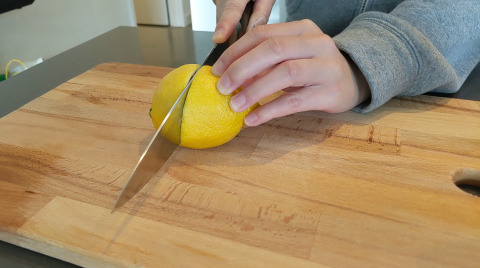} &
        \animage{80 40 130 20}{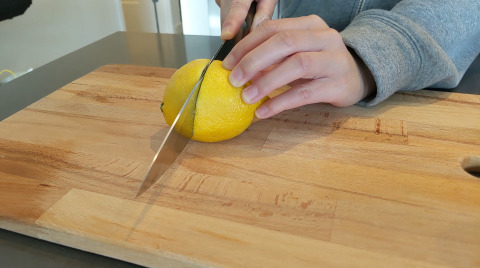} &
        \animage{80 40 130 20}{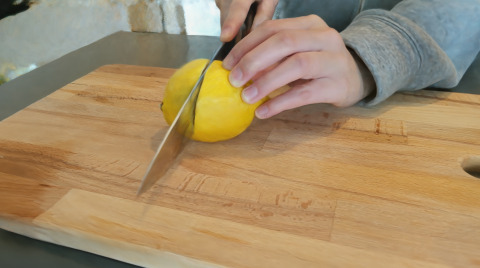} &
        \animage{80 40 130 20}{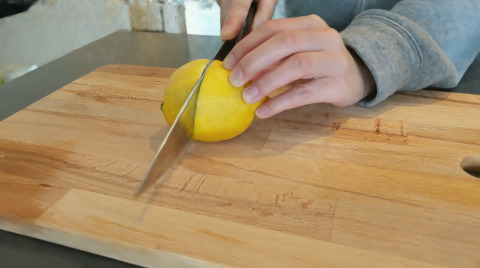} &
        \animage{80 40 130 20}{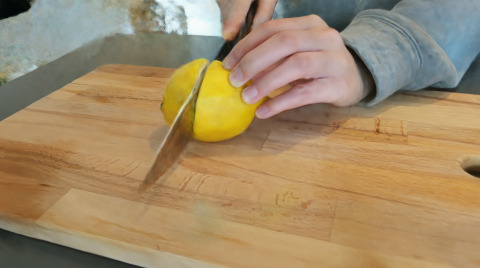} &
        \animage{80 40 130 20}{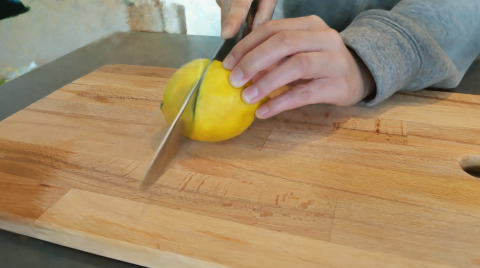} &
        \animage{80 40 130 20}{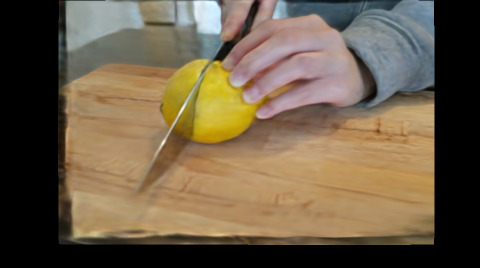} &
        \animage{80 36 130 20}{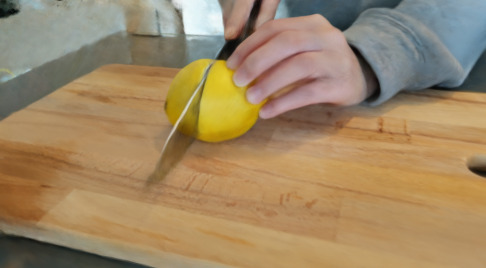} &
        \animage{80 40 130 20}{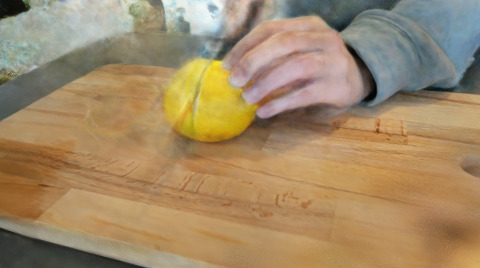} \\
        & & & &
        \animage{80 40 130 20}{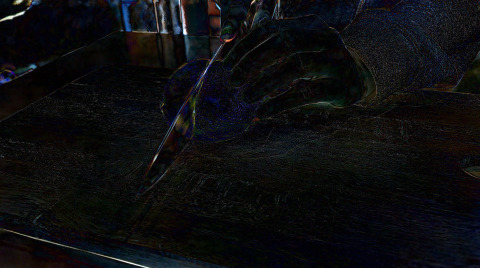} &
        \animage{80 40 130 20}{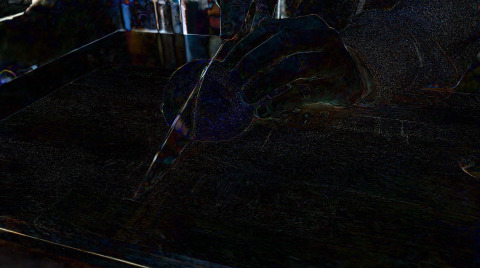} &
        \animage{80 40 130 20}{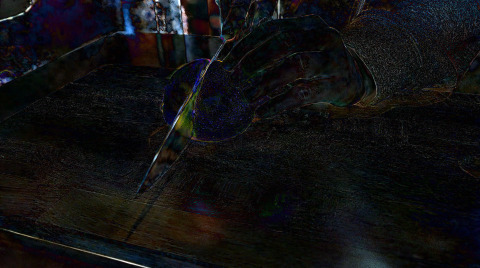} &
        \animage{80 40 130 20}{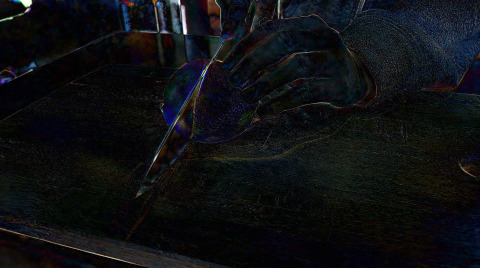} &
        \animage{80 40 130 20}{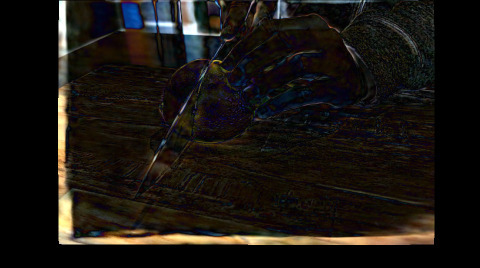} &
        \animage{80 36 130 20}{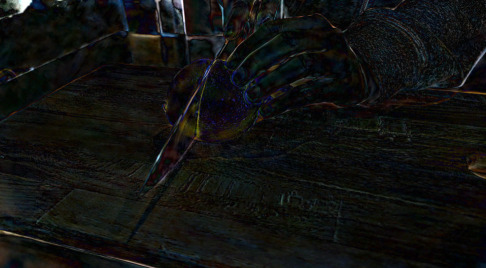} &
        \animage{80 40 130 20}{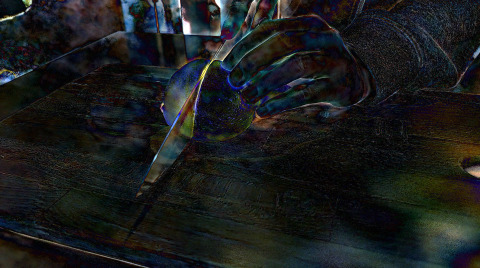} \\
        
        \\[2pt]
        
        & 
        {\smaller Reference 1} & 
        {\smaller Reference 2} & 
        {\smaller \makecell{Interpolated\\(GT)}} & 
        {\smaller \makecell{HyperNeRF\\(DS)}} &
        {\smaller \makecell{HyperNeRF\\(AP)}} &
        {\smaller No Deform} &
        {\smaller Nerfies} &
        {\smaller \makecell{Neural\\Volumes}} &
        {\smaller NSFF} &
        {\smaller NeRF}
        \\
    \end{tabular}
    \vspace{-6pt}
    \caption{(continued) Qualitative comparisons of our method, ablations, and baselines on the interpolation dataset (See \secref{sec:evaluation_quantitative}). The interpolated frame is rendered by linearly interpolating between the latent codes of the two reference frames. The bottom row of each sequence shows the absolute error. Baselines shown are: Nerfies~\cite{park2020nerfies}, Neural Volumes~\cite{lombardi2019neural}, NSFF~\cite{li2020nsff}, and NeRF~\cite{mildenhall2020nerf}.}
    \label{fig:interp_results2}
    \vspace{-20pt}
\end{figure*}

\section{Limitations and Conclusion}
As with all NeRF-like methods, camera registration affects the quality of the reconstruction. And, since our model uses only color images as input, imposing no domain-specific priors, we can only reconstruct what is observed. So, moments which are not captured well in the training data --- such as when there is rapid motion --- cannot be reconstructed by our method. We consider these limitations as fruitful avenues for future work.

We have presented HyperNeRF, an extension to NeRF that reconstructs topologically-varying scenes with discontinuous deformations.
Deformation-based dynamic NeRF models cannot model such topological variations due to their use of continuous deformation fields, encoded within the weights of an MLP. 
HyperNeRF models these variations as slices through a higher-dimensional space. To keep the space compact and avoid overfitting, we delay the use of these ambient dimensions, and then use deformable hyperplanes to extract these slices. Thus, we have combined insights from level-set methods and deformation-based dynamic NeRF models to reconstruct both large motions and topological variations.

\begin{anonsuppress}
\section*{Acknowledgments}
We thank Xuan Luo, Aleksander Holynski, and Hyunjeong Cho for their help with collecting data. We thank Pratul Srinivasan and John Flynn for their insightful discussions. This work was supported in part by the UW Reality Lab, Amazon, Facebook, Futurewei, and Google.
\end{anonsuppress}

\appendix

\section{Network Architecture}
\begin{figure}[t]
\centering
\includegraphics*[width=0.99\linewidth,clip,trim=0 5 0 3]{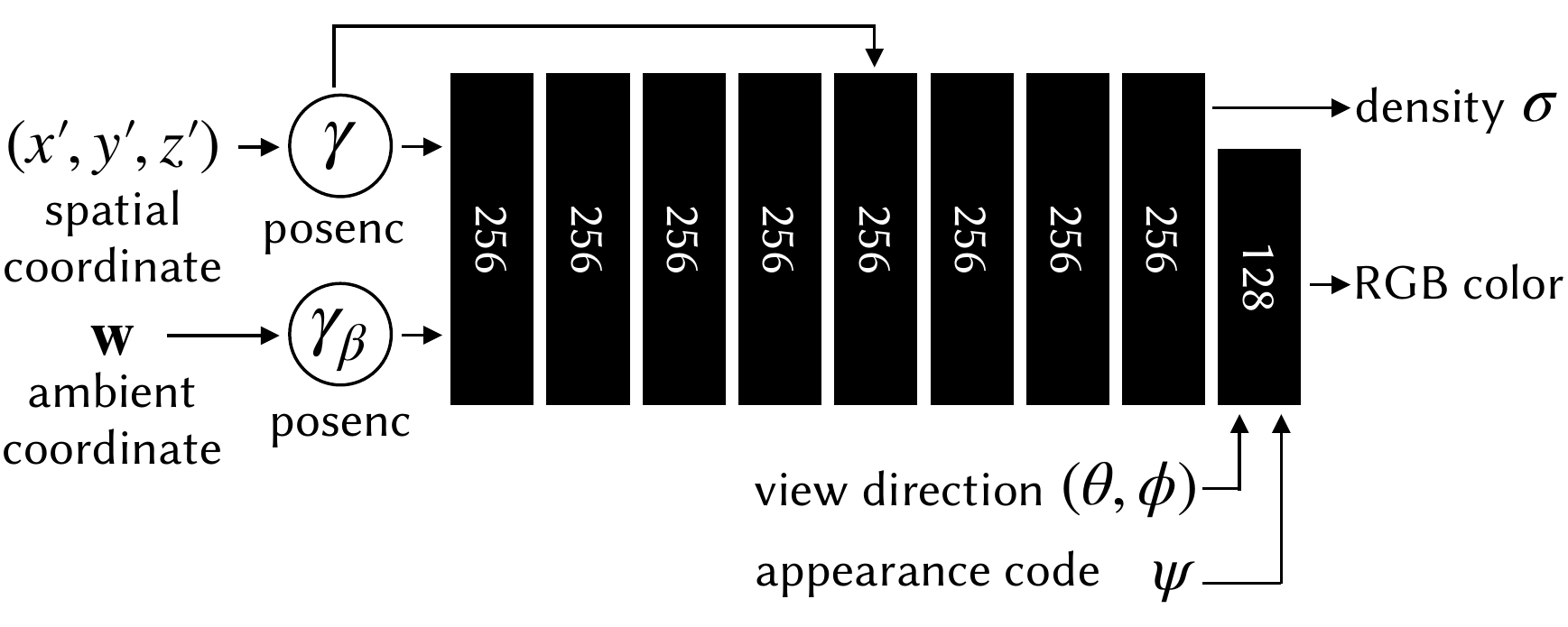}
\caption{A diagram of our hyper-space template network, which is identical to the original NeRF MLP, except it takes an additional ambient coordinate and an appearance latent code $\mat{\psi}$ as in \citet{martinbrualla2020nerfw}}
\label{fig:detailed_architecture_template}
\end{figure}

\begin{figure}[t]
\centering
\includegraphics*[width=0.99\linewidth,clip,trim=0 65 0 3]{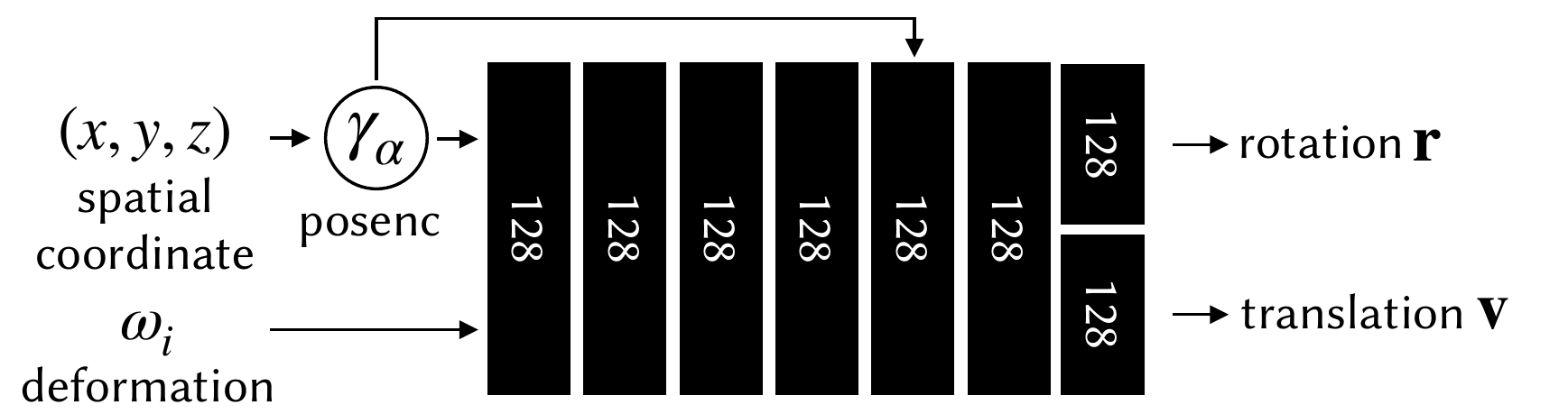}
\caption{A diagram of our deformation network. It is identical to the deformation MLP of Nerfies~\cite{park2020nerfies}.}
\label{fig:detailed_architecture_deformation}
\end{figure}

\begin{figure}[t]
\centering
\includegraphics*[width=0.99\linewidth,clip,trim=0 65 0 3]{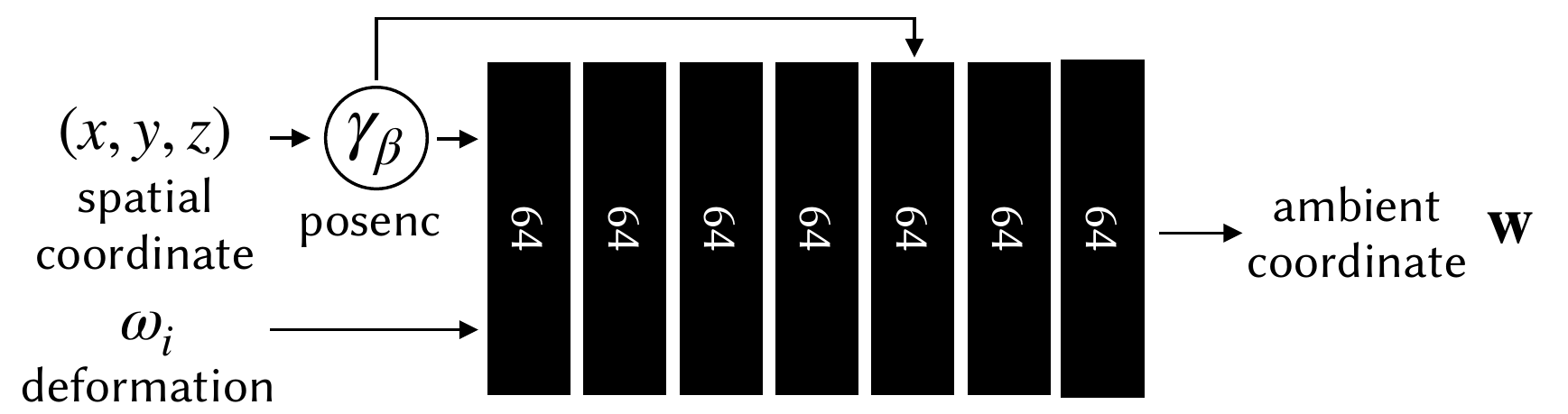}
\caption{A diagram of our ambient slicing surface network. It shares the input deformation code $\mat{\omega}_i$ with the deformation network. The ambient slicing surface network uses a different windowed positional encoding parameter $\beta$, and directly outputs an ambient coordinate $\mat{w}$.}
\label{fig:detailed_architecture_ambient}
\end{figure}

We provide detailed architecture diagrams for the template MLP in \figref{fig:detailed_architecture_template}, deformation MLP in \figref{fig:detailed_architecture_deformation}, and the ambient slicing surface MLP in \figref{fig:detailed_architecture_ambient}.

\section{Evaluation}

\subsection{Additional Qualitative Results}
We show qualitative results from the interpolation experiments in \figref{fig:interp_results1} and \figref{fig:interp_results2}.

\subsection{Additional Quantitative Metrics}
We provide per-sequence evaluations metrics for the interpolation task: \tabref{tab:evaluation_interpolation_lpips} reports LPIPS~\cite{zhang2018unreasonable}, \tabref{tab:evaluation_interpolation_ssim} report MS-SSIM~\cite{wang2003multiscale}, and \tabref{tab:evaluation_interpolation_psnr} report PSNR scores for each sequence and method.

\subsection{Details of the NSFF Experiments}
We use an updated version of the NSFF~\cite{li2020nsff} code, provided by the authors, which includes numerous bug fixes and improvements. The default hyper-parameters provided did not perform well on our sequences and we therefore tune them:
\begin{center}
    \resizebox{0.7\linewidth}{!}{
    \begin{tabular}{c|c|c|c|c}
    \toprule
    $\beta_z$  & $\beta_\text{optical\,flow}$ & $\beta_\text{flow\,smoothness}$ & $\beta_\text{cyc}$ & $\beta_w$ \\ \hline
    0.04 & 0.02 & 0.1 & 1.0 & 0.1 \\
    \bottomrule
    \end{tabular}
    }
\end{center}
In addition, the original code decays the supervised losses (depth and flow) over 25,000 iterations regardless of the length of the dataset resulting in poor performance for longer sequences. Therefore we instead decay the losses over $1000N$ iterations, where $N$ is the number of frames in the dataset.

\subsection{Details of the Neural Volumes Experiments}
We closely follow the experiment procedure described in Sec. E.2. of Nerfies~\cite{park2020nerfies}.

\section{2D Level Set Experiments}
Here we describe our method for learning a family of topologically varying 2D signed distance functions (SDF). This method was used to generate Fig. 3.

\begin{figure}[t]
\centering
\includegraphics*[width=0.99\linewidth,clip,trim=0 45 0 3]{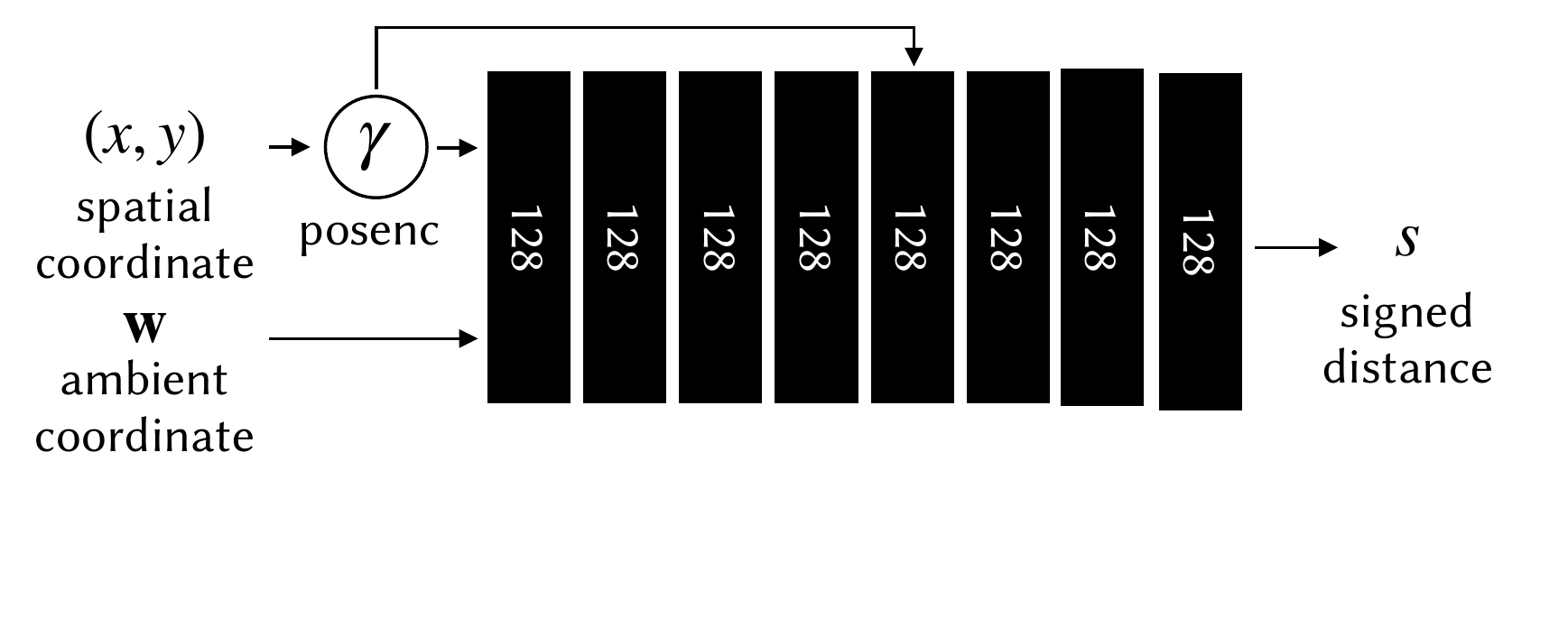}
\caption{A diagram of architecture for the 2D template MLP used in the 2D level set experiments.}
\label{fig:detailed_architecture_2d_template}
\end{figure}

\begin{figure}[t]
\centering
\includegraphics*[width=0.99\linewidth,clip,trim=0 45 0 3]{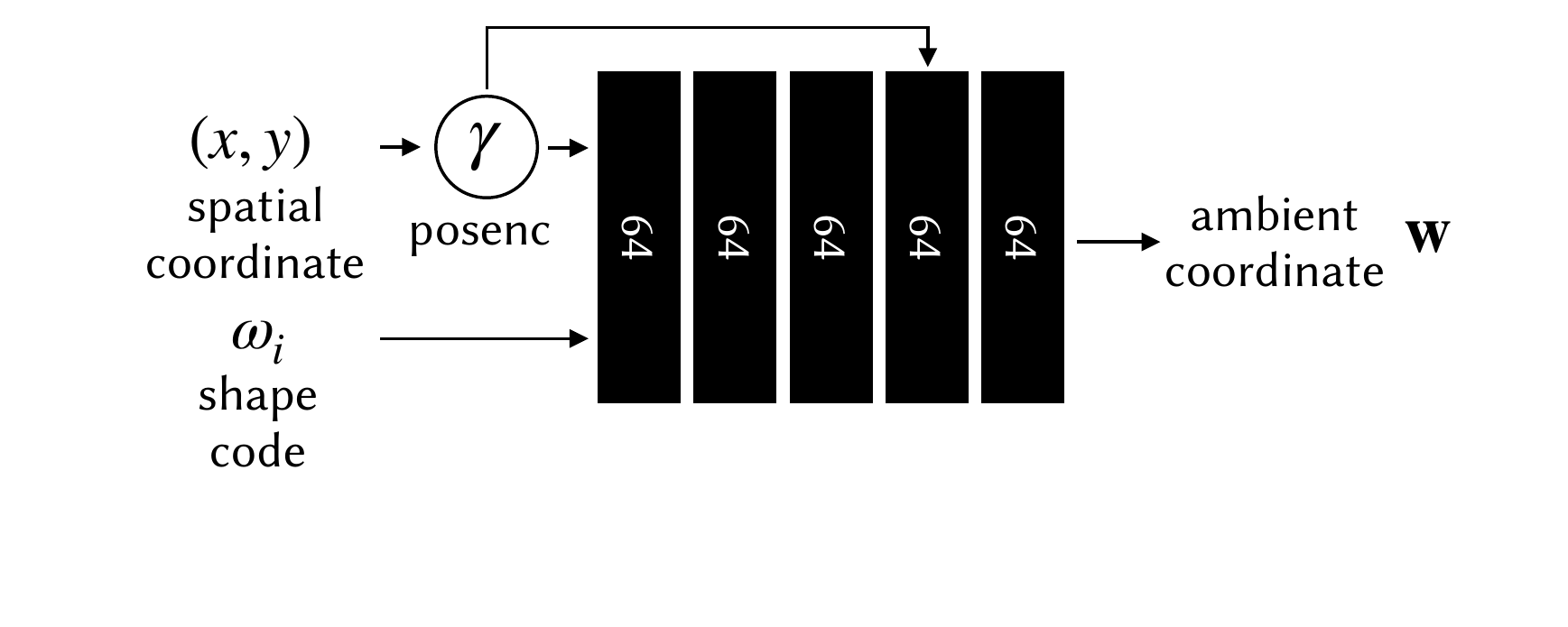}
\caption{A diagram of architecture for the ambient slicing surface MLP used in the 2D level set experiments.}
\label{fig:detailed_architecture_2d_ambient}
\end{figure}

We used truncated signed distance functions, truncated to range between -0.05 and 0.05. We learn a template MLP
\begin{equation}
    F : (x,y,w) \rightarrow s\,,
\end{equation}
which takes as input a normalized 2D coordinate $(x,y)\in[-1,1]^2$ an ambient coordinate $w\in\mathbb{R}$, and outputs a signed distance $s$. This function defines a 3D surface which is sliced by a slicing surface. This be an axis-aligned plane, defined by a value of $w$ defining a plane that spans the $x$ and $y$ axes, or a deformable slicing surface defined by an ambient slicing surface MLP
\begin{equation}
    H : (x,y,\omega_i) \rightarrow w\,,
\end{equation}
where $(x,y)$ are again the spatial coordinates, $\mat{\omega}_i$ is a per-shape latent code, and $w$ is the output ambient coordinate. We show a detailed diagram of the 2D template in \figref{fig:detailed_architecture_2d_template} and deformable slicing surface in \figref{fig:detailed_architecture_2d_ambient}.

\paragraph{Training} Training batches are generated by randomly sampling points from the continuous, truncated SDFs of each shape $i$, resulting in data of the form $\left\{((x,y,\mat{\omega}_i),s)\right\}$. We use a Pseudo-Huber loss~\cite{charbonnier1997deterministic} function:
\begin{equation}
    L_\delta(s-s^*) = \delta^2 \left(\sqrt{1+\left(\frac{s-s^*}{\delta}\right)^2} - 1\right)\,,
\end{equation}
where s is the predicted SDF value, $s^*$ is the ground truth SDF value, and $\delta$ is a hyper-parameter that controls the steepness which we set to $\delta=0.005$.

\paragraph{Implementation Details} We use a positional encoding with a minimum degree as well as a maximum degree:
\begin{equation}
\resizebox{0.92\linewidth}{!}{%
$
\gamma(\spatial) = \left[\sin(2^{m_-}\spatial),\cos(2^{m_-}\spatial),\ldots,\sin(2^{m_+-1}\spatial),\cos(2^{m_+-1}\spatial)\right]^\mathrm{T}\,,
$}
\end{equation}
where $m_-$ is the minimum degree and $m_+$ is the maximum degree of the positional encoding. For the template MLP we $m_-=-2$ and $m_+=3$, for the ambient slicing surface MLP we use $m_-=-2$ and $m_+=2$. We train using the Adam optimizer with a fixed learning rate of $10^{-3}$ for 2000 iterations with a batch size of 512 which takes around a minute on a Google Colab TPU.

\paragraph{Interpolating Shapes} We generate interpolated shaped by linearly interpolating the shape code $\omega_i$ between instances.

\renewcommand{\tablefirst}[0]{\cellcolor{myred}}
\renewcommand{\tablesecond}[0]{\cellcolor{myorange}}
\renewcommand{\tablethird}[0]{\cellcolor{myyellow}}
\begin{table*}[t]
\caption{
Per-sequence PSNR metrics on the interpolation task. We color code each row as \colorbox{myred}{\textbf{best}}, \colorbox{myorange}{\textbf{second best}}, and \colorbox{myyellow}{\textbf{third best}}. 
\label{tab:evaluation_interpolation_psnr}
}
\resizebox{\linewidth}{!}{
\centering
\setlength{\tabcolsep}{2pt}


\begin{tabular}{l||c|c|c|c|c|c|c|c|c|c}

\toprule
& \multicolumn{ 1 }{c}{
  \makecell{
  \textsc{\small Expressions 1 }
\\(391 images)
  }
}
& \multicolumn{ 1 }{c}{
  \makecell{
  \textsc{\small Expressions 2 }
\\(126 images)
  }
}
& \multicolumn{ 1 }{c}{
  \makecell{
  \textsc{\small Expressions 3 }
\\(110 images)
  }
}
& \multicolumn{ 1 }{c}{
  \makecell{
  \textsc{\small Teapots }
\\(243 images)
  }
}
& \multicolumn{ 1 }{c}{
  \makecell{
  \textsc{\small Chicken }
\\(113 images)
  }
}
& \multicolumn{ 1 }{c}{
  \makecell{
  \textsc{\small Fist }
\\(433 images)
  }
}
& \multicolumn{ 1 }{c}{
  \makecell{
  \textsc{\small Slice Banana }
\\(82 images)
  }
}
& \multicolumn{ 1 }{c}{
  \makecell{
  \textsc{\small Torch }
\\(173 images)
  }
}
& \multicolumn{ 1 }{c}{
  \makecell{
  \textsc{\small Lemon }
\\(415 images)
  }
}
& \multicolumn{ 1 }{c}{
  \makecell{
  \textsc{\small Mean }
  }
}
\\

& \multicolumn{1}{c}{ \footnotesize PSNR$\uparrow$ }
& \multicolumn{1}{c}{ \footnotesize PSNR$\uparrow$ }
& \multicolumn{1}{c}{ \footnotesize PSNR$\uparrow$ }
& \multicolumn{1}{c}{ \footnotesize PSNR$\uparrow$ }
& \multicolumn{1}{c}{ \footnotesize PSNR$\uparrow$ }
& \multicolumn{1}{c}{ \footnotesize PSNR$\uparrow$ }
& \multicolumn{1}{c}{ \footnotesize PSNR$\uparrow$ }
& \multicolumn{1}{c}{ \footnotesize PSNR$\uparrow$ }
& \multicolumn{1}{c}{ \footnotesize PSNR$\uparrow$ }
& \multicolumn{1}{c}{ \footnotesize PSNR$\uparrow$ }
\\
\hline

  NeRF~\cite{mildenhall2020nerf}
  &$21.6$
  
  &$21.1$
  
  &$15.6$
  
  &$23.6$
  
  &$18.8$
  
  &$23.8$
  
  &$20.8$
  
  &$22.5$
  
  &$24.1$
  
  &$21.3$
  
  \\   NV~\cite{lombardi2019neural}
  &$26.7$
  
  &$25.6$
  
  &$18.6$
  
  &$26.2$
  
  &$22.6$
  
  &$29.3$
  
  &$24.8$
  
  &$24.6$
  
  &$28.8$
  
  &$25.3$
  
  \\   NSFF~\cite{li2020nsff}
  &$26.6$
  
  &$29.8$
  
  &$18.3$
  
  &$25.8$
  
  &$27.7$
  
  &$24.9$
  
  &$26.1$
  
  &$22.3$
  
  &$28.0$
  
  &$25.5$
  
  \\   Nerfies~\cite{park2020nerfies}
  &$27.5$
  
  &$31.6$
  
  &$19.7$
  
  &$25.7$
  
  &$28.7$
  
  &$29.9$
  
  &$27.9$
  
  &$27.8$
  
  &$30.8$
  
  &$27.7$
  
  \\   Nerfies (w/o elastic)
  &$27.7$
  
  &\tablethird$32.5$
  
  &\tablethird$20.0$
  
  &$25.8$
  
  &\tablethird$28.8$
  
  &\tablethird$30.5$
  
  &$28.1$
  
  &\tablethird$27.9$
  
  &$30.8$
  
  &$28.0$
  
  \\ \hline  Hyper-NeRF (DS)
  &\tablesecond$27.9$
  
  &\tablesecond$32.9$
  
  &\tablefirst$20.1$
  
  &\tablesecond$26.4$
  
  &$28.7$
  
  &\tablefirst$30.7$
  
  &\tablefirst$28.4$
  
  &\tablesecond$28.0$
  
  &\tablefirst$31.8$
  
  &\tablefirst$28.3$
  
  \\   Hyper-NeRF (DS, w/ elastic)
  &\tablethird$27.8$
  
  &$32.4$
  
  &$19.9$
  
  &\tablefirst$26.4$
  
  &\tablesecond$28.8$
  
  &$30.2$
  
  &\tablethird$28.3$
  
  &$27.9$
  
  &\tablesecond$31.4$
  
  &\tablethird$28.1$
  
  \\   Hyper-NeRF (AP)
  &\tablefirst$27.9$
  
  &\tablefirst$33.1$
  
  &\tablesecond$20.0$
  
  &\tablethird$26.3$
  
  &\tablefirst$28.9$
  
  &\tablesecond$30.5$
  
  &\tablesecond$28.4$
  
  &\tablefirst$28.0$
  
  &\tablethird$31.3$
  
  &\tablesecond$28.3$
  
  \\   Hyper-NeRF (w/o deform)
  &$26.9$
  
  &$31.1$
  
  &$19.4$
  
  &$26.0$
  
  &$27.7$
  
  &$29.3$
  
  &$28.1$
  
  &$27.3$
  
  &$30.5$
  
  &$27.4$
  
  \\ \bottomrule

\end{tabular}

}
\end{table*}
\begin{table*}[t]
\caption{
Per-sequence MS-SSIM metrics on the interpolation task. We color code each row as \colorbox{myred}{\textbf{best}}, \colorbox{myorange}{\textbf{second best}}, and \colorbox{myyellow}{\textbf{third best}}. 
\label{tab:evaluation_interpolation_ssim}
}
\resizebox{\linewidth}{!}{
\centering
\setlength{\tabcolsep}{2pt}


\begin{tabular}{l||c|c|c|c|c|c|c|c|c|c}

\toprule
& \multicolumn{ 1 }{c}{
  \makecell{
  \textsc{\small Expressions 1 }
\\(391 images)
  }
}
& \multicolumn{ 1 }{c}{
  \makecell{
  \textsc{\small Expressions 2 }
\\(126 images)
  }
}
& \multicolumn{ 1 }{c}{
  \makecell{
  \textsc{\small Expressions 3 }
\\(110 images)
  }
}
& \multicolumn{ 1 }{c}{
  \makecell{
  \textsc{\small Teapots }
\\(243 images)
  }
}
& \multicolumn{ 1 }{c}{
  \makecell{
  \textsc{\small Chicken }
\\(113 images)
  }
}
& \multicolumn{ 1 }{c}{
  \makecell{
  \textsc{\small Fist }
\\(433 images)
  }
}
& \multicolumn{ 1 }{c}{
  \makecell{
  \textsc{\small Slice Banana }
\\(82 images)
  }
}
& \multicolumn{ 1 }{c}{
  \makecell{
  \textsc{\small Torch }
\\(173 images)
  }
}
& \multicolumn{ 1 }{c}{
  \makecell{
  \textsc{\small Lemon }
\\(415 images)
  }
}
& \multicolumn{ 1 }{c}{
  \makecell{
  \textsc{\small Mean }
  }
}
\\

& \multicolumn{1}{c}{ \footnotesize MS-SSIM$\uparrow$ }
& \multicolumn{1}{c}{ \footnotesize MS-SSIM$\uparrow$ }
& \multicolumn{1}{c}{ \footnotesize MS-SSIM$\uparrow$ }
& \multicolumn{1}{c}{ \footnotesize MS-SSIM$\uparrow$ }
& \multicolumn{1}{c}{ \footnotesize MS-SSIM$\uparrow$ }
& \multicolumn{1}{c}{ \footnotesize MS-SSIM$\uparrow$ }
& \multicolumn{1}{c}{ \footnotesize MS-SSIM$\uparrow$ }
& \multicolumn{1}{c}{ \footnotesize MS-SSIM$\uparrow$ }
& \multicolumn{1}{c}{ \footnotesize MS-SSIM$\uparrow$ }
& \multicolumn{1}{c}{ \footnotesize MS-SSIM$\uparrow$ }
\\
\hline

  NeRF~\cite{mildenhall2020nerf}
  &$.708$
  
  &$.722$
  
  &$.449$
  
  &$.875$
  
  &$.761$
  
  &$.773$
  
  &$.721$
  
  &$.866$
  
  &$.826$
  
  &$.745$
  
  \\   NV~\cite{lombardi2019neural}
  &\tablefirst$.900$
  
  &$.910$
  
  &$.647$
  
  &$.917$
  
  &$.861$
  
  &$.912$
  
  &$.907$
  
  &$.917$
  
  &\tablethird$.951$
  
  &$.880$
  
  \\   NSFF~\cite{li2020nsff}
  &$.885$
  
  &$.933$
  
  &$.654$
  
  &$.915$
  
  &$.939$
  
  &$.797$
  
  &$.858$
  
  &$.883$
  
  &$.904$
  
  &$.863$
  
  \\   Nerfies~\cite{park2020nerfies}
  &$.877$
  
  &$.963$
  
  &$.698$
  
  &$.922$
  
  &\tablethird$.948$
  
  &$.940$
  
  &$.916$
  
  &$.959$
  
  &$.946$
  
  &$.908$
  
  \\   Nerfies (w/o elastic)
  &$.883$
  
  &\tablethird$.967$
  
  &$.698$
  
  &$.920$
  
  &$.947$
  
  &\tablethird$.946$
  
  &$.915$
  
  &\tablesecond$.961$
  
  &$.946$
  
  &$.909$
  
  \\ \hline  Hyper-NeRF (DS)
  &\tablesecond$.887$
  
  &\tablefirst$.970$
  
  &\tablesecond$.703$
  
  &\tablesecond$.931$
  
  &$.948$
  
  &\tablefirst$.950$
  
  &\tablefirst$.920$
  
  &\tablefirst$.962$
  
  &\tablefirst$.956$
  
  &\tablefirst$.914$
  
  \\   Hyper-NeRF (DS, w/ elastic)
  &$.885$
  
  &$.966$
  
  &\tablethird$.700$
  
  &\tablefirst$.932$
  
  &\tablefirst$.949$
  
  &$.943$
  
  &\tablesecond$.918$
  
  &$.960$
  
  &\tablesecond$.952$
  
  &\tablethird$.912$
  
  \\   Hyper-NeRF (AP)
  &\tablethird$.885$
  
  &\tablesecond$.970$
  
  &\tablefirst$.703$
  
  &\tablethird$.927$
  
  &\tablesecond$.949$
  
  &\tablesecond$.947$
  
  &\tablethird$.917$
  
  &\tablethird$.960$
  
  &$.947$
  
  &\tablesecond$.912$
  
  \\   Hyper-NeRF (w/o deform)
  &$.857$
  
  &$.947$
  
  &$.669$
  
  &$.922$
  
  &$.936$
  
  &$.924$
  
  &$.915$
  
  &$.951$
  
  &$.937$
  
  &$.895$
  
  \\ \bottomrule

\end{tabular}

}
\end{table*}
\begin{table*}[t]
\caption{
Per-sequence LPIPS metrics on the interpolation task. We color code each row as \colorbox{myred}{\textbf{best}}, \colorbox{myorange}{\textbf{second best}}, and \colorbox{myyellow}{\textbf{third best}}. 
\label{tab:evaluation_interpolation_lpips}
}
\resizebox{\linewidth}{!}{
\centering
\setlength{\tabcolsep}{2pt}


\begin{tabular}{l||c|c|c|c|c|c|c|c|c|c}

\toprule
& \multicolumn{ 1 }{c}{
  \makecell{
  \textsc{\small Expressions 1 }
\\(391 images)
  }
}
& \multicolumn{ 1 }{c}{
  \makecell{
  \textsc{\small Expressions 2 }
\\(126 images)
  }
}
& \multicolumn{ 1 }{c}{
  \makecell{
  \textsc{\small Expressions 3 }
\\(110 images)
  }
}
& \multicolumn{ 1 }{c}{
  \makecell{
  \textsc{\small Teapots }
\\(243 images)
  }
}
& \multicolumn{ 1 }{c}{
  \makecell{
  \textsc{\small Chicken }
\\(113 images)
  }
}
& \multicolumn{ 1 }{c}{
  \makecell{
  \textsc{\small Fist }
\\(433 images)
  }
}
& \multicolumn{ 1 }{c}{
  \makecell{
  \textsc{\small Slice Banana }
\\(82 images)
  }
}
& \multicolumn{ 1 }{c}{
  \makecell{
  \textsc{\small Torch }
\\(173 images)
  }
}
& \multicolumn{ 1 }{c}{
  \makecell{
  \textsc{\small Lemon }
\\(415 images)
  }
}
& \multicolumn{ 1 }{c}{
  \makecell{
  \textsc{\small Mean }
  }
}
\\

& \multicolumn{1}{c}{ \footnotesize LPIPS$\downarrow$ }
& \multicolumn{1}{c}{ \footnotesize LPIPS$\downarrow$ }
& \multicolumn{1}{c}{ \footnotesize LPIPS$\downarrow$ }
& \multicolumn{1}{c}{ \footnotesize LPIPS$\downarrow$ }
& \multicolumn{1}{c}{ \footnotesize LPIPS$\downarrow$ }
& \multicolumn{1}{c}{ \footnotesize LPIPS$\downarrow$ }
& \multicolumn{1}{c}{ \footnotesize LPIPS$\downarrow$ }
& \multicolumn{1}{c}{ \footnotesize LPIPS$\downarrow$ }
& \multicolumn{1}{c}{ \footnotesize LPIPS$\downarrow$ }
& \multicolumn{1}{c}{ \footnotesize LPIPS$\downarrow$ }
\\
\hline

  NeRF~\cite{mildenhall2020nerf}
  &$.582$
  
  &$.492$
  
  &$.747$
  
  &$.339$
  
  &$.453$
  
  &$.469$
  
  &$.513$
  
  &$.373$
  
  &$.437$
  
  &$.490$
  
  \\   NV~\cite{lombardi2019neural}
  &\tablefirst$.215$
  
  &$.130$
  
  &$.318$
  
  &$.216$
  
  &$.243$
  
  &$.213$
  
  &\tablethird$.209$
  
  &$.189$
  
  &\tablefirst$.190$
  
  &$.214$
  
  \\   NSFF~\cite{li2020nsff}
  &$.283$
  
  &$.134$
  
  &$.317$
  
  &\tablefirst$.210$
  
  &$.173$
  
  &$.329$
  
  &$.243$
  
  &$.253$
  
  &$.238$
  
  &$.242$
  
  \\   Nerfies~\cite{park2020nerfies}
  &$.224$
  
  &$.0952$
  
  &\tablethird$.275$
  
  &$.225$
  
  &\tablefirst$.141$
  
  &$.171$
  
  &$.209$
  
  &\tablesecond$.169$
  
  &\tablethird$.223$
  
  &\tablesecond$.193$
  
  \\   Nerfies (w/o elastic)
  &\tablethird$.219$
  
  &\tablesecond$.0857$
  
  &$.279$
  
  &$.229$
  
  &$.160$
  
  &\tablethird$.158$
  
  &\tablesecond$.200$
  
  &\tablefirst$.168$
  
  &$.239$
  
  &\tablethird$.193$
  
  \\ \hline  Hyper-NeRF (DS)
  &\tablesecond$.218$
  
  &\tablefirst$.0825$
  
  &\tablesecond$.274$
  
  &\tablesecond$.212$
  
  &\tablethird$.156$
  
  &\tablefirst$.150$
  
  &\tablefirst$.191$
  
  &\tablethird$.172$
  
  &\tablesecond$.210$
  
  &\tablefirst$.185$
  
  \\   Hyper-NeRF (DS, w/ elastic)
  &$.220$
  
  &$.0956$
  
  &\tablefirst$.269$
  
  &\tablethird$.212$
  
  &\tablesecond$.145$
  
  &$.162$
  
  &$.249$
  
  &$.173$
  
  &$.230$
  
  &$.195$
  
  \\   Hyper-NeRF (AP)
  &$.240$
  
  &\tablethird$.0894$
  
  &$.290$
  
  &$.234$
  
  &$.162$
  
  &\tablesecond$.158$
  
  &$.241$
  
  &$.183$
  
  &$.276$
  
  &$.208$
  
  \\   Hyper-NeRF (w/o deform)
  &$.312$
  
  &$.131$
  
  &$.352$
  
  &$.238$
  
  &$.203$
  
  &$.212$
  
  &$.299$
  
  &$.224$
  
  &$.303$
  
  &$.253$
  
  \\ \bottomrule

\end{tabular}

}
\end{table*}
\begin{table*}[t]
\caption{
Per-sequence ``average'' metrics~\cite{barron2021mipnerf} on the interpolation task. We color each row as \colorbox{myred}{\textbf{best}}, \colorbox{myorange}{\textbf{second best}}, and \colorbox{myyellow}{\textbf{third best}}. 
\label{tab:evaluation_interpolation_average}
}
\resizebox{\linewidth}{!}{
\centering
\setlength{\tabcolsep}{2pt}


\begin{tabular}{l||c|c|c|c|c|c|c|c|c|c}

\toprule
& \multicolumn{ 1 }{c}{
  \makecell{
  \textsc{\small Expressions 1 }
\\(391 images)
  }
}
& \multicolumn{ 1 }{c}{
  \makecell{
  \textsc{\small Expressions 2 }
\\(126 images)
  }
}
& \multicolumn{ 1 }{c}{
  \makecell{
  \textsc{\small Expressions 3 }
\\(110 images)
  }
}
& \multicolumn{ 1 }{c}{
  \makecell{
  \textsc{\small Teapots }
\\(243 images)
  }
}
& \multicolumn{ 1 }{c}{
  \makecell{
  \textsc{\small Chicken }
\\(113 images)
  }
}
& \multicolumn{ 1 }{c}{
  \makecell{
  \textsc{\small Fist }
\\(433 images)
  }
}
& \multicolumn{ 1 }{c}{
  \makecell{
  \textsc{\small Slice Banana }
\\(82 images)
  }
}
& \multicolumn{ 1 }{c}{
  \makecell{
  \textsc{\small Torch }
\\(173 images)
  }
}
& \multicolumn{ 1 }{c}{
  \makecell{
  \textsc{\small Lemon }
\\(415 images)
  }
}
& \multicolumn{ 1 }{c}{
  \makecell{
  \textsc{\small Mean }
  }
}
\\

& \multicolumn{1}{c}{ \footnotesize Avg$\downarrow$ }
& \multicolumn{1}{c}{ \footnotesize Avg$\downarrow$ }
& \multicolumn{1}{c}{ \footnotesize Avg$\downarrow$ }
& \multicolumn{1}{c}{ \footnotesize Avg$\downarrow$ }
& \multicolumn{1}{c}{ \footnotesize Avg$\downarrow$ }
& \multicolumn{1}{c}{ \footnotesize Avg$\downarrow$ }
& \multicolumn{1}{c}{ \footnotesize Avg$\downarrow$ }
& \multicolumn{1}{c}{ \footnotesize Avg$\downarrow$ }
& \multicolumn{1}{c}{ \footnotesize Avg$\downarrow$ }
& \multicolumn{1}{c}{ \footnotesize Avg$\downarrow$ }
\\
\hline

  NeRF~\cite{mildenhall2020nerf}
  &$.130$
  
  &$.127$
  
  &$.248$
  
  &$.0809$
  
  &$.143$
  
  &$.0973$
  
  &$.131$
  
  &$.0916$
  
  &$.0889$
  
  &$.126$
  
  \\   NV~\cite{lombardi2019neural}
  &$.0524$
  
  &$.0475$
  
  &$.138$
  
  &$.0532$
  
  &$.0791$
  
  &$.0419$
  
  &$.0596$
  
  &$.0574$
  
  &$.0381$
  
  &$.0630$
  
  \\   NSFF~\cite{li2020nsff}
  &$.0596$
  
  &$.0331$
  
  &$.140$
  
  &$.0546$
  
  &$.0416$
  
  &$.0780$
  
  &$.0608$
  
  &$.0798$
  
  &$.0490$
  
  &$.0663$
  
  \\   Nerfies~\cite{park2020nerfies}
  &$.0519$
  
  &$.0234$
  
  &$.118$
  
  &$.0552$
  
  &\tablefirst$.0351$
  
  &$.0350$
  
  &\tablethird$.0462$
  
  &$.0385$
  
  &\tablethird$.0351$
  
  &$.0487$
  
  \\   Nerfies (w/o elastic)
  &\tablethird$.0501$
  
  &\tablethird$.0206$
  
  &\tablethird$.115$
  
  &$.0554$
  
  &$.0366$
  
  &\tablethird$.0321$
  
  &\tablesecond$.0447$
  
  &\tablefirst$.0377$
  
  &$.0359$
  
  &\tablethird$.0476$
  
  \\ \hline  Hyper-NeRF (DS)
  &\tablefirst$.0494$
  
  &\tablefirst$.0194$
  
  &\tablefirst$.114$
  
  &\tablesecond$.0504$
  
  &$.0363$
  
  &\tablefirst$.0305$
  
  &\tablefirst$.0426$
  
  &\tablesecond$.0377$
  
  &\tablefirst$.0308$
  
  &\tablefirst$.0456$
  
  \\   Hyper-NeRF (DS, w/ elastic)
  &\tablesecond$.0499$
  
  &$.0217$
  
  &\tablesecond$.115$
  
  &\tablefirst$.0501$
  
  &\tablesecond$.0352$
  
  &$.0332$
  
  &$.0471$
  
  &\tablethird$.0383$
  
  &\tablesecond$.0332$
  
  &\tablesecond$.0471$
  
  \\   Hyper-NeRF (AP)
  &$.0509$
  
  &\tablesecond$.0196$
  
  &$.116$
  
  &\tablethird$.0527$
  
  &\tablethird$.0361$
  
  &\tablesecond$.0319$
  
  &$.0464$
  
  &$.0386$
  
  &$.0361$
  
  &$.0476$
  
  \\   Hyper-NeRF (w/o deform)
  &$.0623$
  
  &$.0287$
  
  &$.132$
  
  &$.0552$
  
  &$.0442$
  
  &$.0410$
  
  &$.0513$
  
  &$.0453$
  
  &$.0406$
  
  &$.0557$
  
  \\ \bottomrule

\end{tabular}

}
\vspace{-12pt}
\end{table*}

\bibliographystyle{ACM-Reference-Format}
\bibliography{bibliography}

\end{document}